\documentclass[10pt,twocolumn,letterpaper]{article}

\usepackage{cvpr}              
\usepackage{fontawesome5} 
\usepackage{tikz}

\def\eg{\emph{e.g.~}} 
\def\ie{\emph{i.e.~}}

\usepackage[T1]{fontenc}
\usepackage{amssymb}   

\usepackage{cuted} 
\usepackage{graphicx}
\usepackage{tabularx}  
\usepackage{makecell}  
\usepackage{multirow}   
\usepackage{graphicx}   
\usepackage{colortbl} 
\usepackage{xcolor}
\usepackage{amsmath}
\usepackage{url}
\usepackage{enumitem}
\usepackage{algorithm}
\usepackage{algpseudocode}
\usepackage{booktabs}
\usepackage{tcolorbox}
\tcbuselibrary{breakable}  
\usepackage{caption}    
\usepackage{etoolbox}     

\definecolor{za-pink}{rgb}{0.79, 0.62, 0.70}
\definecolor{za-grey}{rgb}{0.43, 0.45, 0.48}
\definecolor{za-dark}{rgb}{0.19, 0.16, 0.12}
\definecolor{za-green}{rgb}{0.25, 0.71, 0.48}

\newcommand{\pokeza}{\textbf{\textcolor{black}{Pokémon} \textcolor{black}{Legends}:~\textcolor{black}{Z}-\textcolor{black}{A}}}

\newcommand{\no}{$\times$}
\newcommand{\yes}{\checkmark}

\setlist[itemize]{leftmargin=2em, nosep, topsep=0pt}    
\setlist[enumerate]{leftmargin=2em, nosep, topsep=0pt}

\makeatletter
\def\cvprsection{\@startsection{section}{1}{\z@}%
   {-1.5ex plus -0.5ex minus -.2ex}%
   {1pt plus 1pt minus 0.5pt}%
   {\large\bf}}     
\def\cvprsubsection{\@startsection{subsection}{2}{\z@}%
   {-1.5ex plus -0.5ex minus -.2ex}%
   {1pt plus 1pt minus 0.5pt}%
   {\elvbf}}       
\makeatother

\newtcolorbox{promptbox}[1][]{
	colback=yellow!15!white,   
	colframe=yellow!90!black, 
	fonttitle=\bfseries,    
	title=#1,               
	boxrule=0.5mm,          
	arc=2mm,                
	before upper={\setlength{\parskip}{0pt}\setlength{\parindent}{0pt}\small},
	left=2mm, right=2mm, top=2mm, bottom=2mm 
}
\definecolor{highlightgray}{gray}{0.9}

\AtBeginDocument{%
  \setlength{\abovedisplayskip}{3pt}%
  \setlength{\belowdisplayskip}{3pt}%
  \setlength{\abovedisplayshortskip}{2pt}%
  \setlength{\belowdisplayshortskip}{2pt}%
}

\makeatletter
\patchcmd{\abstract}{\vspace*{12pt}}{\vspace*{4pt}}{}{}
\patchcmd{\endabstract}{\vspace*{12pt}}{\vspace*{2pt}}{}{}
\makeatother

\definecolor{cvprblue}{rgb}{0.21,0.49,0.74}
\usepackage[pagebackref,breaklinks,colorlinks,allcolors=cvprblue]{hyperref}

\def\confName{CVPR}
\def\confYear{2027}

\title{\vspace{-40pt}PokeGym: Mastering Long-Horizon, Open-World 3D Games via \\Graph-Guided Multimodal Evolution}

\author{
Ruizhi Zhang\textsuperscript{\rm 1} \quad
Ye Huang\textsuperscript{\rm 1~\faIcon[regular]{envelope}} \quad
Yuangang Pan\textsuperscript{\rm 1} \quad
Chuanfu Shen\textsuperscript{\rm 1} \quad
Zhilin Liu\textsuperscript{\rm 1} \\
Ting Xie\textsuperscript{\rm 1} \quad
Haijun Lei\textsuperscript{\rm 2}  \quad
Wen Li\textsuperscript{\rm 1} \quad
Lixin Duan\textsuperscript{\rm 1} 
\\
\textsuperscript{\rm 1}Shenzhen Institute for Advanced Study, University of Electronic Science and Technology of China \\
    \textsuperscript{\rm 2} College of Computer Science and Software Engineering, Shenzhen University\\
}

\begin{document}

\maketitle
\begin{tikzpicture}[remember picture, overlay]
  \node[anchor=north east, inner sep=0pt] at
    ([xshift=-0.8in, yshift=-0.2in] current page.north east)
    {\includegraphics[height=0.5in]{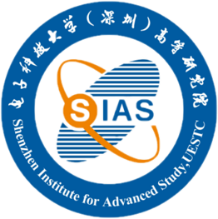}};
  \node[anchor=north east, inner sep=0pt] at
    ([xshift=-0.2in, yshift=-0.2in] current page.north east)
    {\includegraphics[height=0.5in]{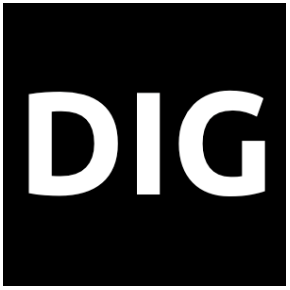}};
\end{tikzpicture}
\addtocontents{toc}{\protect\setcounter{tocdepth}{-10}}

\begin{strip}
\vspace{-18mm}
\centering
\includegraphics[width=\linewidth]{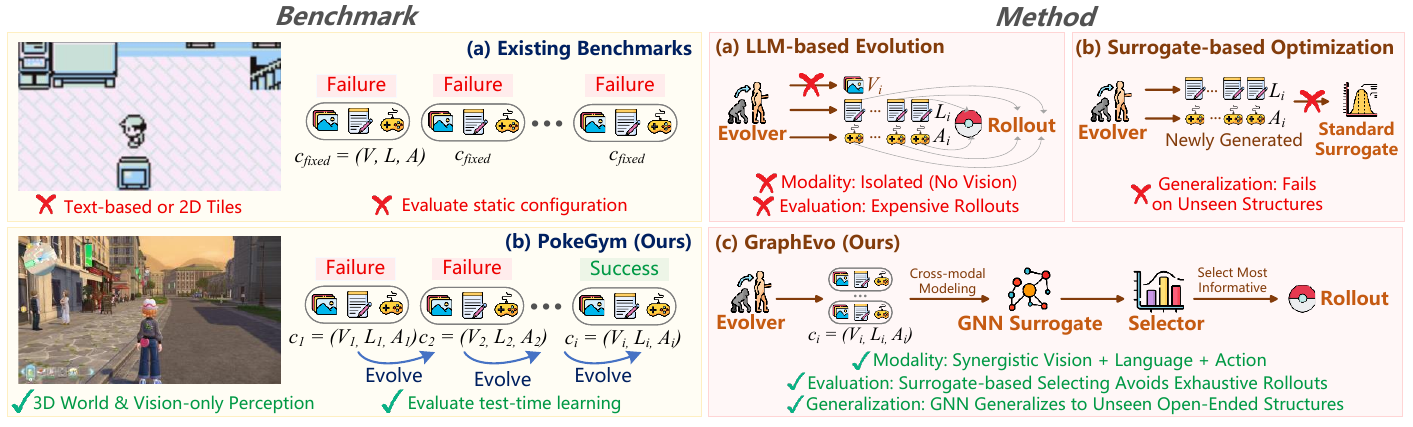}\hfill
\captionsetup{hypcap=false}
\captionof{figure}{Overview of PokeGym and GraphEvo. \textbf{Left:} PokeGym is a 3D vision-only benchmark for evaluating test-time learning through iterative evolution. \textbf{Right:} GraphEvo evolves synergistic multimodal configurations, and iterates a loop through a GNN surrogate and a selector to identify the most promising candidates among newly generated structures for evaluation.}
\captionsetup{hypcap=true}
\label{fig:teaser}%
\vspace{-3mm}
\end{strip}

\begin{abstract}

Mastering long-horizon, vision-only 3D games requires agents to improve from experience across consecutive attempts while coordinating perception, strategy, and action.
We introduce PokeGym, a benchmark built upon \pokeza, comprising 114 navigation, interaction, and target-localization tasks in dense, dynamic urban scenes. 
PokeGym evaluates test-time improvement across repeated attempts at the same task without exposing game states to the agent. 
To master PokeGym, we propose GraphEvo, a graph-guided multimodal test-time learning framework that jointly evolves visual enhancements, language strategies, and action macros while keeping the agent backbone frozen. 
A cross-modal graph predictor estimates the performance of newly generated configurations, and a disagreement-aware selector chooses candidates for execution, enabling efficient search in an open-ended configuration space without exhaustive rollouts. 
GraphEvo achieves a 60.18\% average success rate on PokeGym, outperforming the strongest baseline by over 6\%. 
Ablations demonstrate the contributions of visual adaptation, cross-modal modeling, and graph-guided selection, while experiments on ALFRED show transfer to household tasks.
\end{abstract}

\section{Introduction}
\label{sec:intro}

Superhuman performance has been witnessed in structured games such as chess, shogi, and Go \cite{silver2018general}. However, agents remain substantially less reliable in open-world, visually driven 3D games, where decisions must be grounded in visual observations rather than privileged game states \cite{balrog,lvlm_playground}.
These environments require agents to recognize task-relevant objects, reason about 3D space, and act coherently over hundreds of steps under dense scenes, dynamic actors, and changing viewpoints.
Moreover, perception, planning, and control are tightly coupled: errors in one component can propagate through the entire decision process, making a fixed agent configuration brittle across diverse situations.

Success in such environments therefore requires not only strong performance in isolated episodes, but also the ability to \textbf{improve through experience at test time}.
We refer to this setting as \textbf{test-time learning (TTL)}: an agent refines its configuration across consecutive attempts at the same task using its own trajectory feedback, without modifying its neural weights \cite{evotest}.
Existing game benchmarks, from text-based and 2D environments \cite{nethack,bellemare2013arcade} to recent 3D role-playing games \cite{doom,cradle,lumine}, mainly evaluate independent episodes with fixed configurations or transfer experience across different tasks.
They therefore do not directly measure whether an agent improves at a task by learning from its previous failures.
Although online fine-tuning or reinforcement learning could provide such adaptation, their interaction and optimization costs are impractical within a single playthrough \cite{bernerdota}.
\textbf{Self-evolving agents} offer a more feasible alternative by updating their configurations between attempts based on trajectory feedback \cite{reflexion,voyager}.

Nevertheless, existing self-evolving frameworks do not adequately address the multimodal nature of visually driven 3D interaction.
They primarily evolve components in the \textbf{language modality}, such as prompts, memory, or strategies \cite{reasoningbank}, or in the \textbf{action modality}, such as tools and executable code \cite{voyager,agentsquare,evotest}, while leaving visual perception fixed.
This separation is limiting because visual enhancements determine which scene details become salient, language strategies determine how they are interpreted, and action macros determine whether the resulting plan can be executed.
Improving one component in isolation may therefore provide little benefit when the others remain mismatched.
Effective TTL in these environments instead requires \textbf{synergistic adaptation across vision, language, and action}.

Joint multimodal adaptation also creates a difficult optimization problem: the configuration space grows rapidly, and an LLM-based evolver can generate components that did not exist when optimization began.
Existing evolutionary methods generally require every new candidate to undergo at least one expensive environment rollout before selection \cite{evotest,darwingodelmachine,agentsquare}.
Although surrogate-based optimization can reduce evaluation costs \cite{srinivas2012information,white2021bananas,moss2020boss}, conventional surrogates rely on fixed search spaces and predefined candidate coordinates, making their predictions ill-defined for newly generated components.
The community therefore lacks both \textbf{a testbed for multimodal TTL in visually driven 3D tasks} and \textbf{an efficient method for optimizing newly generated multimodal configurations}.

To address the first gap, we introduce \textbf{PokeGym}, a benchmark built upon the 3D open-world game \pokeza.
Its dense urban scenes, dynamic actors, complex spatial layouts, and multi-stage quests provide a representative testbed for visually driven, long-horizon interaction.
Agents receive only RGB observations and must complete 114 navigation, interaction, and target-localization tasks without access to privileged game states.
Crucially, PokeGym evaluates consecutive attempts at the same task and allows configurations to evolve from in-session feedback between attempts, directly measuring whether agents \textbf{improve through test-time experience}.

To address the second gap, we propose Graph-Guided Evolutionary Multimodal Agent Configuration (\textbf{GraphEvo}), which iterates among an evolver, a performance predictor, and a selector.
The \textbf{evolver} analyzes previous trajectories and proposes configurations containing visual enhancements, language strategies, and action macros.
The \textbf{performance predictor} represents each candidate as a cross-modal graph and estimates its performance with a graph neural network.
Because this graph is constructed from the candidate's content and structure rather than predefined search coordinates, the predictor remains applicable to newly generated configurations.
Finally, the \textbf{selector} uses disagreement among dropout-activated predictor sub-networks as an uncertainty signal, selecting candidates that trade off predicted performance against structural novelty.
Only selected candidates are executed, and their observed scores update the predictor for subsequent rounds.

Our main contributions are fourfold:
\begin{itemize}
    \item We introduce \textbf{PokeGym}, a vision-only 3D benchmark for evaluating whether agents can refine their configurations across consecutive attempts at the same long-horizon task.

    \item We formulate TTL as \textbf{synergistic multimodal configuration evolution} and propose GraphEvo to jointly adapt visual perception, strategies, and action.

    \item We extend surrogate-based optimization to \textbf{open-ended configuration spaces} through a graph-encoded predictor and a disagreement-aware selector, avoiding exhaustive evaluation of newly generated candidates.

    \item Extensive experiments show that GraphEvo achieves a 60.18\% average success rate on PokeGym, surpassing the strongest baseline by over 6 percentage points.
\end{itemize}

\section{Related Work}
\label{sec:related_work}

\subsection{Game-based Environments}
Games have served as ideal testbeds because they provide diverse gameplay and controllable environment \cite{hokoff, rpgbench, orak, omniplay, triangulating}. 
Text-based games such as NetHack \cite{nethack} and TextWorld \cite{cote2018textworld} have been used to study long-horizon planning and reasoning from textual observation. 
2D and grid-world games such as Pok{\'e}mon Red \cite{pokemonred} and StarDojo \cite{stardojo} provide testbeds for long-term progression and resource management from simplified 2D visuals and states. 
However, their textual or constrained visuals cannot test depth perception and spatial reasoning in complex 3D worlds.

With the rise of vision-language agents, recent works have shifted towards RGB-observation environments \cite{mcu, robust, larp, shall, wargames, lvlm_playground}.
In Minecraft, the agent enters each episode with a RL-trained frozen policy and a language prompt \cite{minedojo}. 
Cradle builds agents across diverse RPGs: its self-reflection accumulates experience within a single run, while its skill curation grows a reusable library across different tasks \cite{cradle}.
However, in both paradigms, episodes enter evaluation as independent trials: MineDojo provides no channel for context transfer between episodes at all, and Cradle's reuse operates across different tasks rather than across consecutive attempts of the same task---it even resets the agent so that trials of the same task stay mutually independent. As a result, how much an agent improves from one attempt of a task to the next remains unmeasured.

To bridge this gap, we propose PokeGym, a 3D benchmark where agents
can only perceive the game through RGB frames, and are evaluated across
consecutive episodes of the same task, where the configuration
evolves from in-session experience between attempts.
Tab.~\ref{tab:benchmark_comparison} summarizes how PokeGym differs from existing game environments.

\subsection{Self-Evolving Agents}
Instead of fine-tuning, a parallel line of work seeks to boost performance by evolving the agent prompts \cite{yang2024large,zhou2022large}.
EvoPrompt \cite{evoprompt} casts LLMs as the mutation and crossover operators of an evolutionary algorithm, generating offspring prompts from elite ones, while PromptBreeder
\cite{promptbreeder} goes one step further by co-evolving the
mutation prompts themselves, so the operators adapt alongside the population.

A second line extends this idea from single prompts to the whole agent configurations.
AgentSquare \cite{agentsquare} searches over modular combinations of planning, reasoning, memory, and tool-use components, filtering candidate combinations with an in-context LLM predictor, while EvoTest \cite{evotest} evolves the agent's system prompt, memory, hyperparameters, and tool-use routines across repeated episodes.

These methods primarily optimize isolated textual prompts and tool sets, and despite their different selection mechanisms---an LLM-as-judge
predictor confined to previously evaluated examples, or none at
all---both end up requiring every newly created component to be
evaluated directly.
By contrast, GraphEvo targets the configuration from three synergistic modalities, and brings GNN-surrogate-based optimization to newly generated configurations, avoiding exhaustive rollouts.

\section{PokeGym Benchmark}
\label{sec:pokegym}
\begin{figure*}[t] 
	\centering
	\includegraphics[width=\textwidth]{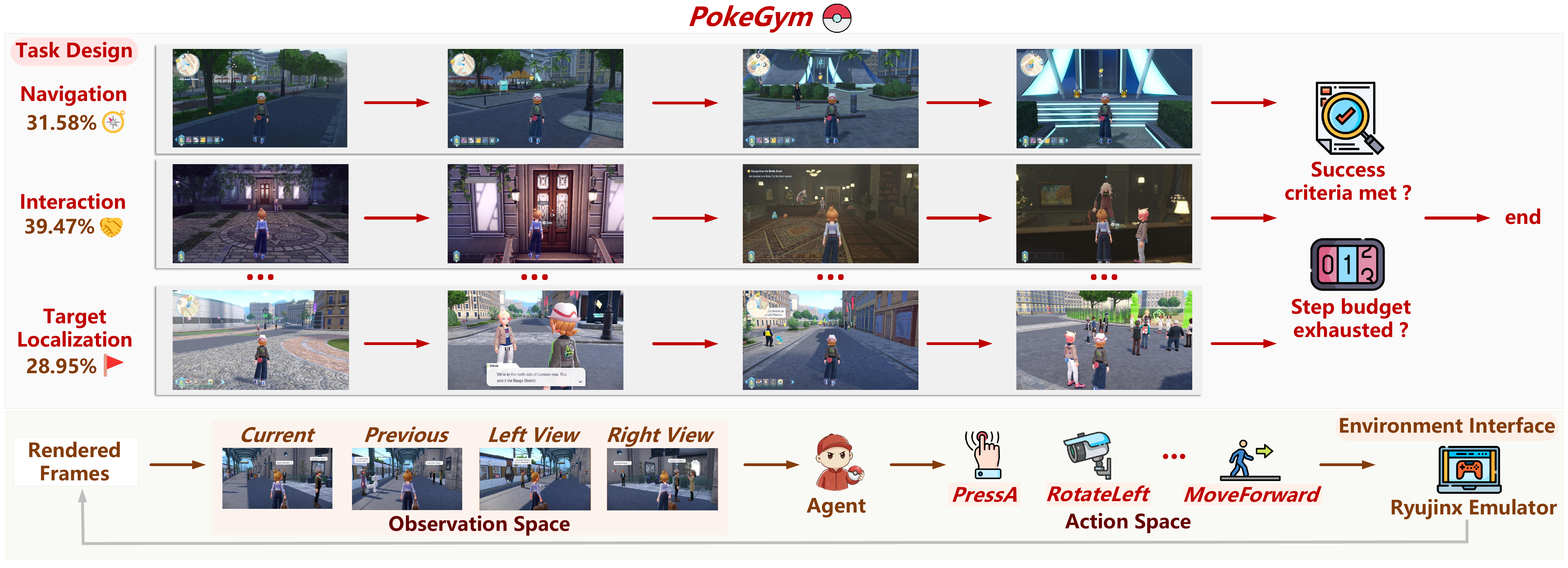} 
	
	\caption{Overview of PokeGym. Top: Task design, showing the proportion of the three task categories along with example visual trajectories. Bottom: The environment interface, where the agent perceives the game through visual observations and interacts with the emulator via high-level actions.}
\label{pokegym}
\end{figure*}
\begin{table}[t]
\centering
\caption{Comparison of game environments. 
Open World indicates unconstrained exploration; Narrative-driven indicates multi-stage narrative quests; Urban Environment indicates densely populated urban games; TTL indicates test-time learning evaluation.}
\label{tab:benchmark_comparison}
\resizebox{0.95\linewidth}{!}{%
\begin{tabular}{l l c c c c c}
\toprule
\textbf{Environment} & \textbf{Approach} & \textbf{Observation} & \textbf{Open} & \textbf{Narrative-} & \textbf{Urban} & \textbf{TTL} \\
& & & \textbf{World} & \textbf{driven} & \textbf{Env} & \\
\midrule

NetHack & IMPALA & Text & \no & \no & \no & \no \\
TextWorld & BYU & Text & \no & \no & \no & \no \\
Pok{\'e}mon Red & DRL & 2D + Text & \no & \yes & \no & \no \\
StarDojo & Zero-shot & 2D + Text & \yes & \no & \no & \no \\
MineDojo & MineAgent & 3D + Text & \yes & \no & \no & \no \\
Simulation Games & Cradle & 3D & \yes & \yes & \no & \no \\
Genshin Impact & Lumine & 3D & \yes & \yes & \yes & \no \\

\midrule
\rowcolor{gray!15} \textbf{PokeGym} & \textbf{GraphEvo} & \textbf{3D} & \textbf{\yes} & \textbf{\yes} & \textbf{\yes} & \textbf{\yes}\\
\bottomrule
\end{tabular}%
}
\end{table}

\begin{table}[t]
\centering
\caption{Robustness of the automatic evaluator.
Hit Rate denotes successful first-try detection of the task-relevant states.}
\label{tab:platform_hit_rates}
\small
\renewcommand{\arraystretch}{0.85}
\setlength{\tabcolsep}{9pt}
\begin{tabular}{l c c c}
\toprule
\textbf{Platform} & \textbf{Emulator} & \textbf{Game} & \textbf{Hit rate} \\
\midrule
Windows 10 & v1.3.2 & v1.0.0 & 100\% \\
Windows 11 & v1.3.2 & v1.0.0 & 100\% \\
Windows 11 & v1.3.3 & v1.0.0 & 100\% \\
Windows 11 & v1.3.2 & v2.0.2 & 100\% \\
Ubuntu 22.04 & v1.3.2 & v1.0.0 & 100\% \\
\bottomrule
\end{tabular}
\end{table}

\subsection{Game Environment}
We choose \pokeza{} as the evaluation environment because it poses unique visual challenges---dense scenes, dynamic actors, and changing viewpoints---while offering diverse multi-stage quests, making it well-suited for evaluating multimodal test-time learning.

\subsection{Task Design}
\label{sec::task_definitions}
PokeGym contains 114 long-horizon tasks across navigation, interaction, and target localization. 
%
Each task starts from a pre-configured file, ends upon success or step exhaustion (360 environment steps), and is verified by an automatic evaluator.

\subsection{Environment Interface}
\textbf{Observation Space.} 
At each step, the agent receives 4 RGB observations: the current front-view frame, the previous frame, and left/right view frames. 

\noindent\textbf{Action Space.} 
The action space consists of defined high-level discrete commands (\eg, MoveForward, RotateRight, PressA). 
At each step, the agent outputs an ordered sequence of up to three actions.

\subsection{Evaluation Metrics}
We evaluate agent performance with two metrics:

\begin{itemize}[nosep, topsep=0pt]
    \item \textbf{Success Rate (SR):} The percentage of episodes that successfully complete the task. 
    \item \textbf{Ineffective Move Rate (IMR):} The percentage of steps with movement actions that resulted in no spatial displacement. It evaluates the process performance, \ie whether the agent keeps repeating movements without progress.
\end{itemize}

For each episode, the task success ($v \in \{0,1\}$), the normalized IMR ($I$), and the recovery rate ($r$) are aggregated into a score used to rank configurations and train GNN:
\begin{equation}
    \text{score} = 0.5 \times v + 0.5 \times \big[0.5 \times r + 0.5 \times (1 - I)\big],
\end{equation}
where $r$ is measured as the fraction of effective-move steps immediately following an ineffective move.


\subsection{Benchmark Access}

PokeGym distributes no proprietary assets; users must legally obtain the game ROM. The task definitions, code, and evaluator will be released. 
%
A runtime memory scanner locates task-state variables to automatically verify task completion, as detailed in the appendix. \textbf{The scanner is used exclusively by the evaluator}: scanned states are never exposed to the agent, evolver, or predictor, nor included in observations or trajectory feedback. 
All methods therefore use the same vision-only interface. 
The scanner identifies the correct variables on its first attempt in all 100 runs per setting across the tested game versions, emulators, and operating systems (Tab.~\ref{tab:platform_hit_rates}), demonstrating reliable evaluation without revealing game states to the evaluated agents.

\section{The GraphEvo Framework}
\label{sec:method}

GraphEvo iterates among three modules: the evolver generates candidates from trajectories; the predictor estimates their performance; and the selector picks the most informative one to evaluate, using the observed score to update the predictor. 
As iterations proceed, this loop reinforces itself: each real evaluation refines the predictor, whose increasingly accurate estimates let the selector single out better configurations to run; the trajectories of these better runs in turn guide the evolver toward stronger candidates, so the agent improves across consecutive attempts, which is exactly what TTL demands.

\subsection{Multimodal Configuration Evolver}
The evolver applies a two-stage LLM-driven analysis to propose candidate configurations in each round.

\textbf{Trajectory Summarization.}
For each task, given the agent's rollout history including the reasoning, the LLM groups continuous steps into high-level phases and identifies critical errors, deadlocks, and visual bottlenecks. In this way, trajectories of hundreds of steps are condensed into the few decision points that matter, which the cross-trajectory analysis below can then process effectively.

\textbf{Cross-Trajectory Evolution.}
Given summarized trajectories of the task, which may contain no successful episodes, the LLM compares and analyzes them to produce a configuration containing three components:
\begin{itemize}[nosep, topsep=0pt]
    \item \textbf{Visual Enhancement:} a per-frame image-processing pipeline applied to each of the four perception frames (previous, front, left, right) at every step. Operations include crop, brightness, contrast, saturation, gamma correction, median blur, and sharpen;
    In visually-driven games, frame enhancements help the model focus on task-relevant regions under complex scenes. 
    \item \textbf{Strategy:} high-level guidelines for the agent. It shapes how the agent interprets observations, decomposes the task, and prioritizes actions;
    \item \textbf{Action Macros:} composite action functions with descriptions and executable code. They provide composite maneuvers that primitive actions cannot express---such as escaping collision deadlocks---and thereby adapt to the specific scenes.
\end{itemize}
\begin{figure*}[!t]
    \centering
    \includegraphics[width=\textwidth]{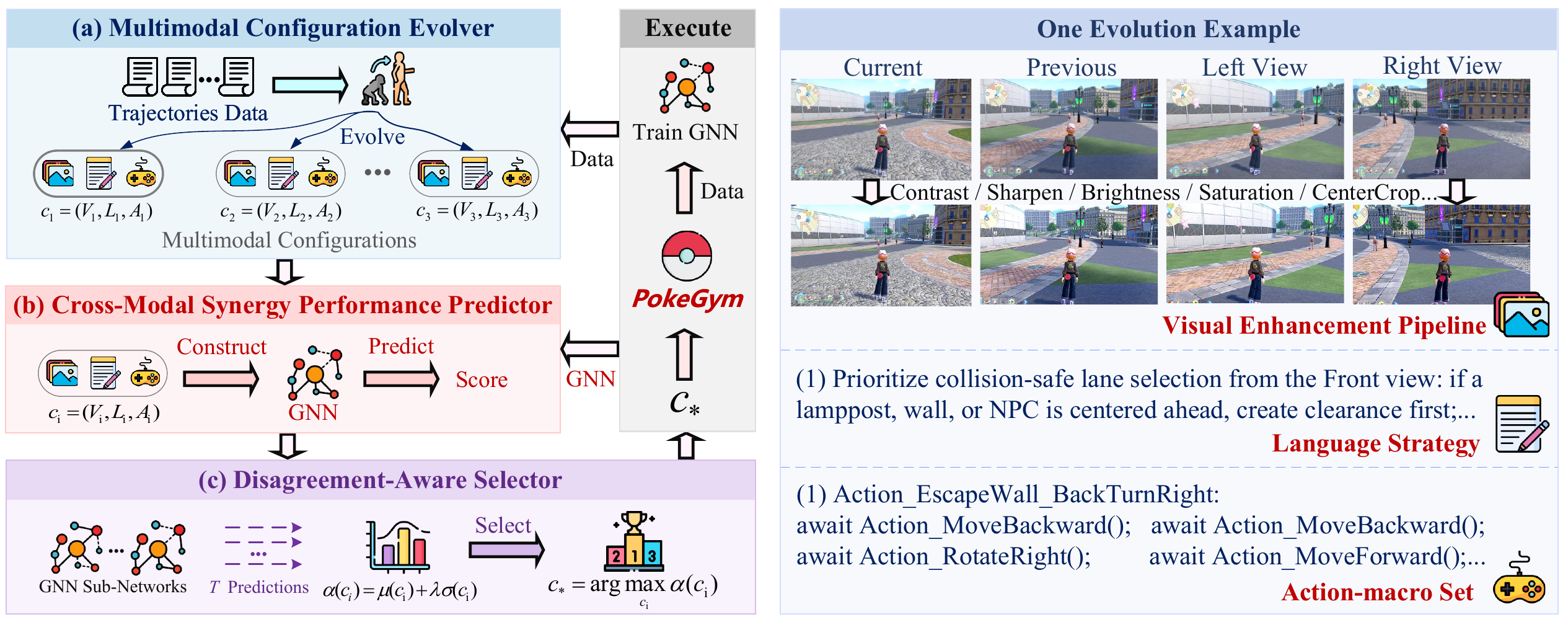}
    \caption{The GraphEvo framework: (a) The multimodal configuration evolver generates candidate configurations from historical rollouts; (b) the cross-modal synergy performance predictor estimates each candidate's score and disagreement; (c) the disagreement-aware selector picks the most informative configuration to evaluate, and the observed score is used to update the predictor for the next round.}
    \label{fig:framework}
\end{figure*}

\subsection{Cross-Modal Synergy Predictor}
Given the candidate configurations generated by the evolver, the predictor estimates the performance of each candidate without real execution.

\textbf{Configuration Encoding.}
The strategy and each macro-action description are encoded by the SigLIP2 text transformer.
For vision, we sample the four views from key frames, apply the visual-enhancement pipeline to them, and encode each enhanced view with the SigLIP2 NaFlex-ViT.

Using descriptions rather than code for macros keeps all three modalities in the same pretrained embedding space, facilitating cross-modal alignment.
The key frames are selected from a bank built from reference episodes, covering typical visual states such as the starting view, the target area, and regions with occlusion.
Using fixed key frames provides coverage of the task's visual difficulty while guaranteeing determinism and reproducibility for every candidate.

\textbf{Graph Construction.}
A configuration is represented as a graph $\mathcal{G}=(\mathcal{N},\mathcal{E},\mathbf{W})$.
Nodes in $\mathcal{N}$ are typed by modality: one for the language strategy, one per enhanced view, and one per macro action.
Each node is initialized with the corresponding encoded embedding that is $\ell_2$-normalized.
The edge set $\mathcal{E}$ connects every pair of nodes from different modalities.
Edge weight $\mathbf{W}$ stores the pairwise similarity between connected nodes.

\textbf{GNN Architecture.}
We employ a Graph Attention Network (GAT) \cite{velivckovic2017graph} as the surrogate model.
The edge weight $w_{ij}$ between nodes $i$ and $j$ is derived from cosine similarity and rescaled to $[0,1]$:
\begin{equation}
    w_{ij} = \frac{\cos(\mathbf{e}_i,\mathbf{e}_j)+1}{2},
\end{equation}
where $\mathbf{e}_i$ and $\mathbf{e}_j$ are node embeddings.
Each GAT layer updates node states by multi-head attention over their neighborhoods, with the attention logits conditioned on the edge weight $w_{ij}$.
After $D$ layers, global mean pooling aggregates the node states into a graph-level representation $\mathbf{g}$, and a two-layer MLP maps $\mathbf{g}$ to the predicted score $\hat{s}\in[0,1]$.

\textbf{Smoothness of the Predictor.}
We show that the predictor's score is Lipschitz-continuous in the component embeddings: perturbing a component by $\varepsilon$ changes the predicted score by at most $L\,\varepsilon$.
This smoothness anchors the score of an unevaluated candidate to those of similar evaluated ones, so its prediction stays within a bounded neighborhood of them rather than being arbitrary.

Let $c$ and $c'$ be two configurations whose nodes are in one-to-one correspondence, with $\|\mathbf{e}_i-\mathbf{e}_i'\|_2\le\varepsilon$ for every matched pair.
(i) The edge weight is parameter-free and Lipschitz: since node embeddings are $\ell_2$-normalized, cosine similarity is $1$-Lipschitz in each argument. 
Applying the triangle inequality separates the two endpoints, and Cauchy--Schwarz with $\|\mathbf{e}\|_2=1$ bounds each single-endpoint term; rescaling by the edge-weight factor $\tfrac{1}{2}$ gives
\begin{equation}
    |w_{ij}-w'_{ij}| \le \tfrac{1}{2}\big(\|\mathbf{e}_i-\mathbf{e}_i'\|_2+\|\mathbf{e}_j-\mathbf{e}_j'\|_2\big) \le \varepsilon.
\end{equation}
(ii) The per-layer induction (appendix) shows that each GAT layer amplifies an existing perturbation by at most a constant factor $q$ while injecting the edge-weight perturbation at rate $r$, so that $\delta^{\ell+1}\le q\,\delta^{\ell}+r\,\varepsilon$ for every layer. Unrolling it over $D$ layers, applying mean pooling---whose perturbation is independent of the node count---and the $L_{\mathrm{MLP}}$-Lipschitz head yields
\begin{gather}
    |\hat{s}(c)-\hat{s}(c')| \le L\,\varepsilon, \\
    L = L_{\mathrm{MLP}}\Big(q^D + r\sum_{t=0}^{D-1}q^{t}\Big).
\end{gather}
The constant $L$ depends only on the depth, head count, and weight norms---not on the identity of any component---so semantically similar components inherit similar scores.
In particular, when the selector faces a candidate whose components are close to evaluated ones, its prediction stays within a bounded neighborhood of theirs; a large disagreement therefore signals a structurally unseen configuration rather than local jitter of the predictor.

\textbf{Training Pipeline.}
The predictor is initialized from a warm-up set of one evaluated trajectory per task (114 in total), providing the initial supervision for score estimation and disagreement.
After each round of evaluation, the graph representations and observed scores of evaluated configurations are appended to the training set, and the GNN is further trained on the accumulated data.
This procedure repeats for $P$ rounds, progressively refining the predictor's score estimates and disagreement quantification.

\subsection{Disagreement-Aware Selector}
Given the scores produced by the predictor, the selector decides which candidate should be run next.

\textbf{Disagreement Estimation via MC-Dropout.}
The GNN is equipped with dropout layers; keeping them enabled at inference yields $T$ stochastic forward passes $\{\hat{s}^{(t)}(c_i)\}_{t=1}^{T}$ on the same graph, whose mean and standard deviation serve as $\mu(c_i)$ and $\sigma(c_i)$.
A large $\sigma(c_i)$ indicates high disagreement among the sub-networks, for configurations whose graph structure differs from evaluated ones.

\textbf{Exploration-Exploitation Selection.}
We combine the predicted mean and disagreement into a acquisition score:
\begin{equation}
    U(c_i) = \mu(c_i) + \lambda \sigma(c_i),
\end{equation}
where $\lambda\ge 0$ controls the exploration-exploitation trade-off.
The selector computes $U(c_i)$ for all candidate configurations generated in each round and selects the one with the highest score:
\begin{equation}
    c^* = \arg\max_{c_i}U(c_i).
\end{equation}
The selected configuration $c^*$ is executed on the emulator, and its ground-truth score $s^*$ is added to the training set.

\section{Experiments}
\label{sec:experiments}

\subsection{Experiment Setup}

\subsubsection{Baselines.}
We compare GraphEvo against three categories of baselines: \textbf{(1) Zero-shot VLMs} (Qwen3.6-35B \cite{qwen3.6-35b-a3b}, Claude-Sonnet-4.6 \cite{claudesonnet4.6}, GPT-5.4 \cite{gpt5.4}), directly prompted without task-specific adaptation; \textbf{(2) game-agent frameworks} (Cradle \cite{cradle} and Voyager \cite{voyager}), which learn reusable executable skills from gameplay experience; and \textbf{(3) self-evolving agents} (Reflexion \cite{reflexion}, EvoPrompt \cite{evoprompt}, PromptBreeder \cite{promptbreeder}, AgentSquare \cite{agentsquare}, and EvoTest \cite{evotest}), which adapt prompts, tool sets, or full configurations at test time. 
Detailed baseline descriptions, including how each method is adapted to PokeGym, are provided in the appendix.

\subsubsection{Implementation Details.}
We sample 10 sub-network predictions to estimate the mean and disagreement, and combine them with $\lambda=1.0$.
The predictor is trained with a learning rate of $10^{-3}$, batch size 4, and early stopping with a patience of 50 epochs.
Except for the zero-shot VLMs, all agents use GPT-5.4 as the backbone and run the evolutionary process for 10 iterations.  
Prompt-search methods (EvoPrompt and PromptBreeder) generate 6 candidate prompts per round, while configuration-search methods (AgentSquare and EvoTest) generate 3 candidate configurations per round, the same budget as GraphEvo. Other non-search baselines involve no candidate pool, so no candidate count applies: each attempt uses a single configuration directly.
All methods start each task from the same save state under the same step budget, and receive the same feedback signals at every step: RGB frames only, with no access to game states.
Each experimental setting is evaluated with 5 independent trials. The final success rate (SR) and ineffective move rate (IMR) are averaged across trials.

\subsection{Results on PokeGym}
\label{sec:main_results}

\begin{table*}[htbp]
\centering
\renewcommand{\arraystretch}{1.0}
\footnotesize
\setlength{\tabcolsep}{2pt}
\caption{Main results on PokeGym.
Except for the zero-shot VLMs, all other agents use GPT-5.4 as the backbone. 
Values are reported as mean with standard error shown in smaller font, and bold values indicate the best performance.
Modalities column indicates which agent modalities are evolved (V: visual, L: language, A: action).
$\dagger$: variants extended to evolve all three modalities.
The GraphEvo variants remove the nodes of one modality at a time, keeping the cross-modal edges among the remaining ones.}
\label{tab:main_results}
\resizebox{0.96\linewidth}{!}{%
\newcommand{\mse}[2]{#1{\tiny$\pm$#2}}
\begin{tabular*}{0.90\textwidth}{@{\extracolsep{\fill}}l c c c@{\hspace{2pt}}c c@{\hspace{2pt}}c c@{\hspace{2pt}}c c@{\hspace{2pt}}c@{}}
\toprule
\multirow{2}{*}{\textbf{Model}} & \multirow{2}{*}{\textbf{Venues}} & \multirow{2}{*}{\textbf{Modalities}} & \multicolumn{2}{c}{\textbf{Navigation}} & \multicolumn{2}{c}{\textbf{Interaction}} & \multicolumn{2}{c}{\textbf{Localization}} & \multicolumn{2}{c}{\textbf{Average}} \\
\cmidrule(lr){4-5} \cmidrule(lr){6-7} \cmidrule(lr){8-9} \cmidrule(lr){10-11}
 & & & \textbf{SR}$\uparrow$ & \textbf{IMR}$\downarrow$ & \textbf{SR}$\uparrow$ & \textbf{IMR}$\downarrow$ & \textbf{SR}$\uparrow$ & \textbf{IMR}$\downarrow$ & \textbf{SR}$\uparrow$ & \textbf{IMR}$\downarrow$ \\
\midrule
\multicolumn{11}{l}{\textbf{\textit{Zero-shot VLMs}}} \\
Qwen3.6-35B  & \multicolumn{1}{c}{--} & \multicolumn{1}{c}{--} & \mse{18.89}{4.43} & \mse{42.25}{3.45} & \mse{14.67}{3.20} & \mse{23.38}{2.50} & \mse{20.00}{3.59} & \mse{29.94}{3.62} & \mse{17.54}{2.15} & \mse{31.24}{1.94} \\
Claude-Sonnet-4.6 & \multicolumn{1}{c}{--} & \multicolumn{1}{c}{--} & \mse{22.78}{5.41} & \mse{20.97}{2.90} & \mse{16.00}{3.28} & \mse{15.54}{1.83} & \mse{17.58}{5.13} & \mse{15.76}{2.42} & \mse{18.60}{2.60} & \mse{17.32}{1.37} \\
GPT-5.4  & \multicolumn{1}{c}{--} & \multicolumn{1}{c}{--} & \mse{31.11}{5.88} & \mse{17.93}{2.59} & \mse{38.22}{4.66} & \mse{8.74}{1.34}  & \mse{42.42}{4.06} & \mse{15.05}{2.87} & \mse{37.19}{2.87} & \mse{13.47}{1.32} \\
\midrule
\multicolumn{11}{l}{\textbf{\textit{Game Agent Frameworks}}} \\
Cradle  & ICML'25 & L+A & \mse{36.11}{5.51} & \mse{20.90}{3.23} & \mse{42.67}{4.48} & \mse{9.14}{1.53}  & \mse{47.88}{4.08} & \mse{10.76}{2.45} & \mse{42.11}{2.76} & \mse{13.32}{1.45} \\
Voyager  & NeurIPS'23 & A & \mse{40.00}{5.91} & \mse{7.42}{1.36}  & \mse{52.00}{5.02} & \mse{7.69}{1.75}  & \mse{52.73}{5.63} & \mse{6.97}{1.24}  & \mse{48.42}{3.19} & \mse{7.40}{0.88}  \\
\midrule
\multicolumn{11}{l}{\textbf{\textit{Self-evolving Methods}}} \\
Reflexion  & NeurIPS'23 & L & \mse{33.89}{5.90} & \mse{13.71}{2.32} & \mse{43.11}{5.12} & \mse{9.67}{1.29}  & \mse{36.36}{5.11} & \mse{11.67}{2.09} & \mse{38.25}{3.12} & \mse{11.53}{1.08} \\
EvoPrompt  & ICLR'24 & L & \mse{40.00}{6.02} & \mse{15.43}{2.27} & \mse{50.22}{4.39} & \mse{8.63}{1.32}  & \mse{46.06}{3.64} & \mse{15.56}{2.67} & \mse{45.79}{2.78} & \mse{12.78}{1.21} \\
PromptBreeder  & ICML'24 & L & \mse{42.22}{5.95} & \mse{18.50}{2.35} & \mse{51.56}{4.56} & \mse{7.78}{1.25}  & \mse{49.09}{4.61} & \mse{16.71}{3.00} & \mse{47.89}{2.92} & \mse{13.75}{1.32} \\
AgentSquare  & ICLR'25 & L+A & \mse{37.22}{5.11} & \mse{6.42}{1.68}  & \mse{37.78}{5.21} & \mse{8.62}{1.73}  & \mse{44.85}{5.58} & \mse{8.54}{1.97}  & \mse{39.65}{3.06} & \mse{7.90}{1.03}  \\
EvoTest  & ICLR'26 & L+A & \mse{38.89}{6.32} & \mse{7.62}{1.23}  & \mse{52.00}{4.69} & \mse{7.38}{1.43}  & \mse{52.12}{4.43} & \mse{10.79}{2.46} & \mse{47.89}{3.04} & \mse{8.44}{0.99}  \\
AgentSquare$^\dagger$  & ICLR'25 & V+L+A & \mse{43.89}{5.79} & \mse{6.02}{1.20}  & \mse{55.56}{4.49} & \mse{6.88}{1.25}  & \mse{55.76}{3.77} & \mse{10.20}{2.20}  & \mse{51.93}{2.79} & \mse{7.57}{0.90}  \\
EvoTest$^\dagger$  & ICLR'26 & V+L+A & \mse{46.11}{5.95} & \mse{7.81}{1.19}  & \mse{59.11}{4.45} & \mse{6.28}{1.47}  & \mse{55.76}{4.83} & \mse{9.67}{1.97}  & \mse{54.04}{2.95} & \mse{7.75}{0.90}  \\
\midrule
\multicolumn{11}{l}{\textbf{\textit{Variants of GraphEvo}}} \\
GraphEvo w/o V & \multicolumn{1}{c}{--} & L+A & \mse{34.44}{5.94} & \mse{6.86}{1.19}  & \mse{42.67}{5.23} & \mse{7.13}{1.98}  & \mse{41.82}{5.32} & \mse{5.30}{1.01}  & \mse{39.82}{3.18} & \mse{6.52}{0.91}  \\
GraphEvo w/o L & \multicolumn{1}{c}{--} & V+A & \mse{39.44}{6.25} & \mse{8.55}{1.32}  & \mse{48.00}{4.73} & \mse{10.19}{2.19}  & \mse{53.94}{5.47} & \mse{7.31}{1.46}  & \mse{47.02}{3.16} & \mse{8.84}{1.05}  \\
GraphEvo w/o A & \multicolumn{1}{c}{--} & V+L & \mse{42.78}{5.75} & \mse{8.26}{1.79}  & \mse{55.11}{4.76} & \mse{7.63}{1.65}  & \mse{52.12}{5.83} & \mse{7.57}{1.39}  & \mse{50.35}{3.12} & \mse{7.81}{0.94}  \\
\midrule
\textbf{GraphEvo (Ours)} & \multicolumn{1}{c}{--} & \textbf{V+L+A} & \textbf{\mse{46.11}{6.11}} & \textbf{\mse{5.72}{1.01}}  & \textbf{\mse{68.00}{3.83}} & \textbf{\mse{5.76}{0.89}}  & \textbf{\mse{64.85}{4.27}} & \textbf{\mse{5.07}{0.96}}  & \textbf{\mse{60.18}{2.87}} & \textbf{\mse{5.55}{0.54}}  \\
\bottomrule
\end{tabular*}%
} 
\end{table*}

Tab.~\ref{tab:main_results} reports the main results on PokeGym.

\textbf{Extending modalities helps.}
When extended to all three modalities,AgentSquare and EvoTest gain substantially,  rising from 39.65\% to 51.93\% and from 47.89\% to 54.04\% average SR. 
Our method benefits in the same direction: dropping visual enhancement costs about 20 SR points (60.18\% to 39.82\%), and dropping strategy or action macros costs about 13 and 10 (47.02\% and 50.35\%). 
Evolving multiple modalities therefore pays off, with visual grounding as the dominant bottleneck.

\textbf{But modality extension is not the whole story.} Among zero-shot VLMs, even the strongest backbone, GPT-5.4, only reaches 37.19\% SR, highlighting the benchmark's difficulty and the necessity of test-time adaptation. Game-agent frameworks and self-evolving agents improve over zero-shot models: Voyager obtains 48.42\% SR by learning reusable skills, while EvoTest and PromptBreeder reach around 47--48\% average SR.
GraphEvo achieves an average SR of \textbf{60.18\%} and an average IMR of \textbf{5.55\%}. 
Even when AgentSquare and EvoTest are extended to evolve all three modalities (AgentSquare$^\dagger$, EvoTest$^\dagger$), they reach only 51.93\% and 54.04\% average SR, still trailing GraphEvo by more than 6 points.
%
This consistent lead suggests that the gain comes not merely from extending the evolution space to three modalities---the extended baselines share that space---but from bringing surrogate-based optimization to it: a learned graph predictor models the cross-modal synergy among components and scores newly generated configurations, leveraging the graph's smoothness to bring efficient black-box search into the open configuration space.

\subsection{Ablation Study}
\label{sec:ablation}
\subsubsection{Evidence for Cross-Modal Synergy.}
\begin{table}[htbp]
\centering
\renewcommand{\arraystretch}{1.15}
\setlength{\tabcolsep}{1pt}
\caption{Ablation on cross-modal edges in the GNN.}
\label{tab:ablation_cross}
\resizebox{\linewidth}{!}{%
\newcommand{\msei}[2]{#1{\tiny$\pm$#2}}
\begin{tabular}{l cc cc cc cc}
\toprule
\multirow{2}{*}{\textbf{GNN Variant}} & \multicolumn{2}{c}{\textbf{Navigation}} & \multicolumn{2}{c}{\textbf{Interaction}} & \multicolumn{2}{c}{\textbf{Localization}} & \multicolumn{2}{c}{\textbf{Average}} \\
\cmidrule(lr){2-3} \cmidrule(lr){4-5} \cmidrule(lr){6-7} \cmidrule(lr){8-9}
 & \textbf{SR$\uparrow$} & \textbf{IMR$\downarrow$} & \textbf{SR$\uparrow$} & \textbf{IMR$\downarrow$} & \textbf{SR$\uparrow$} & \textbf{IMR$\downarrow$} & \textbf{SR$\uparrow$} & \textbf{IMR$\downarrow$} \\
\midrule
w/o all edges      & 42.78 & 9.46  & 56.00 & 8.46  & 58.18 & 11.00  & \msei{52.46}{3.11} & \msei{9.51}{1.07}  \\
w/o V--L edges   & 46.67 & 6.85  & 53.78 & 7.56  & 53.94 & 9.99  & \msei{51.58}{2.77} & \msei{8.04}{0.95}  \\
w/o L--A edges      & 46.67 & 6.35  & 55.56 & 6.84  & 55.76 & 11.63  & \msei{52.81}{2.73} & \msei{8.07}{0.93}  \\
w/o V--A edges      & 46.67 & 5.90  & 54.22 & 5.94  & 55.76 & 8.82  & \msei{52.28}{2.86} & \msei{6.76}{0.87}  \\
\midrule
\textbf{Full GNN}   & 46.11 & \textbf{5.72}  & \textbf{68.00} & \textbf{5.76}  & \textbf{64.85} & \textbf{5.07}  & \textbf{\msei{60.18}{2.87}} & \textbf{\msei{5.55}{0.54}}  \\
\bottomrule
\end{tabular}%
} 
\end{table}

\begin{figure}[t]
\centering
\includegraphics[width=\columnwidth]{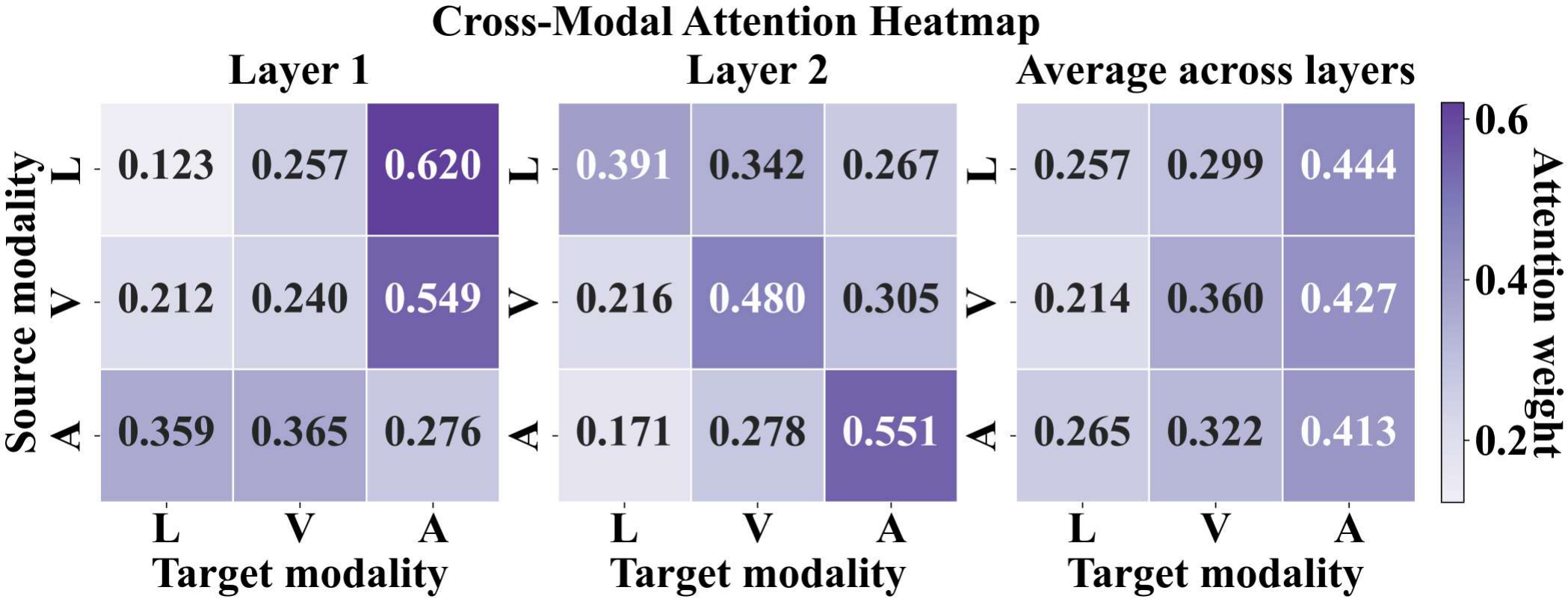}
\caption{Cross-modal attention heatmap of the GNN predictor. Each cell shows the normalized attention weight. Higher values indicate stronger cross-modal dependency.}
\label{fig:heatmap}
\end{figure}

\begin{figure}[t]
	\centering
	\includegraphics[width=\columnwidth]{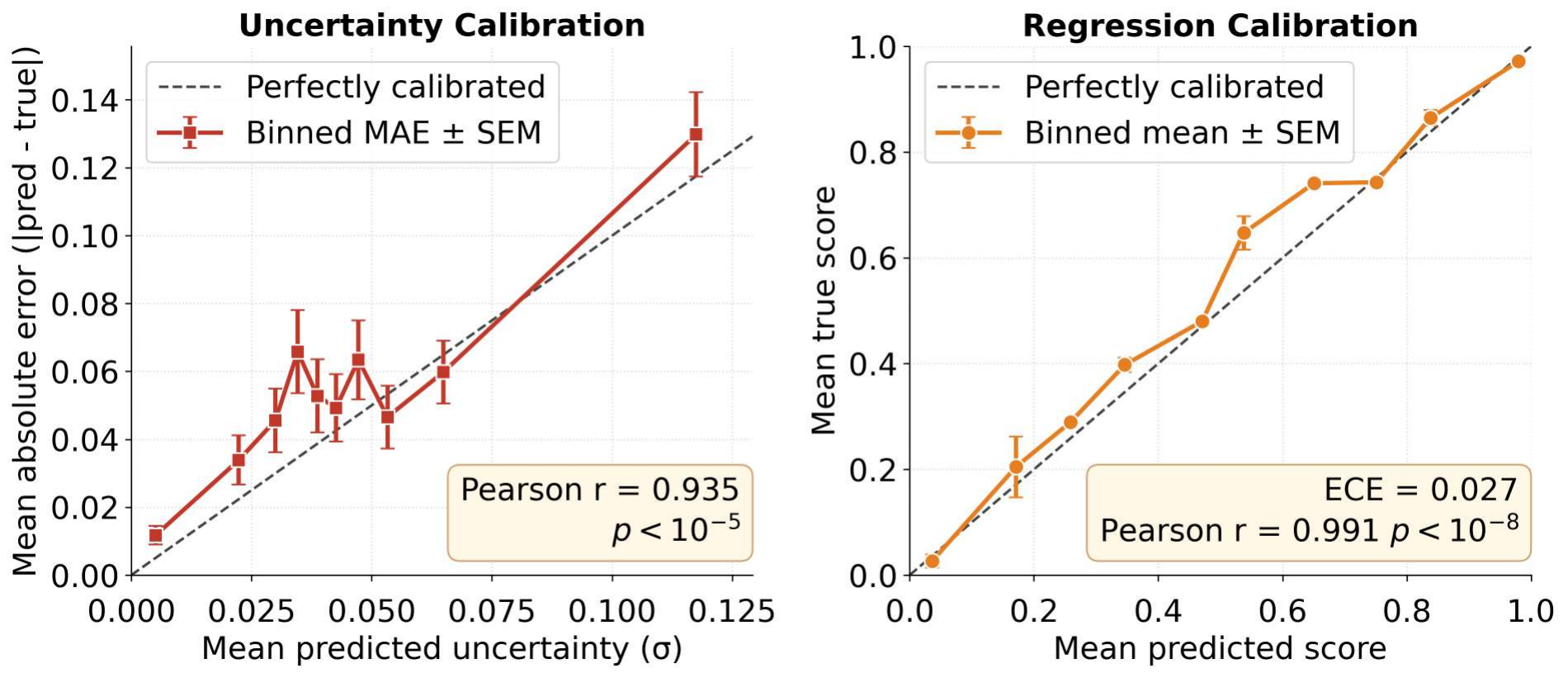}
	\caption{Calibration of the GNN predictor. Left: predicted uncertainty ($\sigma$) versus mean absolute error. Right: mean predicted score versus mean true score.}
	\label{fig:calibration}
\end{figure}

Tab.~\ref{tab:ablation_cross} ablates the cross-modal edges of the GNN while keeping every node and its embedding vector unchanged. 
Removing all cross-modal edges drops the average SR from 60.18\% to 52.46\% and raises the IMR from 5.55\% to 9.51\%; removing each individual edge type also degrades performance, with the vision--language edges contributing the most to SR.
This confirms that our gains come not merely from evolving more modalities, but from letting them mutually enhance each other through the cross-modal interactions.
In Fig.~\ref{fig:heatmap}, the strong links between the language, visual, and action nodes suggest that the predictor captures meaningful cross-modal synergy rather than treating modalities independently.

%

\subsubsection{Effective Search with the Predictor.}
\begin{table}[htbp]
\centering
\renewcommand{\arraystretch}{1.15}
\setlength{\tabcolsep}{1pt}
\caption{Comparison with and without the GNN predictor.}
\label{tab:ablation_exhaust}
\resizebox{\linewidth}{!}{%
\newcommand{\mse}[2]{\begin{tabular}{@{}c@{}}#1\\[-5pt]{\tiny$\pm$#2}\end{tabular}}
\newcommand{\msei}[2]{#1{\tiny$\pm$#2}}
\begin{tabular}{l cc cc cc cc}
\toprule
\multirow{2}{*}{\textbf{Method}} & \multicolumn{2}{c}{\textbf{Navigation}} & \multicolumn{2}{c}{\textbf{Interaction}} & \multicolumn{2}{c}{\textbf{Localization}} & \multicolumn{2}{c}{\textbf{Average}} \\
\cmidrule(lr){2-3} \cmidrule(lr){4-5} \cmidrule(lr){6-7} \cmidrule(lr){8-9}
 & \textbf{SR$\uparrow$} & \textbf{IMR$\downarrow$} & \textbf{SR$\uparrow$} & \textbf{IMR$\downarrow$} & \textbf{SR$\uparrow$} & \textbf{IMR$\downarrow$} & \textbf{SR$\uparrow$} & \textbf{IMR$\downarrow$} \\
\midrule
w/o Predictor & 41.11 & 9.66  & 51.56 & 7.75  & 56.36 & 11.92  & \msei{49.65}{2.93} & \msei{9.56}{1.10}  \\
\midrule
\textbf{GNN Predictor} & \textbf{46.11} & \textbf{5.72}  & \textbf{68.00} & \textbf{5.76}  & \textbf{64.85} & \textbf{5.07}  & \textbf{\msei{60.18}{2.87}} & \textbf{\msei{5.55}{0.54}}  \\
\bottomrule
\end{tabular}%
} 
\end{table}

Under the same simulator rollouts (Tab.~\ref{tab:ablation_exhaust}), the no-predictor variant cannot estimate the performance of generated configurations and must roll out every one of them to obtain the true score, whereas our surrogate scores all candidates and rolls out only the selected one, updating the GNN after each round. The no-predictor variant therefore iterates fewer rounds: the average SR drops from 60.18\% to 49.65\% and the IMR rises from 5.55\% to 9.56\%. This confirms that the predictor is not merely a speedup: by screening candidates before evaluation, it converts the same rollout count into a far more effective search.

In Fig.~\ref{fig:calibration}, the binned mean absolute error closely follows the predicted uncertainty (Pearson $r=0.935$, $p<10^{-5}$), and the mean predicted scores align with the mean true scores (Pearson $r=0.991$, ECE $=0.027$). 
Thus, the predictor's scores and disagreement are calibrated enough to serve as the selector's decision basis: high scores safely support
exploitation, while high disagreement correctly flags candidates worth exploring.

\subsubsection{Topology Awareness of the Predictor.}
\begin{table}[htbp]
\centering
\renewcommand{\arraystretch}{1.15}
\setlength{\tabcolsep}{1pt}
\caption{Comparison of different predictors.}
\label{tab:ablation_gnn}
\resizebox{\linewidth}{!}{%
\newcommand{\mse}[2]{\begin{tabular}{@{}c@{}}#1\\[-5pt]{\tiny$\pm$#2}\end{tabular}}
\newcommand{\msei}[2]{#1{\tiny$\pm$#2}}
\begin{tabular}{l cc cc cc cc}
\toprule
\multirow{2}{*}{\textbf{Predictor}} & \multicolumn{2}{c}{\textbf{Navigation}} & \multicolumn{2}{c}{\textbf{Interaction}} & \multicolumn{2}{c}{\textbf{Localization}} & \multicolumn{2}{c}{\textbf{Average}} \\
\cmidrule(lr){2-3} \cmidrule(lr){4-5} \cmidrule(lr){6-7} \cmidrule(lr){8-9}
 & \textbf{SR$\uparrow$} & \textbf{IMR$\downarrow$} & \textbf{SR$\uparrow$} & \textbf{IMR$\downarrow$} & \textbf{SR$\uparrow$} & \textbf{IMR$\downarrow$} & \textbf{SR$\uparrow$} & \textbf{IMR$\downarrow$} \\
\midrule
LLM-as-Judge    & 30.00 & 7.92  & 36.44 & 8.42  & 38.18 & 9.83  & \msei{34.91}{3.04} & \msei{8.67}{0.98}  \\
MLP             & 38.33 & 9.07  & 51.56 & 6.09  & 56.36 & 11.69 & \msei{48.77}{3.12} & \msei{8.65}{0.99}  \\
\midrule
\textbf{GNN}    & \textbf{46.11} & \textbf{5.72}  & \textbf{68.00} & \textbf{5.76}  & \textbf{64.85} & \textbf{5.07}  & \textbf{\msei{60.18}{2.87}} & \textbf{\msei{5.55}{0.54}}  \\
\bottomrule
\end{tabular}%
} 
\end{table}

Tab.~\ref{tab:ablation_gnn} compares three choices for the performance predictor.
A plain MLP predictor reaches 48.77\% average SR and 8.65\% IMR, which is better than LLM-as-Judge (34.91\% SR and 8.67\% IMR), but still far below the GNN predictor (60.18\% SR and 5.55\% IMR).
The gap has a structural explanation: the GNN captures the topology of a configuration---how its components relate to one another---whereas the MLP flattens the same configuration into a fixed vector and discards those relations, and LLM-as-Judge has no learnable structural representation at all, parsing raw text for every prediction.

\subsubsection{Disagreement Awareness of the Selector.}
\begin{table}[htbp]
\centering
\renewcommand{\arraystretch}{1.15}
\setlength{\tabcolsep}{1pt}
\caption{Ablation on the configuration selection strategy.}
\label{tab:ablation_selector}
\resizebox{\linewidth}{!}{%
\newcommand{\mse}[2]{\begin{tabular}{@{}c@{}}#1\\[-5pt]{\tiny$\pm$#2}\end{tabular}}
\newcommand{\msei}[2]{#1{\tiny$\pm$#2}}
\begin{tabular}{l cc cc cc cc}
\toprule
\multirow{2}{*}{\textbf{Selector}} & \multicolumn{2}{c}{\textbf{Navigation}} & \multicolumn{2}{c}{\textbf{Interaction}} & \multicolumn{2}{c}{\textbf{Localization}} & \multicolumn{2}{c}{\textbf{Average}} \\
\cmidrule(lr){2-3} \cmidrule(lr){4-5} \cmidrule(lr){6-7} \cmidrule(lr){8-9}
 & \textbf{SR$\uparrow$} & \textbf{IMR$\downarrow$} & \textbf{SR$\uparrow$} & \textbf{IMR$\downarrow$} & \textbf{SR$\uparrow$} & \textbf{IMR$\downarrow$} & \textbf{SR$\uparrow$} & \textbf{IMR$\downarrow$} \\
\midrule
Explore-only    & 29.44 & 8.57  & 40.44 & 8.94  & 35.15 & 8.50  & \msei{35.44}{3.18} & \msei{8.70}{0.97}  \\
Exploit-only    & 38.89 & 5.95  & 62.22 & 6.59  & 52.12 & 6.50  & \msei{51.93}{2.99} & \msei{6.36}{0.63}  \\
Thompson Sampling    & 35.00 & 6.05  & 45.33 & 7.69  & 41.21 & 7.08  & \msei{40.88}{3.23} & \msei{7.00}{0.93}  \\
Deep Ensembles    & 46.11 & 9.54  & 60.44 & 6.89  & 56.36 & 10.88  & \msei{54.74}{2.87} & \msei{8.88}{0.99}  \\
LinUCB    & 41.11 & 7.23  & 58.22 & 7.12  & 52.12 & 10.75  & \msei{51.05}{2.99} & \msei{8.21}{0.94}  \\
\midrule
\textbf{GraphEvo} & 46.11 & \textbf{5.72}  & \textbf{68.00} & \textbf{5.76}  & \textbf{64.85} & \textbf{5.07}  & \textbf{\msei{60.18}{2.87}} & \textbf{\msei{5.55}{0.54}}  \\
\bottomrule
\end{tabular}%
} 
\end{table}

Tab.~\ref{tab:ablation_selector} studies the configuration selection strategy. 
Pure exploitation and pure exploration both underperform, since one-sided strategies fail to balance the two. 
Alternative uncertainty baselines such as Thompson Sampling, Deep Ensembles, and LinUCB also fall short: their exploration signals come either from training randomness or from how often a candidate has been tried---never from the candidate's own structure. 
In contrast, our selector reaches 60.18\% SR and 5.55\% IMR: dropout disagreement over the GNN's sub-networks is computed from the candidate's own graph structure, and flags what has never been seen, providing a more reliable exploration-exploitation balance.

\subsection{Generalization Study}

\subsubsection{Generalization to Another VLM Backbone.}
\begin{table}[htbp]
\centering
\renewcommand{\arraystretch}{1.15}
\setlength{\tabcolsep}{1pt}
\caption{Performance comparison using Qwen3.6-35B.}
\label{tab:ablation_qwen}
\resizebox{\linewidth}{!}{%
\newcommand{\mse}[2]{\begin{tabular}{@{}c@{}}#1\\[-5pt]{\tiny$\pm$#2}\end{tabular}}
\newcommand{\msei}[2]{#1{\tiny$\pm$#2}}
\begin{tabular}{l cc cc cc cc}
\toprule
\multirow{2}{*}{\textbf{Model}} & \multicolumn{2}{c}{\textbf{Navigation}} & \multicolumn{2}{c}{\textbf{Interaction}} & \multicolumn{2}{c}{\textbf{Localization}} & \multicolumn{2}{c}{\textbf{Average}} \\
\cmidrule(lr){2-3} \cmidrule(lr){4-5} \cmidrule(lr){6-7} \cmidrule(lr){8-9}
 & \textbf{SR$\uparrow$} & \textbf{IMR$\downarrow$} & \textbf{SR$\uparrow$} & \textbf{IMR$\downarrow$} & \textbf{SR$\uparrow$} & \textbf{IMR$\downarrow$} & \textbf{SR$\uparrow$} & \textbf{IMR$\downarrow$} \\
\midrule
Zero-shot              & 18.89 & 42.25 & 14.67 & 23.38 & 20.00 & 29.94 & \msei{17.54}{2.15} & \msei{31.24}{1.94} \\
PromptBreeder          & 22.78 & 33.41 & 17.78 & 17.55 & 29.09 & 24.55 & \msei{22.63}{2.33} & \msei{24.59}{1.73} \\
Voyager                & 26.67 & 18.08 & 21.78 & 15.29 & 28.48 & 19.73 & \msei{25.26}{2.38} & \msei{17.46}{1.38} \\
\midrule
\textbf{GraphEvo} & \textbf{31.11} & \textbf{17.96} & \textbf{32.44} & \textbf{13.70} & \textbf{36.36} & \textbf{15.64} & \textbf{\msei{33.16}{2.54}} & \textbf{\msei{15.61}{1.13}} \\
\bottomrule
\end{tabular}%
} 
\end{table}

Tab.~\ref{tab:ablation_qwen} validates GraphEvo with the Qwen3.6-35B backbone. Starting from a 17.54\% zero-shot SR, GraphEvo raises it to 33.16\% SR with gains across all task types, outperforming both Voyager and PromptBreeder. This shows that the framework is not tied to a single VLM and can improve weaker backbones by adapting their visual, language, and action configurations at test time.
Generalization to newer frontier backbones (\eg, GPT-6.1-sol) is reported in the appendix.

\subsubsection{Cross-Domain Generalization on ALFRED.}
\label{sec:alfred}
\begin{table}[htbp]
\centering
\renewcommand{\arraystretch}{1.0}
\fontsize{7.5pt}{8pt}\selectfont
\setlength{\tabcolsep}{0pt}
\caption{Results on ALFRED using the Qwen3.6-35B. We report goal-condition success (GCS) and path-weighted goal-condition success (PW-GCS, abbreviated as PW).}
\label{tab:alfred_cross_domain}
\newcommand{\mse}[2]{#1{\tiny$\pm$#2}}
\begin{tabular*}{\columnwidth}{@{\extracolsep{\fill}}l*{8}{c}@{}}
\toprule
\multirow{2}{*}{\textbf{Model}} & \multicolumn{2}{c}{\textbf{State}} & \multicolumn{2}{c}{\textbf{Inspection}} & \multicolumn{2}{c}{\textbf{Spatial}} & \multicolumn{2}{c}{\textbf{Average}} \\
\cmidrule(lr){2-3} \cmidrule(lr){4-5} \cmidrule(lr){6-7} \cmidrule(lr){8-9}
 & \textbf{GCS}$\uparrow$ & \textbf{PW}$\uparrow$ & \textbf{GCS}$\uparrow$ & \textbf{PW}$\uparrow$ & \textbf{GCS}$\uparrow$ & \textbf{PW}$\uparrow$ & \textbf{GCS}$\uparrow$ & \textbf{PW}$\uparrow$ \\
\midrule
Zero-shot & 12.22 & 10.34 & 10.00 & 6.55 & 0.00 & 0.00 & \mse{8.05}{2.02} & \mse{6.48}{1.72} \\
Reflexion & 12.22 & 6.79 & 5.00 & 3.64 & 0.00 & 0.00 & \mse{7.18}{1.88} & \mse{4.14}{1.29} \\
EvoTest & 11.11 & 9.24 & 20.00 & 8.74 & 0.00 & 0.00 & \mse{9.20}{2.21} & \mse{6.28}{1.67} \\
Voyager & 11.11 & 7.94 & 30.00 & 25.00 & 0.00 & 0.00 & \mse{10.92}{2.41} & \mse{8.42}{2.04} \\
\textbf{GraphEvo} & 12.22 & \textbf{12.22} & \textbf{45.00} & \textbf{43.91} & \textbf{5.56} & \textbf{4.93} & \textbf{\mse{15.80}{3.26}} & \textbf{\mse{15.42}{3.15}} \\
\bottomrule
\end{tabular*}%
\end{table}

We assess cross-domain generalization by adapting GraphEvo and the baselines on ALFRED \cite{alfred}.
ALFRED is a household-task benchmark built on the interactive AI2-THOR simulator \cite{ai2thor}, where agents must locate objects in indoor rooms, manipulate them, and place them at designated spots---shifting the challenge from PokeGym's urban navigation to object-centric manipulation.
We use the 174 validation-unseen task instances spanning the six most challenging types and group them into three categories: State-Changing (State), Spatial, and Inspection.
Detailed task descriptions are provided in the appendix.

\Cref{tab:alfred_cross_domain} reports results on ALFRED.
GraphEvo improves over all baselines on Spatial, Inspection, and average metrics, achieving the highest average GCS (15.80\%) and PW-GCS (15.42\%). 
The largest gains fall on Inspection tasks, which require locating an object and examining it under a light source---the most perception-intensive category---where GraphEvo reaches 45.00\% GCS versus 30.00\% for Voyager and 10.00\% for zero-shot. 
These results show that our framework transfers across task families, and that its benefit concentrates where the visual bottleneck is heaviest, effectively mitigating it.

\section{Conclusion}


We present PokeGym, a vision-only benchmark for studying test-time learning through consecutive attempts at long-horizon tasks in open-world 3D games. 
To jointly adapt perception, strategy, and action, we introduce GraphEvo, which evolves multimodal configurations and selects promising candidates with a graph-based predictor and a disagreement-aware selector, avoiding exhaustive evaluation.
GraphEvo achieves a 60.18\% average success rate on PokeGym, outperforming the strongest baseline by over 6\%. 
Ablations further highlight the importance of visual adaptation, cross-modal modeling, and efficient candidate selection. 
Together, PokeGym and GraphEvo provide a testbed and an effective approach for multimodal test-time learning in visually driven 3D environments. 

\clearpage


{
    \small
    \bibliographystyle{ieeenat_fullname}
    \bibliography{main}
}

\clearpage

\appendix

\renewcommand{\contentsname}{Appendix Contents}   
\addtocontents{toc}{\protect\setcounter{tocdepth}{2}}  
\tableofcontents


\section{Extended Environment Comparison}
\label{app:extended_benchmark_comparison}

PokeGym complements existing game benchmarks by combining visually grounded 3D interaction, multi-stage narrative tasks, and repeated-episode evaluation of agent adaptation.
The distinction is not whether an environment permits learning, but whether its evaluation protocol explicitly measures how an agent improves its configuration across successive attempts at the same task.
We compare PokeGym with two representative environments below.

\subsection{Comparison with Pok\'{e}mon Red}
\label{app:cmp_pokemon_red}

Pok\'{e}mon Red~\cite{pokemonred} provides a tile-based 2D world with a relatively stable top-down view.
Although agents may still operate from pixels, its discrete spatial layout and recurring visual patterns provide consistent cues for navigation and interaction.
In contrast, PokeGym uses perspective-rendered 3D scenes with an avatar-following camera.
Changes in viewpoint, occlusion, and moving characters make task-relevant cues less consistent across observations.
Agents must therefore coordinate visual interpretation, navigation strategies, and action execution under changing views.
This provides a complementary setting for evaluating whether test-time configuration updates improve visually grounded behavior.

\subsection{Comparison with Minecraft}
\label{app:cmp_minecraft}

Minecraft-based environments~\cite{minedojo,voyager} support diverse tasks, including resource gathering, crafting, exploration, and construction, and have enabled substantial work on open-ended skill acquisition.
Their voxel-based worlds and commonly used task formulations offer a different combination of visual structure and objectives from PokeGym.

PokeGym emphasizes multi-stage narrative quests in a 3D urban environment.
Progress requires interpreting dialogue and completing ordered interactions with characters and locations, rather than satisfying only an inventory or construction objective.
Its repeated-episode protocol and automatic task-state evaluation make improvements across attempts measurable and reproducible.
PokeGym thus complements open-ended skill-learning environments with a controlled setting for evaluating multimodal configuration adaptation on narrative tasks.

\section{Proof of the Lipschitz Bound}
\label{app:lipschitz}

We prove that the GNN surrogate is \textbf{Lipschitz-continuous with respect to its component embeddings} below. Lipschitz continuity means that the change in predicted score is bounded by a fixed constant times the size of the input perturbation; it therefore formalizes the predictor's smoothness by limiting how strongly small embedding changes can affect its output.

This property provides a stability basis for surrogate-guided search over newly generated configurations. Under the graph-matching assumptions of the proof, a candidate close to an evaluated configuration in component-embedding space cannot receive an arbitrarily different predicted score. The surrogate can thus extend its learned scoring function smoothly to nearby, unevaluated candidates, supporting candidate screening before expensive emulator rollouts. This is a guarantee of prediction stability.

\subsection{Statement and Assumptions}
Let $c$ and $c'$ be two configurations whose nodes can be matched one-to-one by modality. Let $\varepsilon\ge0$ bound the Euclidean distance between every pair of matched, normalized component embeddings.
Our goal is to show that, for a fixed trained predictor, there exists a finite constant $L$, independent of the identities of these components, such that
\begin{equation}
|\hat{s}(c)-\hat{s}(c')|\le L\varepsilon,
\end{equation}
where $\hat{s}(c)$ and $\hat{s}(c')$ are the predicted scores and $L$ is a Lipschitz constant determined by the network architecture and parameter norms.
The proof propagates the input perturbation $\varepsilon$ through similarity-based edge weights, multi-head graph attention, global mean pooling, and the score head, deriving an explicit expression for $L$.

Let $\mathcal{N}$ be the common node-index set obtained by matching nodes of the same modality, and let $\mathcal{E}$ be the common set of cross-modal edges. Indices $i$ and $j$ denote nodes, while $m$ is a dummy neighbor index. We use $\mathcal{N}(i)$ to denote the attention neighborhood of node $i$. Let $\mathbf{e}_i$ denote the normalized embedding of component $i$ in configuration $c$. A prime denotes the corresponding quantity for $c'$, so $\mathbf{e}'_i$ is the embedding of its matched component. These embeddings satisfy
\begin{equation}
\|\mathbf{e}_i\|_2=\|\mathbf{e}'_i\|_2=1,
\qquad
\max_i\|\mathbf{e}_i-\mathbf{e}'_i\|_2\le\varepsilon.
\end{equation}
We compare predictions from the same fixed network parameters. All matrix norms below are spectral norms, and vector norms are Euclidean unless otherwise indicated.

Let $D$ be the number of GAT layers and $K$ the number of attention heads per layer. The indices $\ell\in\{0,\ldots,D-1\}$ and $k\in\{1,\ldots,K\}$ identify a layer and an attention head, respectively. For head $k$, $\mathbf{W}^{\ell,k}$ projects node states at layer $\ell$, $\mathbf{E}^{k}$ projects the scalar edge weight into the attention feature space, and $\mathbf{u}^{k}$ is the attention vector that maps concatenated features to a scalar logit. Let $B_W$, $B_E$, and $B_u$ be finite, nonnegative upper bounds on their norms, uniform over all relevant layers and heads:
\begin{equation}
\|\mathbf{W}^{\ell,k}\|_2\le B_W,\quad
\|\mathbf{E}^{k}\|_2\le B_E,\quad
\|\mathbf{u}^{k}\|_2\le B_u.
\end{equation}
The proof uses the bias-free GAT update below, with ReLU for message updates and a LeakyReLU whose negative slope lies in $[0,1]$ for attention logits. Both activations are $1$-Lipschitz. Fixed biases in the attention logits cancel when taking differences; biases in message updates can also be accommodated by enlarging the hidden-state bound.
Let $\mathbf{h}_i^\ell$ be the hidden state of node $i$ after $\ell$ GAT layers, with layer $0$ denoting the input embeddings. Define $\delta^\ell$ as the largest hidden-state difference between matched nodes at that layer:
\begin{equation}
\delta^\ell=\max_i\|\mathbf{h}_i^\ell-\mathbf{h}_i'^{\ell}\|_2,
\qquad \mathbf{h}_i^0=\mathbf{e}_i.
\end{equation}
Thus $\delta^0\le\varepsilon$. The same definitions apply to the primed graph.

\subsection{Step 1: Edge-Weight Stability}
For an edge connecting nodes $i$ and $j$, let $w_{ij}\in[0,1]$ be its similarity-based weight and $w'_{ij}$ its counterpart in the second graph. For normalized embeddings, this weight is
\begin{equation}
w_{ij}=\frac{\mathbf{e}_i^\top\mathbf{e}_j+1}{2}.
\end{equation}
Adding and subtracting $\mathbf{e}_i'^\top\mathbf{e}_j$ separates the effects of the two endpoints:
\begin{align}
2|w_{ij}-w'_{ij}|
&=|\mathbf{e}_i^\top\mathbf{e}_j-\mathbf{e}_i'^\top\mathbf{e}'_j|\notag\\
&\le |(\mathbf{e}_i-\mathbf{e}'_i)^\top\mathbf{e}_j|
 +|\mathbf{e}_i'^\top(\mathbf{e}_j-\mathbf{e}'_j)|\notag\\
&\le \|\mathbf{e}_i-\mathbf{e}'_i\|_2
 +\|\mathbf{e}_j-\mathbf{e}'_j\|_2.
\end{align}
The final inequality follows from Cauchy--Schwarz and unit embedding norms. Consequently,
\begin{equation}
\max_{(i,j)\in\mathcal{E}}|w_{ij}-w'_{ij}|\le\varepsilon.
\label{eq:app_edge_bound}
\end{equation}
Edge weights are computed from the initial embeddings and reused across layers; their perturbation therefore remains bounded by $\varepsilon$ throughout message passing.

\subsection{Step 2: Hidden-State Bounds}
Let $\beta_{ij}^{\ell,k}$ be the unnormalized attention logit from node $i$ to neighbor $j$, $\alpha_{ij}^{\ell,k}$ its normalized attention coefficient, and $\mathbf{z}_i^{\ell,k}$ the updated node representation produced by head $k$. Their computation and the concatenated node update are
\begin{equation}
\resizebox{0.86\columnwidth}{!}{$\displaystyle
\beta_{ij}^{\ell,k}
=\operatorname{LeakyReLU}\!\left((\mathbf{u}^{k})^\top
\left[\mathbf{W}^{\ell,k}\mathbf{h}_i^\ell
\,\|\,\mathbf{W}^{\ell,k}\mathbf{h}_j^\ell
\,\|\,\mathbf{E}^{k}w_{ij}\right]\right)$}.
\end{equation}
\begin{gather}
\alpha_{ij}^{\ell,k}
=\frac{\exp(\beta_{ij}^{\ell,k})}
{\sum_{m\in\mathcal{N}(i)}\exp(\beta_{im}^{\ell,k})},\\
\mathbf{z}_i^{\ell,k}
=\operatorname{ReLU}\!\left(\mathbf{W}^{\ell,k}
\sum_{j\in\mathcal{N}(i)}\alpha_{ij}^{\ell,k}\mathbf{h}_j^\ell\right),\\
\mathbf{h}_i^{\ell+1}
=\mathbin\|_{k=1}^{K}\mathbf{z}_i^{\ell,k}.
\end{gather}
Here $[\cdot\,\|\,\cdot]$ denotes feature concatenation and $\mathbin\|_{k=1}^K$ concatenates the head outputs. Since attention coefficients are nonnegative and sum to one, each head aggregates a convex combination of node states. Let $H_\ell$ bound $\|\mathbf{h}_i^\ell\|_2$ and $\|\mathbf{h}_i'^{\ell}\|_2$ for all nodes in both graphs. Then
\begin{equation}
\|\mathbf{z}_i^{\ell,k}\|_2\le B_W H_\ell,
\qquad H_{\ell+1}\le\sqrt{K}B_W H_\ell.
\end{equation}
Starting from $H_0=1$, induction gives $H_\ell\le(\sqrt{K}B_W)^\ell$. Let $H$ denote a single hidden-state norm bound valid for every layer of both graphs. We may choose
\begin{equation}
H=\max_{0\le\ell\le D}(\sqrt{K}B_W)^\ell
=\max\{1,(\sqrt{K}B_W)^D\}.
\label{eq:app_hidden_bound}
\end{equation}
The maximum with $1$ is necessary when $\sqrt{K}B_W<1$; the final-layer bound alone would not bound earlier layers in that case.

\subsection{Step 3: Attention Stability}
Set $b=B_uB_W$ and $e=B_uB_E$, scalar constants bounding sensitivity of the attention logits to node states and edge weights, respectively. The scalar $e$ is distinct from the embedding vector $\mathbf{e}_i$. By the Lipschitz property of LeakyReLU, the weight-norm bounds, and the triangle inequality,
\begin{align}
|\beta_{ij}^{\ell,k}-\beta_{ij}'^{\ell,k}|
&\le B_u\bigl(B_W\|\mathbf{h}_i^\ell-\mathbf{h}_i'^{\ell}\|_2\notag\\
&\quad+B_W\|\mathbf{h}_j^\ell-\mathbf{h}_j'^{\ell}\|_2
+B_E|w_{ij}-w'_{ij}|\bigr)\notag\\
&\le 2b\delta^\ell+e\varepsilon.
\label{eq:app_logit_bound}
\end{align}
To bound the normalized coefficients, let $\mathbf{x}$ and $\mathbf{x}'$ be logit vectors of equal dimension, and let $\mathbf{p}=\operatorname{softmax}(\mathbf{x})$ have entries $p_j=\exp(x_j)/\sum_m\exp(x_m)$. Its Jacobian (derivative matrix) with respect to $\mathbf{x}$ is $\mathbf{J}=\operatorname{diag}(\mathbf{p})-\mathbf{p}\mathbf{p}^\top$, where $\operatorname{diag}(\mathbf{p})$ has diagonal entries $p_j$ and zeros elsewhere. For any same-dimensional perturbation vector $\mathbf{v}$ with entries $v_j$,
\begin{align}
\|\mathbf{J}\mathbf{v}\|_1
&=\sum_j p_j\left|v_j-\sum_m p_m v_m\right|\notag\\
&\le 2\|\mathbf{v}\|_\infty\sum_jp_j
=2\|\mathbf{v}\|_\infty.
\end{align}
Integrating the Jacobian along the segment between two logit vectors yields the dimension-independent bound
\begin{equation}
\|\operatorname{softmax}(\mathbf{x})-
\operatorname{softmax}(\mathbf{x}')\|_1
\le 2\|\mathbf{x}-\mathbf{x}'\|_\infty.
\end{equation}
Applying this to the matched neighborhoods and using Eq.~\eqref{eq:app_logit_bound} gives
\begin{equation}
\sum_{j\in\mathcal{N}(i)}
|\alpha_{ij}^{\ell,k}-\alpha_{ij}'^{\ell,k}|
\le 4b\delta^\ell+2e\varepsilon.
\label{eq:app_attention_bound}
\end{equation}
Importantly, this bound does not grow with the number of neighbors.

\subsection{Step 4: Multi-Head Aggregation}
Let $\mathbf{m}_i^{\ell,k}=\sum_j\alpha_{ij}^{\ell,k}\mathbf{h}_j^\ell$ denote the attention-weighted message before projection and activation. Adding and subtracting $\sum_j\alpha_{ij}^{\ell,k}\mathbf{h}_j'^{\ell}$ separates state perturbations from coefficient perturbations:
\begin{align}
\mathbf{m}_i^{\ell,k}-\mathbf{m}_i'^{\ell,k}
&=\sum_j\alpha_{ij}^{\ell,k}(\mathbf{h}_j^\ell-\mathbf{h}_j'^{\ell})\notag\\
&\quad+\sum_j(\alpha_{ij}^{\ell,k}-\alpha_{ij}'^{\ell,k})
\mathbf{h}_j'^{\ell}.
\end{align}
All sums here run over $\mathcal{N}(i)$. Taking norms and applying Eqs.~\eqref{eq:app_hidden_bound} and~\eqref{eq:app_attention_bound},
\begin{align}
\|\mathbf{m}_i^{\ell,k}-\mathbf{m}_i'^{\ell,k}\|_2
&\le \delta^\ell\sum_j\alpha_{ij}^{\ell,k}
+H\sum_j|\alpha_{ij}^{\ell,k}-\alpha_{ij}'^{\ell,k}|\notag\\
&\le (1+4Hb)\delta^\ell+2He\varepsilon.
\end{align}
The linear projection and ReLU then give
\begin{equation}
\|\mathbf{z}_i^{\ell,k}-\mathbf{z}_i'^{\ell,k}\|_2
\le B_W\bigl((1+4Hb)\delta^\ell+2He\varepsilon\bigr).
\end{equation}
Finally, concatenating $K$ heads combines their squared Euclidean differences:
\begin{align}
\|\mathbf{h}_i^{\ell+1}-\mathbf{h}_i'^{\ell+1}\|_2^2
&=\sum_{k=1}^{K}\|\mathbf{z}_i^{\ell,k}-\mathbf{z}_i'^{\ell,k}\|_2^2.
\end{align}
Let $q$ denote the per-layer amplification bound for hidden-state differences and $r$ the per-layer sensitivity bound for edge-weight differences. Taking the maximum over nodes gives the recurrence used in the main paper, with these constants defined explicitly as
\begin{gather}
\delta^{\ell+1}\le q\delta^\ell+r\varepsilon,\\
q=\sqrt{K}B_W(1+4Hb),\qquad
r=\sqrt{K}B_W\,2He.
\label{eq:app_recurrence}
\end{gather}
Thus $q$ bounds the propagation of an existing hidden-state perturbation, while $r$ bounds the additional effect of perturbed edge weights at each layer.

\subsection{Step 5: Layer-Wise Induction}
Using $t$ as a dummy integer summation index over powers of $q$, we show by induction that, for every layer $1\le\ell\le D$,
\begin{equation}
\delta^\ell\le
\left(q^\ell+r\sum_{t=0}^{\ell-1}q^t\right)\varepsilon.
\end{equation}
For $\ell=1$, Eq.~\eqref{eq:app_recurrence} and $\delta^0\le\varepsilon$ give $\delta^1\le(q+r)\varepsilon$.
Assuming the claim at layer $\ell$, the next layer satisfies
\begin{align}
\delta^{\ell+1}
&\le q\left(q^\ell+r\sum_{t=0}^{\ell-1}q^t\right)\varepsilon
+r\varepsilon\notag\\
&=\left(q^{\ell+1}+r\sum_{t=0}^{\ell}q^t\right)\varepsilon.
\end{align}
Consequently, after $D$ layers,
\begin{equation}
\delta^D\le\left(q^D+r\sum_{t=0}^{D-1}q^t\right)\varepsilon.
\label{eq:app_depth_bound}
\end{equation}
The geometric sum equals $(q^D-1)/(q-1)$ when $q\ne1$, and $D$ when $q=1$. No assumption that $q<1$ is required for a finite-depth Lipschitz bound.

\subsection{Step 6: Pooling and Score Bound}
Let $n=|\mathcal{N}|$ be the number of nodes and $\mathbf{g}=n^{-1}\sum_{i\in\mathcal{N}}\mathbf{h}_i^D$ the graph-level representation obtained by global mean pooling. Let $\mathbf{g}'$ denote the corresponding representation for $c'$. The matched node sets imply
\begin{equation}
\|\mathbf{g}-\mathbf{g}'\|_2
\le\frac{1}{n}\sum_i\|\mathbf{h}_i^D-\mathbf{h}_i'^D\|_2
\le\delta^D.
\end{equation}
Thus mean pooling introduces no multiplicative dependence on $n$.
Let $\mathbf{R}_1$ and $\mathbf{R}_2$ be the first and second weight matrices of the two-layer MLP score head, $\mathbf{a}_1$ its hidden-layer bias vector, and $a_2$ its scalar output bias. The function $\operatorname{sigmoid}(x)=1/(1+\exp(-x))$ maps a scalar logit to $[0,1]$. For a configuration with pooled representation $\mathbf{g}$, write its predicted score $\hat{s}$ as
\begin{equation}
\hat{s}=\operatorname{sigmoid}\!\left(
\mathbf{R}_2\operatorname{ReLU}(\mathbf{R}_1\mathbf{g}+\mathbf{a}_1)+a_2\right).
\end{equation}
Let $L_{\mathrm{MLP}}$ denote a Lipschitz constant of the complete score head, including the final sigmoid. Since ReLU is $1$-Lipschitz and the sigmoid derivative is at most $1/4$, a valid choice satisfies
\begin{equation}
L_{\mathrm{MLP}}\le\tfrac14\|\mathbf{R}_2\|_2\|\mathbf{R}_1\|_2,
\end{equation}
where $L_{\mathrm{MLP}}$ includes the final sigmoid. Combining this with Eq.~\eqref{eq:app_depth_bound} yields
\begin{gather}
|\hat{s}(c)-\hat{s}(c')|\le L\varepsilon,\\
L=L_{\mathrm{MLP}}\left(q^D+r\sum_{t=0}^{D-1}q^t\right).
\end{gather}
This establishes the main-paper bound. Its constant depends on network depth, head count, and parameter norms, but not on component identities or the number of matched nodes.

\subsection{Extension to Variable-Size Configurations}
We extend the smoothness result to configurations with different numbers of action-macro nodes or different node sets. By representing configurations on a common padded graph and assigning bounded embeddings to inserted or deleted nodes, the GNN predictor remains Lipschitz-continuous with respect to both component perturbations and bounded structural changes.

Let $\mathcal{M}\subseteq\mathcal{N}\cap\mathcal{N}'$ be a matching between nodes of the same modality, with $m=|\mathcal{M}|$. Let $n=|\mathcal{N}|$ and $n'=|\mathcal{N}'|$ denote the two node counts. The nodes in $\mathcal{N}\setminus\mathcal{M}$ and $\mathcal{N}'\setminus\mathcal{M}$ represent inserted or deleted components. We place both configurations on the common union node set $\mathcal{N}\cup\mathcal{N}'$, using a bounded null embedding and a mask for absent nodes and edges. This common representation allows the same GNN to score configurations with different node counts without requiring a fixed component catalog.

For matched nodes, assume $\|\mathbf{e}_i-\mathbf{e}'_i\|_2\le\varepsilon$. The layer-wise argument above therefore remains valid:
\begin{equation}
\delta^D\le
\left(q^D+r\sum_{t=0}^{D-1}q^t\right)\varepsilon.
\label{eq:app_variable_depth_bound}
\end{equation}
The effect of unmatched nodes is bounded by the hidden-state bound $H$. Define
\begin{equation}
\rho=\frac{n-m}{n}+\frac{n'-m}{n'}+m\left|\frac{1}{n}-\frac{1}{n'}\right|.
\end{equation}
For global mean pooling, the matched nodes contribute the first term below, while unmatched nodes and the different normalization factors contribute the second:
\begin{equation}
\|\mathbf{g}-\mathbf{g}'\|_2
\le \frac{m}{\max\{n,n'\}}\delta^D+H\rho.
\label{eq:app_variable_pool_bound}
\end{equation}
Combining Eqs.~\eqref{eq:app_variable_depth_bound} and~\eqref{eq:app_variable_pool_bound} with the Lipschitz score head gives
\begin{equation}
\resizebox{0.82\columnwidth}{!}{$\displaystyle
|\hat{s}(c)-\hat{s}(c')|
\le L_{\mathrm{MLP}}\left[\frac{m}{\max\{n,n'\}}
\left(q^D+r\sum_{t=0}^{D-1}q^t\right)\varepsilon+H\rho\right]$}.
\label{eq:app_variable_size_bound}
\end{equation}
This bound establishes smoothness for configurations that differ in both component semantics and node cardinality. The first term captures changes among matched components; the second gives a bounded contribution for inserted or deleted nodes. When the node counts agree and all nodes are matched, $\rho=0$, and Eq.~\eqref{eq:app_variable_size_bound} reduces exactly to the original Lipschitz bound.

When node insertion or deletion also changes neighboring edges, let $\eta_{\mathrm{graph}}$ measure the resulting bounded graph-structure perturbation. The layer recurrence becomes
\begin{equation}
\delta^{\ell+1}\le q\delta^\ell+r\varepsilon+s\eta_{\mathrm{graph}},
\end{equation}
where $s$ depends on the affected neighborhoods and the hidden-state bound. Unrolling the same induction adds a finite term proportional to $\eta_{\mathrm{graph}}$ to Eq.~\eqref{eq:app_variable_size_bound}. Thus, the predictor remains smooth for variable-size configurations with bounded node and graph changes, including newly generated macro-action sets.

This extension provides a theoretical basis for GraphEvo's open-ended configuration search: newly generated configurations do not need to share a fixed number of macros or a predefined component catalog. Once their components and induced graph changes are represented in the common graph space, the surrogate can score them smoothly and the selector can compare them before expensive emulator rollouts.

\subsection{Implications for Surrogate-Guided Search}
The two bounds establish the stability required for GraphEvo's surrogate-based optimization. For configurations with the same node structure, nearby component embeddings lead to nearby predicted scores; for configurations with different macro counts or node sets, the score remains smoothly controlled by the semantic perturbation, the bounded node-set change, and the induced graph edit. Thus, the predictor can score newly generated configurations without a fixed component catalog or a fixed number of macros.

This result directly supports GraphEvo's open-ended search procedure. The evolver may generate new strategies, visual pipelines, or macro-action sets, while the padded graph representation and bounded graph-edit term keep their surrogate scores comparable to previously evaluated configurations. The selector can therefore rank and screen variable-size candidates before expensive emulator rollouts, converting the limited evaluation budget into a more effective search over unseen configurations.

The same conclusion applies to MC-dropout scoring when the same sampled masks are used for both configurations: each sub-network and their predictive mean preserve the corresponding smoothness bound. Together, the fixed-size and variable-size results provide a theoretical basis for GraphEvo's graph-guided surrogate optimization across an open configuration space.
\begin{algorithm*}[t]
\caption{Signature discovery and scanner-based task verification}
\label{alg:memory_evaluation}
\small
\begin{algorithmic}[1]
\Require Reference memory $\mathcal{M}_{\mathrm{ref}}$; map string $m$; reference coordinates $\mathbf{x}_{\mathrm{ref}}$;
\Statex \hspace{\algorithmicindent} coordinate offset $\Delta_{\mathrm{coord}}$; task-state layout $\mathcal{L}$; success predicate $P$
\Ensure Signature $S$, variable offsets $\boldsymbol{\Delta}$, and a task-completion decision
\Statex \textbf{Stage 1: Signature discovery (discovery script)}
\State Scan map-string occurrences: $\mathcal{A}\gets\operatorname{Scan}(\mathcal{M}_{\mathrm{ref}},m)$
\For{each candidate address $a\in\mathcal{A}$}
    \State Read coordinates: $\mathbf{x}\gets\operatorname{Read}_{\mathrm{float}}(\mathcal{M}_{\mathrm{ref}},a+\Delta_{\mathrm{coord}})$
    \State Discard $a$ if $\mathbf{x}$ does not match $\mathbf{x}_{\mathrm{ref}}$
\EndFor
\State Validate remaining addresses against the known task state using $\mathcal{L}$
\If{no unique valid address remains} \State \Return $\bot$ (discovery failure) \EndIf
\State Let $a^*$ be the validated address
\State Extract the surrounding signature and variable offsets:
\Statex \hspace{2\algorithmicindent} $(S,\boldsymbol{\Delta})\gets\operatorname{Extract}(\mathcal{M}_{\mathrm{ref}},a^*,\mathcal{L})$
\Statex \textbf{Stage 2: Runtime scanner: state localization and verification}
\State Scan signature occurrences: $\mathcal{B}\gets\operatorname{Scan}(\mathcal{M},S)$
\State Validate candidates using the map and task-state relations $(m,\mathcal{L},\boldsymbol{\Delta})$
\If{no unique valid address remains} \State \Return $\bot$ (localization failure) \EndIf
\State Let $b^*$ be the validated signature address
\State Resolve task-variable addresses: $\mathbf{a}\gets b^*+\boldsymbol{\Delta}$
\While{episode active}
    \State Read current task state: $\mathbf{z}\gets\operatorname{Read}(\mathcal{M},\mathbf{a})$
    \If{$P(\mathbf{z})$ is true} \State \Return $\mathrm{success}$ \EndIf
\EndWhile
\State \Return $\mathrm{not\ completed}$
\end{algorithmic}
\end{algorithm*}

\section{Reproducibility and Release}
\label{app:reproducibility}

To facilitate reproducibility while respecting intellectual property, we will release all non-proprietary components of PokeGym as an open-source package.

\subsection{Released Artifacts}
We will release the following non-proprietary components as an open-source package: the Ryujinx emulator wrapper that interfaces with the game, signature discovery scripts, the runtime memory scanner for task-state address localization, the automatic evaluator for success-condition verification, task definitions (descriptions, criteria, and budgets), the complete GraphEvo implementation (including evolution, predictor training, and selector logic), and documentation for environment setup and benchmark execution. Proprietary game assets, including the ROM, firmware, and decryption keys, are not redistributed and must be legally obtained by researchers.

\subsection{Signature Discovery and State Localization}
\label{app:signature}
Algorithm~\ref{alg:memory_evaluation} summarizes two stages: the discovery script extracts a signature, and the runtime scanner uses it to locate task-state variables and verify task completion.

\textbf{Signature discovery.} Reverse engineering of open-source game plugins identifies the in-memory representation of the map string and the relative offset $\Delta_{\mathrm{coord}}$ from its address to the player's coordinate fields. The discovery script scans reference memory for all occurrences of the map string. For each occurrence at address $a$, it reads the coordinate fields as single-precision floats at $a+\Delta_{\mathrm{coord}}$ and compares the resulting coordinate vector with the known coordinates of the reference state. This offset-based read distinguishes the correct map-string occurrence from other matches. From the validated location, the script extracts a surrounding memory signature and records relative offsets to the task-state variables used by the runtime scanner.

\textbf{Runtime scanning and task verification.} At the start of an episode, the scanner searches current emulator memory for the stored signature and applies its relative offsets to locate task-state variables. It validates the addresses against the expected map and task-state relations, then reads the located values and applies the task-specific success predicate. The resulting task outcome is used as evaluation feedback; raw memory values are not exposed to the agent, evolver, or predictor.

\textbf{Algorithm notation.} $\mathcal{M}_{\mathrm{ref}}$ and $\mathcal{M}$ denote reference and live emulator memory, respectively; $\mathcal{L}$ specifies the task-state layout and validation relations. The map string is $m$, the reference coordinates are $\mathbf{x}_{\mathrm{ref}}$, and $\Delta_{\mathrm{coord}}$ is their offset from a map-string address. Sets $\mathcal{A}$ and $\mathcal{B}$ contain candidate map-string and signature addresses; $a^*$ and $b^*$ are their unique validated elements. The extracted signature is $S$, $\boldsymbol{\Delta}$ contains the task-variable offsets from its location, $\mathbf{a}$ is the resolved address vector, $\mathbf{z}$ contains current state values, and $P$ is the Boolean success predicate. The operator $\operatorname{Scan}$ returns pattern matches, and $\operatorname{Read}_{\mathrm{float}}$ reads a candidate's single-precision coordinate vector $\mathbf{x}$. The operators $\operatorname{Extract}$ and $\operatorname{Read}$ derive the signature and offsets and read task-state values, respectively. The symbol $\bot$ denotes localization or discovery failure.


\subsection{Reproducibility Across Platforms and Versions}
\label{app:reproducibility_validation}

\textbf{Scanner reliability.} Table~2 in the main paper measures whether the scanner locates the correct task-state variables on its first attempt. It does so in all 100 runs for each of five platform--emulator--game configurations, covering Windows~10/11, Ubuntu~22.04, emulator v1.3.2/1.3.3, and game v1.0.0/2.0.2. This establishes address-localization reliability across the tested setups; after localization, the scanner uses the resolved state variables to check each task's success condition.

\textbf{Reproduction and release.} We make every effort to maximize reproducibility: researchers can choose any compatible game-ROM version, emulator version, and platform, and reproduce the benchmark using our released components (emulator wrapper, task-specific starting states and definitions, signature-discovery script, runtime scanner, GraphEvo implementation, and setup instructions). Since signatures are derived from relative task-state relationships rather than fixed addresses, they can be rediscovered for different compatible versions.

\section{Experimental Details}
\label{app:experimental_details}

\subsection{ALFRED Benchmark}
\label{app:alfred_dataset}

ALFRED~\cite{alfred} is an instruction-following benchmark for household tasks in the AI2-THOR simulator~\cite{ai2thor}.
Agents must navigate indoor scenes, locate and manipulate objects, and complete the sequence of actions specified by a natural-language instruction.
Compared with PokeGym's urban navigation and narrative interactions, ALFRED emphasizes object-centric manipulation and reasoning about object states and spatial relations.

Our evaluation uses 174 task instances from the validation-unseen split, covering six of ALFRED's seven task types and excluding simple pick-and-place tasks.
We group these types into three categories:
\textbf{State-Changing} comprises cleaning, heating, and cooling an object before placing it;
\textbf{Spatial} comprises placing two objects and placing an object using a movable receptacle;
and \textbf{Inspection} requires locating an object and examining it under a light source.

We report \textbf{Goal-Condition Success (GCS)}, the fraction of task goal conditions satisfied, averaged over evaluated episodes, and \textbf{Path-Weighted Goal-Condition Success (PW-GCS)}, which additionally weights goal completion by the ratio of the reference path length to the executed path length, capped at one.
Together, these metrics capture partial-task completion and execution efficiency.

\subsection{Implementation Details}
\label{app:implementation_details}

\subsubsection{Visual Key-Frame Selection.}

For each task, we collect a small set of reference episodes and construct a bank of frames covering representative states, such as the starting location, target area, and scenes with occlusion or moving NPCs.
Before candidate evaluation, we manually select one step that captures the task's main visual challenge and fix its four observation views (previous, front, left, and right) for surrogate scoring.
Each candidate's visual-enhancement pipeline is applied to these same views to construct its visual graph nodes.
This shared visual reference makes candidate representations comparable and reproducible without requiring additional emulator rollouts for surrogate scoring; actual agent rollouts still use live observations.

\subsubsection{Candidate Generation and Evaluation Budget.}

GraphEvo generates $M=3$ candidate configurations per evolution round. The GNN surrogate scores all candidates, and the disagreement-aware selector chooses one for an emulator rollout; its trajectory and measured score then support configuration updates and online GNN training.
Candidate counts for the baselines follow their respective search mechanisms: EvoPrompt and PromptBreeder generate 6 candidate prompts per round, whereas AgentSquare and EvoTest generate 3 candidate configurations, matching GraphEvo's candidate count. Their V+L+A extensions retain the same count.
The zero-shot VLMs, Cradle, Voyager, and Reflexion do not maintain a candidate pool for selection; each attempt directly executes a single configuration.
The adaptation protocol consists of 10 iterations, with each episode capped at 360 environment steps. All methods start each task from the same save state and receive only RGB observations during execution, without access to internal game states. Each experimental setting is evaluated over 5 independent trials, and SR and IMR are averaged across trials.

%

\subsubsection{Warm-Up Training Set for the GNN Predictor.}

The warm-up set addresses the surrogate's cold-start problem: before observing any evaluated configurations, the GNN has no task-specific supervision for ranking candidates, and its predictions and disagreement estimates are not yet grounded in rollout outcomes.
Initial graph--score pairs provide this supervision before the selector begins using the surrogate to guide search.
Before evolutionary search, we train the GNN surrogate on a small warm-up set of configurations proposed by the Multimodal Configuration Evolver.
Each configuration is executed in the emulator to obtain a measured score, yielding a training pair consisting of its graph representation and score.
During search, newly evaluated configurations supply additional training pairs for online updates, allowing the surrogate to adapt as the configuration space expands.

\begin{table*}[t]
\centering
\caption{Overview of baseline adaptations in PokeGym. V/L/A denote visual, language, and action components. $\dagger$ adds visual evolution while retaining the baseline's optimization mechanism. ``---'' indicates no separate population-level selector.}
\label{tab:baselines}
\small
\setlength{\tabcolsep}{4pt}
\renewcommand{\arraystretch}{1.15}

\begin{tabularx}{\textwidth}{@{}l c >{\raggedright\arraybackslash}X >{\raggedright\arraybackslash}X l@{}}
\toprule
\textbf{Method} & \textbf{Modalities} & \textbf{Evolving components} & \textbf{Update mechanism} & \textbf{Selection} \\
\midrule
\multicolumn{5}{@{}l}{\hspace{3pt}\textcolor{black!75}{\textbf{Game agent frameworks}}} \\
\addlinespace[3pt]
Cradle & L+A & Reflections + skills & Action-level reflection; episode-level skill curation & --- \\
Voyager & A & Skill library & Visual-feedback refinement & --- \\
\midrule
\multicolumn{5}{@{}l}{\hspace{3pt}\textcolor{black!75}{\textbf{Self-evolving methods}}} \\
\addlinespace[3pt]
Reflexion & L & Strategy text & Episode-level self-reflection & --- \\
EvoPrompt & L & System instructions & LLM crossover + mutation & Score-proportional \\
PromptBreeder & L & Instructions + mutation prompts & Prompt-guided mutation & Binary tournament \\
AgentSquare & L+A & Strategies + macros & Module evolution + recombination & LLM-based \\
EvoTest & L+A & Strategies + macros & Episode-level configuration revision & Bandit \\
\midrule
\multicolumn{5}{@{}l}{\hspace{3pt}\textcolor{black!75}{\textbf{Extensions to all three modalities ($\dagger$)}}} \\
\addlinespace[3pt]
AgentSquare$^\dagger$ & V+L+A & Strategies + macros + visual pipeline & Same as AgentSquare & LLM-based \\
EvoTest$^\dagger$ & V+L+A & Strategies + macros + visual pipeline & Same as EvoTest & Bandit \\
\bottomrule
\end{tabularx}
\end{table*}

\subsection{Baselines}
\label{app:baselines}
Except for the zero-shot VLMs, the evolutionary process is run for 10 iterations.
Below we provide detailed descriptions of the baselines compared in the main experiments, covering the modalities each evolves, its original evolution mechanism, and how it is adapted to PokeGym. Table~\ref{tab:baselines} summarizes these properties at a glance.

\subsubsection{Game Agent Frameworks.}
These frameworks were originally designed for general video-game playing and are adapted to PokeGym.

\textbf{Cradle} \cite{cradle} evolves the \emph{language} and \emph{action} modalities through self-reflection and skill curation. Designed for general computer control, it takes screenshots as input and interacts through keyboard and mouse commands. Its Self-Reflection module reviews executed actions and stores diagnostic notes in episodic memory to guide subsequent decisions. Its Skill Curation module generates executable skills from in-game tutorials and manuals, revises incorrect skills, and maintains a growing library in procedural memory for retrieval and reuse across tasks.

%
%
Adapted to PokeGym, Cradle uses the benchmark's visual observations and emulator controls. At each decision step, it identifies sub-tasks (Task Inference), interprets the current observations (Information Gathering), and produces a three-action plan (Action Planning). After each executed action, Self-Reflection compares the previous and current frames to assess its effect and diagnose failures before the next decision. Skill Curation operates across episodes: the accumulated trajectories are analyzed after each episode to update a macro-action library, which supplies reusable skills for subsequent planning.

\textbf{Voyager} \cite{voyager} evolves the \emph{action} modality by acquiring reusable executable skills. Originally developed for Minecraft, it combines an automatic curriculum that proposes progressively harder tasks with an iterative prompting loop that refines skill programs using execution feedback and error messages. Successful programs are stored in a skill library and indexed by semantic embeddings for later retrieval and composition.

%
Adapted to PokeGym, the benchmark's predefined tasks replace the automatic curriculum. Iterative refinement uses visual feedback: before each decision, the model compares consecutive frames to assess the previous action and adjust its next plan when necessary. The skill library is represented by a macro-action library, with each skill mapped to an executable sequence of controller inputs.

\subsubsection{Self-evolving Methods.}
These methods adapt the agent at test time by evolving prompts, skills or configurations.

\textbf{Reflexion} \cite{reflexion} evolves the \emph{language} modality by accumulating verbal reflections across episodes. In the original framework, an Actor executes the task, an Evaluator scores the resulting trajectory, and a Self-Reflection model uses this feedback to identify mistakes and formulate lessons for future attempts. These reflections are stored in episodic memory and provided to the Actor in subsequent episodes, enabling adaptation without updating model weights.

Adapted to PokeGym, verbal reflections are represented as a strategy text that guides the Actor's subsequent attempts. After each episode, the Evolver performs Self-Reflection by analyzing the trajectory and the binary task outcome supplied by the benchmark, then updates the strategy with the resulting lessons. As with the other methods, task outcomes are provided only to the Evolver for configuration updates and are not directly exposed to the Actor. Visual processing and the macro-action library remain unchanged.

\textbf{EvoPrompt} \cite{evoprompt} evolves the \emph{language} modality by applying evolutionary search to a population of candidate prompts. An LLM performs crossover and mutation on selected parent prompts to generate offspring. Candidates are evaluated on the target task, and better-performing prompts are retained for the next generation, enabling prompt optimization without updating model weights.

Adapted to PokeGym, each candidate is a system instruction that guides the agent's reasoning and action generation, initialized from a prompt template. At each generation, candidates are scored through rollouts on the current task, and two parents are selected with probabilities proportional to their scores. An LLM then applies crossover and mutation to produce offspring instructions for subsequent evaluation. Only the system instruction evolves; the visual processing and macro-action library remain unchanged.

\textbf{PromptBreeder} \cite{promptbreeder} evolves the \emph{language} modality by jointly optimizing task prompts and mutation prompts. Each individual contains a task prompt that determines the agent's behavior and a mutation prompt that guides its revision. Binary tournaments select the fitter of two sampled individuals for mutation, while the mutation prompts themselves also evolve, allowing the method to improve both the instructions and the rules used to revise them.

Adapted to PokeGym, a separate population of candidate instructions is maintained for each task, with each candidate paired with a thinking-style prompt and a mutation prompt. At each round, candidates are evaluated through task rollouts and randomly paired for binary tournaments based on their scores. An LLM revises each winning instruction using its associated mutation prompt, and the winners and their offspring form the population for the next round.

\textbf{AgentSquare} \cite{agentsquare} evolves the \emph{language} and \emph{action} modalities by searching over modular agent configurations. The original framework organizes each agent into four module categories: planning, reasoning, memory, and tool use. Search alternates between \emph{module evolution}, where an LLM generates new module implementations, and \emph{module recombination}, where existing modules are assembled into new configurations. Configurations containing newly generated modules are evaluated directly in the environment, as these modules are absent from the experience pool used by the predictor. Recombinations of existing modules are first screened by an LLM performance predictor, reserving rollouts for promising candidates.

Adapted to PokeGym, planning, reasoning, and memory are consolidated into a strategy text, while tool use is represented by a macro-action library. We retain the two-phase search over these components, starting from randomly generated candidate libraries. During module evolution, an LLM uses previously evaluated trajectories to propose revised strategies and macro-action sets; configurations containing these new components are evaluated directly in the emulator. During module recombination, an LLM selects one strategy and one macro-action library from the evaluated pool based on the task description and accumulated experience, and the resulting configuration is executed in the emulator. The visual modality is left unchanged.

\textbf{EvoTest} \cite{evotest} jointly evolves the \emph{language} and \emph{action} modalities through episode-level configuration updates. The original framework represents an agent configuration as a policy prompt, memory, hyperparameters, and tool-use routines. An Actor executes an episode with the selected configuration, after which an Evolver analyzes the complete trajectory to propose revisions. A bandit-style selector uses empirical performance and evaluation counts to balance exploration and exploitation when choosing the configuration for the next episode.

Adapted to PokeGym, the policy prompt, memory, and hyperparameter settings are consolidated into a strategy text, while tool-use routines are represented by a macro-action library. After each episode, the Evolver uses the trajectory and task outcome to revise both components. The bandit selector determines which configuration the Actor executes in the next emulator rollout, preserving the original exploration--exploitation mechanism. The visual modality is left unchanged.

\textbf{Extensions to V+L+A.} We additionally evaluate AgentSquare$^\dagger$ and EvoTest$^\dagger$, which extend the above adaptations to evolve a visual-enhancement pipeline alongside the strategy text and macro-action library. The visual component uses the same image-processing operations as GraphEvo and is updated from trajectory feedback within each baseline's configuration search. AgentSquare$^\dagger$ includes this component in module evolution and recombination, while EvoTest$^\dagger$ includes it in the Evolver's configuration updates and retains bandit selection. These variants use the same rollout budget as GraphEvo, allowing comparison at matched modality coverage and evaluation budget.

\begin{table*}[htbp]
\centering
\renewcommand{\arraystretch}{1.0}
\small
\setlength{\tabcolsep}{2pt}
\caption{Comparison with test-time parameter adaptation. V-GPS updates a value function's parameters at test time and selects candidate actions by learned value.}
\label{tab:ttl_baseline}
\resizebox{0.90\linewidth}{!}{%
\newcommand{\mse}[2]{#1{\tiny$\pm$#2}}
\begin{tabular*}{0.9\textwidth}{@{\extracolsep{\fill}}l c@{\hspace{2pt}}c c@{\hspace{2pt}}c c@{\hspace{2pt}}c c@{\hspace{2pt}}c@{}}
\toprule
\multirow{2}{*}{\textbf{Method}} & \multicolumn{2}{c}{\textbf{Navigation}} & \multicolumn{2}{c}{\textbf{Interaction}} & \multicolumn{2}{c}{\textbf{Localization}} & \multicolumn{2}{c}{\textbf{Average}} \\
\cmidrule(lr){2-3} \cmidrule(lr){4-5} \cmidrule(lr){6-7} \cmidrule(lr){8-9}
 & \textbf{SR}$\uparrow$ & \textbf{IMR}$\downarrow$ & \textbf{SR}$\uparrow$ & \textbf{IMR}$\downarrow$ & \textbf{SR}$\uparrow$ & \textbf{IMR}$\downarrow$ & \textbf{SR}$\uparrow$ & \textbf{IMR}$\downarrow$ \\
\midrule
V-GPS & \mse{42.78}{5.92} & \mse{12.99}{1.92} & \mse{54.67}{4.98} & \mse{7.32}{0.92} & \mse{49.09}{4.09} & \mse{13.65}{2.47} & \mse{49.30}{2.97} & \mse{10.94}{1.03} \\
\textbf{GraphEvo (Ours)} & \textbf{\mse{46.11}{6.11}} & \textbf{\mse{5.72}{1.01}} & \textbf{\mse{68.00}{3.83}} & \textbf{\mse{5.76}{0.89}} & \textbf{\mse{64.85}{4.27}} & \textbf{\mse{5.07}{0.96}} & \textbf{\mse{60.18}{2.87}} & \textbf{\mse{5.55}{0.54}} \\
\bottomrule
\end{tabular*}%
} 
\end{table*}

\begin{table*}[t]
\centering
\caption{Zero-shot and GraphEvo performance across backbone VLMs on PokeGym. All SR and IMR values are percentages, with standard errors in smaller font. Bold indicates the better mean within each backbone pair.}
\label{tab:new_backbones}
\footnotesize
\setlength{\tabcolsep}{3pt}
\renewcommand{\arraystretch}{1.05}
\newcommand{\backbonese}[2]{#1{\tiny$\pm$#2}}
\resizebox{\textwidth}{!}{%
\begin{tabular}{llcccccccc}
\toprule
\multirow{2}{*}{\textbf{Backbone}} & \multirow{2}{*}{\textbf{Method}} & \multicolumn{2}{c}{\textbf{Navigation}} & \multicolumn{2}{c}{\textbf{Interaction}} & \multicolumn{2}{c}{\textbf{Localization}} & \multicolumn{2}{c}{\textbf{Average}} \\
\cmidrule(lr){3-4}\cmidrule(lr){5-6}\cmidrule(lr){7-8}\cmidrule(lr){9-10}
& & \textbf{SR}$\uparrow$ & \textbf{IMR}$\downarrow$ & \textbf{SR}$\uparrow$ & \textbf{IMR}$\downarrow$ & \textbf{SR}$\uparrow$ & \textbf{IMR}$\downarrow$ & \textbf{SR}$\uparrow$ & \textbf{IMR}$\downarrow$ \\
\midrule
\multirow{2}{*}{Qwen3.6-35B} & Zero-shot & \backbonese{18.89}{4.43} & \backbonese{42.25}{3.45} & \backbonese{14.67}{3.20} & \backbonese{23.38}{2.50} & \backbonese{20.00}{3.59} & \backbonese{29.94}{3.62} & \backbonese{17.54}{2.15} & \backbonese{31.24}{1.94} \\
 & GraphEvo & \backbonese{\textbf{31.11}}{4.88} & \backbonese{\textbf{17.96}}{2.17} & \backbonese{\textbf{32.44}}{4.13} & \backbonese{\textbf{13.70}}{1.47} & \backbonese{\textbf{36.36}}{4.22} & \backbonese{\textbf{15.64}}{2.35} & \backbonese{\textbf{33.16}}{2.54} & \backbonese{\textbf{15.61}}{1.13} \\
\midrule
\multirow{2}{*}{GPT-5.4} & Zero-shot & \backbonese{31.11}{5.88} & \backbonese{17.93}{2.59} & \backbonese{38.22}{4.66} & \backbonese{8.74}{1.34} & \backbonese{42.42}{4.06} & \backbonese{15.05}{2.87} & \backbonese{37.19}{2.87} & \backbonese{13.47}{1.32} \\
 & GraphEvo & \backbonese{\textbf{46.11}}{6.11} & \backbonese{\textbf{5.72}}{1.01} & \backbonese{\textbf{68.00}}{3.83} & \backbonese{\textbf{5.76}}{0.89} & \backbonese{\textbf{64.85}}{4.27} & \backbonese{\textbf{5.07}}{0.96} & \backbonese{\textbf{60.18}}{2.87} & \backbonese{\textbf{5.55}}{0.54} \\
\midrule
\multirow{2}{*}{GPT-6.1-sol} & Zero-shot & \backbonese{56.11}{6.27} & \backbonese{13.62}{1.89} & \backbonese{62.67}{4.31} & \backbonese{7.11}{1.33} & \backbonese{62.42}{5.35} & \backbonese{13.95}{2.53} & \backbonese{60.53}{3.03} & \backbonese{11.15}{1.11} \\
 & GraphEvo & \backbonese{\textbf{74.44}}{5.44} & \backbonese{\textbf{6.39}}{1.55} & \backbonese{\textbf{79.11}}{3.36} & \backbonese{\textbf{4.48}}{0.94} & \backbonese{\textbf{82.42}}{3.23} & \backbonese{\textbf{7.00}}{2.35} & \backbonese{\textbf{78.60}}{2.36} & \backbonese{\textbf{5.81}}{0.91} \\
\midrule
\multirow{2}{*}{Claude-Sonnet-5-5} & Zero-shot & \backbonese{43.33}{6.40} & \backbonese{22.10}{2.99} & \backbonese{54.67}{5.02} & \backbonese{9.21}{2.02} & \backbonese{35.76}{5.69} & \backbonese{21.95}{3.98} & \backbonese{45.61}{3.33} & \backbonese{16.97}{1.77} \\
 & GraphEvo & \backbonese{\textbf{65.00}}{5.66} & \backbonese{\textbf{10.66}}{1.46} & \backbonese{\textbf{75.56}}{3.81} & \backbonese{\textbf{7.76}}{1.43} & \backbonese{\textbf{55.15}}{5.84} & \backbonese{\textbf{12.05}}{2.58} & \backbonese{\textbf{66.32}}{2.96} & \backbonese{\textbf{9.92}}{1.05} \\
\bottomrule
\end{tabular}%
}
\end{table*}

\begin{table*}[t]
\centering
\small
\setlength{\tabcolsep}{4pt}
\caption{Complete modality ablation study. The first three columns indicate which modalities (Visual, Strategy, Action) are included in each configuration.}
\label{tab:complete_modal_ablation}
\begin{tabular}{ccc cc cc cc cc}
\toprule
\multicolumn{3}{c}{\textbf{Modality}} & \multicolumn{2}{c}{\textbf{Navigation}} & \multicolumn{2}{c}{\textbf{Interaction}} & \multicolumn{2}{c}{\textbf{Localization}} & \multicolumn{2}{c}{\textbf{Average}} \\
\cmidrule(lr){1-3} \cmidrule(lr){4-5} \cmidrule(lr){6-7} \cmidrule(lr){8-9} \cmidrule(lr){10-11}
\textbf{Visual} & \textbf{Strategy} & \textbf{Action} & \textbf{SR}$\uparrow$ & \textbf{IMR}$\downarrow$ & \textbf{SR}$\uparrow$ & \textbf{IMR}$\downarrow$ & \textbf{SR}$\uparrow$ & \textbf{IMR}$\downarrow$ & \textbf{SR}$\uparrow$ & \textbf{IMR}$\downarrow$ \\
\midrule
 & \checkmark & & 42.22{\scriptsize \,$\pm$6.00} & 7.02{\scriptsize \,$\pm$1.54} & 41.33{\scriptsize \,$\pm$4.56} & 6.95{\scriptsize \,$\pm$1.52} & 49.09{\scriptsize \,$\pm$6.10} & 7.33{\scriptsize \,$\pm$1.49} & 43.86{\scriptsize \,$\pm$3.14} & 7.09{\scriptsize \,$\pm$0.88} \\
\checkmark & & & 36.67{\scriptsize \,$\pm$5.99} & 19.80{\scriptsize \,$\pm$2.39} & 42.67{\scriptsize \,$\pm$4.20} & 11.83{\scriptsize \,$\pm$1.72} & 43.64{\scriptsize \,$\pm$4.80} & 16.07{\scriptsize \,$\pm$2.89} & 41.05{\scriptsize \,$\pm$2.86} & 15.58{\scriptsize \,$\pm$1.34} \\
 & & \checkmark & 37.22{\scriptsize \,$\pm$6.02} & 8.00{\scriptsize \,$\pm$1.41} & 37.33{\scriptsize \,$\pm$4.52} & 9.66{\scriptsize \,$\pm$1.96} & 38.79{\scriptsize \,$\pm$5.43} & 8.01{\scriptsize \,$\pm$1.81} & 37.72{\scriptsize \,$\pm$3.02} & 8.66{\scriptsize \,$\pm$1.03} \\
\midrule
 & \checkmark & \checkmark & 34.44{\scriptsize \,$\pm$5.94} & 6.86{\scriptsize \,$\pm$1.19} & 42.67{\scriptsize \,$\pm$5.23} & 7.13{\scriptsize \,$\pm$1.98} & 41.82{\scriptsize \,$\pm$5.32} & 5.30{\scriptsize \,$\pm$1.01} & 39.82{\scriptsize \,$\pm$3.18} & 6.52{\scriptsize \,$\pm$0.91} \\
\checkmark & \checkmark & & 42.78{\scriptsize \,$\pm$5.75} & 8.26{\scriptsize \,$\pm$1.79} & 55.11{\scriptsize \,$\pm$4.76} & 7.63{\scriptsize \,$\pm$1.65} & 52.12{\scriptsize \,$\pm$5.83} & 7.57{\scriptsize \,$\pm$1.39} & 50.35{\scriptsize \,$\pm$3.12} & 7.81{\scriptsize \,$\pm$0.94} \\
\checkmark & & \checkmark & 39.44{\scriptsize \,$\pm$6.25} & 8.55{\scriptsize \,$\pm$1.32} & 48.00{\scriptsize \,$\pm$4.73} & 10.19{\scriptsize \,$\pm$2.19} & 53.94{\scriptsize \,$\pm$5.47} & 7.31{\scriptsize \,$\pm$1.46} & 47.02{\scriptsize \,$\pm$3.16} & 8.84{\scriptsize \,$\pm$1.05} \\
\midrule
\checkmark & \checkmark & \checkmark & \textbf{46.11{\scriptsize \,$\pm$6.11}} & \textbf{5.72{\scriptsize \,$\pm$1.01}} & \textbf{68.00{\scriptsize \,$\pm$3.83}} & \textbf{5.76{\scriptsize \,$\pm$0.89}} & \textbf{64.85{\scriptsize \,$\pm$4.27}} & \textbf{5.07{\scriptsize \,$\pm$0.96}} & \textbf{60.18{\scriptsize \,$\pm$2.87}} & \textbf{5.55{\scriptsize \,$\pm$0.54}} \\
\bottomrule
\end{tabular}
\end{table*}

\subsection{Configuration Selection Baselines}
\label{app:selector_baselines}

Below we describe the three baselines used to ablate the disagreement-aware selector.

\subsubsection{Thompson Sampling.}
This variant replaces GraphEvo's acquisition rule with a Thompson-style sampling rule while retaining the GNN surrogate and its MC-dropout uncertainty estimate.
For each candidate $c_i$, we estimate the predictive mean $\mu(c_i)$ and standard deviation $\sigma(c_i)$ from $T$ stochastic forward passes, then draw a score $\tilde{s}_i\sim\mathcal{N}(\mu(c_i),\sigma(c_i)^2)$.
The candidate with the largest sampled score is evaluated in the emulator.
Selection therefore depends on a random draw rather than an explicit uncertainty bonus: a candidate can be selected either for its high predicted performance or for a favorable draw under high uncertainty.

\subsubsection{Deep Ensembles.}
This variant estimates uncertainty across independently trained GNNs rather than stochastic passes through a single GNN.
We train $M=10$ predictors with the same architecture and training data but independent random initializations and dropout masks.
All members are trained on the same warm-up set and updated online with newly evaluated configurations.
For candidate $c_i$, member $m$ produces a prediction $\hat{s}_m(c_i)$; the ensemble mean and sample standard deviation define its predicted score and uncertainty,
\begin{equation}
    \mu(c_i) = \frac{1}{M}\sum_{m=1}^{M}\hat{s}_m(c_i),
\end{equation}
\begin{equation}
    \sigma(c_i) = \sqrt{\frac{1}{M-1}\sum_{m=1}^{M}\bigl(\hat{s}_m(c_i)-\mu(c_i)\bigr)^2},
\end{equation}
We evaluate the candidate maximizing $U(c_i)=\mu(c_i)+\lambda\sigma(c_i)$, with $\lambda=1.0$ as in GraphEvo.
This comparison retains the mean-plus-uncertainty acquisition rule but replaces within-model disagreement with between-model disagreement induced by independent training.

Ensemble disagreement comes mainly from initialization and training randomness on shared data, rather than from how structurally novel a candidate is; GraphEvo's MC-dropout instead probes prediction stability around the candidate's own graph representation, and at a fraction of the cost of maintaining ten predictors.

\subsubsection{LinUCB.}
This variant applies a linear upper-confidence-bound rule to the GNN graph-level embedding $\mathbf{g}_i\in\mathbb{R}^d$ of each candidate $c_i$.
Using previously evaluated embeddings $\mathbf{g}_t$ and rollout scores $s_t$, it maintains the regularized feature matrix $\mathbf{A}=\lambda\mathbf{I}+\sum_t\mathbf{g}_t\mathbf{g}_t^\top$ and reward vector $\mathbf{b}=\sum_t s_t\mathbf{g}_t$, with $\lambda=1.0$.
The linear score estimate is parameterized by $\boldsymbol{\theta}=\mathbf{A}^{-1}\mathbf{b}$, and candidates are ranked by
\begin{equation}
    U(c_i) = \mathbf{g}_i^\top\boldsymbol{\theta} + \alpha\sqrt{\mathbf{g}_i^\top\mathbf{A}^{-1}\mathbf{g}_i},
\end{equation}
with exploration coefficient $\alpha=1.0$.
The highest-ranked candidate is evaluated, after which $\mathbf{A}$ and $\mathbf{b}$ are updated with its embedding and measured score.
The exploration bonus thus rewards feature directions under-explored in the linear embedding space, whereas GraphEvo's disagreement directly measures prediction instability around the candidate's graph structure; under the non-linear reward landscape induced by cross-modal interactions, the structure-based signal selects informative candidates more reliably.

\section{Additional Analysis}
\label{app:additional_analysis}

We further examine GraphEvo's adaptation mechanism, modality contributions, selection hyperparameters, and predictor quality, followed by trajectory comparisons and recurring failure cases.

\subsection{Comparison with Parameter Adaptation}
\label{app:ttl_baseline}

Table~\ref{tab:ttl_baseline} compares GraphEvo with V-GPS~\cite{vgps}, which adapts a trainable value function at test time to rank actions sampled by the VLM. GraphEvo achieves higher SR and lower IMR across three task categories, improving average SR from 49.30\% to 60.18\% and reducing IMR from 10.94\% to 5.55\%. These results favor configuration-level adaptation in PokeGym: evolving perception, strategy, and macros addresses bottlenecks beyond individual action selection.

\subsection{Generalization to Additional Backbones}
\label{app:new_backbones}

Table~\ref{tab:new_backbones} compares zero-shot execution with GraphEvo across four backbones, including Qwen3.6-35B and GPT-5.4 from the main paper. The additional GPT-6.1-sol \cite{gpt6.1} and Claude-Sonnet-5-5 \cite{claudesonnet5.5} results further assess backbone generalization. With GPT-6.1-sol, GraphEvo raises average SR from 60.53\% to 78.60\% (+18.07 points) and lowers IMR from 11.15\% to 5.81\%. With Claude-Sonnet-5-5, average SR increases from 45.61\% to 66.32\% (+20.71 points), while IMR decreases from 16.97\% to 9.92\%. Both backbones improve in every task category, showing that GraphEvo's benefits extend beyond the backbones used in the main paper.

The gains remain substantial even with the stronger zero-shot performance of GPT-6.1-sol, indicating that backbone capability and configuration adaptation provide complementary benefits. For both models, the largest absolute IMR reductions occur in Navigation and Localization, consistent with evolved configurations improving spatially grounded execution rather than merely increasing task completion. These results support GraphEvo as a backbone-general approach to test-time configuration adaptation.

\begin{table*}[htbp!]
\centering
\renewcommand{\arraystretch}{1.0}
\small
\setlength{\tabcolsep}{2pt}
\caption{Sensitivity of the disagreement-aware selector to the exploration-exploitation trade-off weight $\lambda$, measured by Success Rate (SR) and Ineffective Move Rate (IMR). Values are reported as mean with standard error shown in smaller font. $\lambda=0$ corresponds to pure exploitation, $\lambda=2$ to a more exploratory setting, and $\lambda=1$ to the default GraphEvo.}
\label{tab:lambda_sensitivity}
\resizebox{0.90\linewidth}{!}{%
\newcommand{\mse}[2]{#1{\tiny$\pm$#2}}
\begin{tabular*}{0.9\textwidth}{@{\extracolsep{\fill}}l c@{\hspace{2pt}}c c@{\hspace{2pt}}c c@{\hspace{2pt}}c c@{\hspace{2pt}}c@{}}
\toprule
\multirow{2}{*}{$\boldsymbol{\lambda}$} & \multicolumn{2}{c}{\textbf{Navigation}} & \multicolumn{2}{c}{\textbf{Interaction}} & \multicolumn{2}{c}{\textbf{Localization}} & \multicolumn{2}{c}{\textbf{Average}} \\
\cmidrule(lr){2-3} \cmidrule(lr){4-5} \cmidrule(lr){6-7} \cmidrule(lr){8-9}
 & \textbf{SR}$\uparrow$ & \textbf{IMR}$\downarrow$ & \textbf{SR}$\uparrow$ & \textbf{IMR}$\downarrow$ & \textbf{SR}$\uparrow$ & \textbf{IMR}$\downarrow$ & \textbf{SR}$\uparrow$ & \textbf{IMR}$\downarrow$ \\
\midrule
0 & \mse{38.89}{5.85} & \mse{5.95}{1.11} & \mse{62.22}{3.93} & \mse{6.59}{1.00} & \mse{52.12}{5.36} & \mse{6.50}{1.21} & \mse{51.93}{2.99} & \mse{6.36}{0.63} \\
1 & \textbf{\mse{46.11}{6.11}} & \textbf{\mse{5.72}{1.01}} & \textbf{\mse{68.00}{3.83}} & \textbf{\mse{5.76}{0.89}} & \textbf{\mse{64.85}{4.27}} & \textbf{\mse{5.07}{0.96}} & \textbf{\mse{60.18}{2.87}} & \textbf{\mse{5.55}{0.54}} \\
2 & \textbf{\mse{46.11}{6.01}} & \mse{7.46}{1.12} & \mse{56.00}{4.52} & \mse{7.05}{1.23} & \mse{54.55}{4.71} & \mse{8.86}{1.85} & \mse{52.46}{2.94} & \mse{7.71}{0.80} \\
\bottomrule
\end{tabular*}%
} 
\end{table*}

\begin{table*}[htbp]
\centering
\renewcommand{\arraystretch}{1.0}
\small
\setlength{\tabcolsep}{2pt}
\caption{Sensitivity of the disagreement-aware selector to the number of MC-Dropout samples $T$, measured by Success Rate (SR) and Ineffective Move Rate (IMR). $T=10$ is the default GraphEvo setting.}
\label{tab:mcdropout_sensitivity}
\resizebox{0.80\linewidth}{!}{%
\newcommand{\mse}[2]{#1{\tiny$\pm$#2}}
\begin{tabular*}{0.8\textwidth}{@{\extracolsep{\fill}}l c@{\hspace{2pt}}c c@{\hspace{2pt}}c c@{\hspace{2pt}}c c@{\hspace{2pt}}c@{}}
\toprule
\multirow{2}{*}{$\boldsymbol{T}$} & \multicolumn{2}{c}{\textbf{Navigation}} & \multicolumn{2}{c}{\textbf{Interaction}} & \multicolumn{2}{c}{\textbf{Localization}} & \multicolumn{2}{c}{\textbf{Average}} \\
\cmidrule(lr){2-3} \cmidrule(lr){4-5} \cmidrule(lr){6-7} \cmidrule(lr){8-9}
 & \textbf{SR}$\uparrow$ & \textbf{IMR}$\downarrow$ & \textbf{SR}$\uparrow$ & \textbf{IMR}$\downarrow$ & \textbf{SR}$\uparrow$ & \textbf{IMR}$\downarrow$ & \textbf{SR}$\uparrow$ & \textbf{IMR}$\downarrow$ \\
\midrule
5 & \mse{41.11}{5.91} & \mse{9.55}{1.74} & \mse{66.67}{3.92} & \mse{6.36}{0.88} & \mse{56.36}{5.60} & \mse{8.62}{1.93} & \mse{55.61}{3.06} & \mse{8.02}{0.86} \\
10 & \textbf{\mse{46.11}{6.11}} & \textbf{\mse{5.72}{1.01}} & \textbf{\mse{68.00}{3.83}} & \textbf{\mse{5.76}{0.89}} & \textbf{\mse{64.85}{4.27}} & \textbf{\mse{5.07}{0.96}} & \textbf{\mse{60.18}{2.87}} & \textbf{\mse{5.55}{0.54}} \\
\bottomrule
\end{tabular*}%
} 
\end{table*}

\begin{table*}[htbp!]
\centering
\renewcommand{\arraystretch}{1.0}
\small
\setlength{\tabcolsep}{2pt}
\caption{Sensitivity of the GNN predictor to the score definition used for training. The default score combines task success $v$, recovery rate $r$, and ineffective move rate $I$ as $0.5v + 0.5[0.5r + 0.5(1-I)]$. The ``w/o $r$'' variant removes the recovery term, using $0.5v + 0.5(1-I)$.}
\label{tab:score_sensitivity}
\resizebox{0.90\linewidth}{!}{%
\newcommand{\mse}[2]{#1{\tiny$\pm$#2}}
\begin{tabular*}{0.9\textwidth}{@{\extracolsep{\fill}}l c@{\hspace{2pt}}c c@{\hspace{2pt}}c c@{\hspace{2pt}}c c@{\hspace{2pt}}c@{}}
\toprule
\multirow{2}{*}{\textbf{Score}} & \multicolumn{2}{c}{\textbf{Navigation}} & \multicolumn{2}{c}{\textbf{Interaction}} & \multicolumn{2}{c}{\textbf{Localization}} & \multicolumn{2}{c}{\textbf{Average}} \\
\cmidrule(lr){2-3} \cmidrule(lr){4-5} \cmidrule(lr){6-7} \cmidrule(lr){8-9}
 & \textbf{SR}$\uparrow$ & \textbf{IMR}$\downarrow$ & \textbf{SR}$\uparrow$ & \textbf{IMR}$\downarrow$ & \textbf{SR}$\uparrow$ & \textbf{IMR}$\downarrow$ & \textbf{SR}$\uparrow$ & \textbf{IMR}$\downarrow$ \\
\midrule
w/o $r$ & \textbf{\mse{46.67}{5.69}} & \textbf{\mse{5.24}{0.88}} & \mse{57.33}{3.95} & \mse{9.81}{1.81} & \mse{55.76}{4.98} & \mse{10.64}{2.38} & \mse{53.51}{2.79} & \mse{8.60}{1.05} \\
Default & \mse{46.11}{6.11} & \mse{5.72}{1.01} & \textbf{\mse{68.00}{3.83}} & \textbf{\mse{5.76}{0.89}} & \textbf{\mse{64.85}{4.27}} & \textbf{\mse{5.07}{0.96}} & \textbf{\mse{60.18}{2.87}} & \textbf{\mse{5.55}{0.54}} \\
\bottomrule
\end{tabular*}%
} 
\end{table*}

\begin{figure*}[t]
	\centering
	\includegraphics[width=0.8\linewidth]{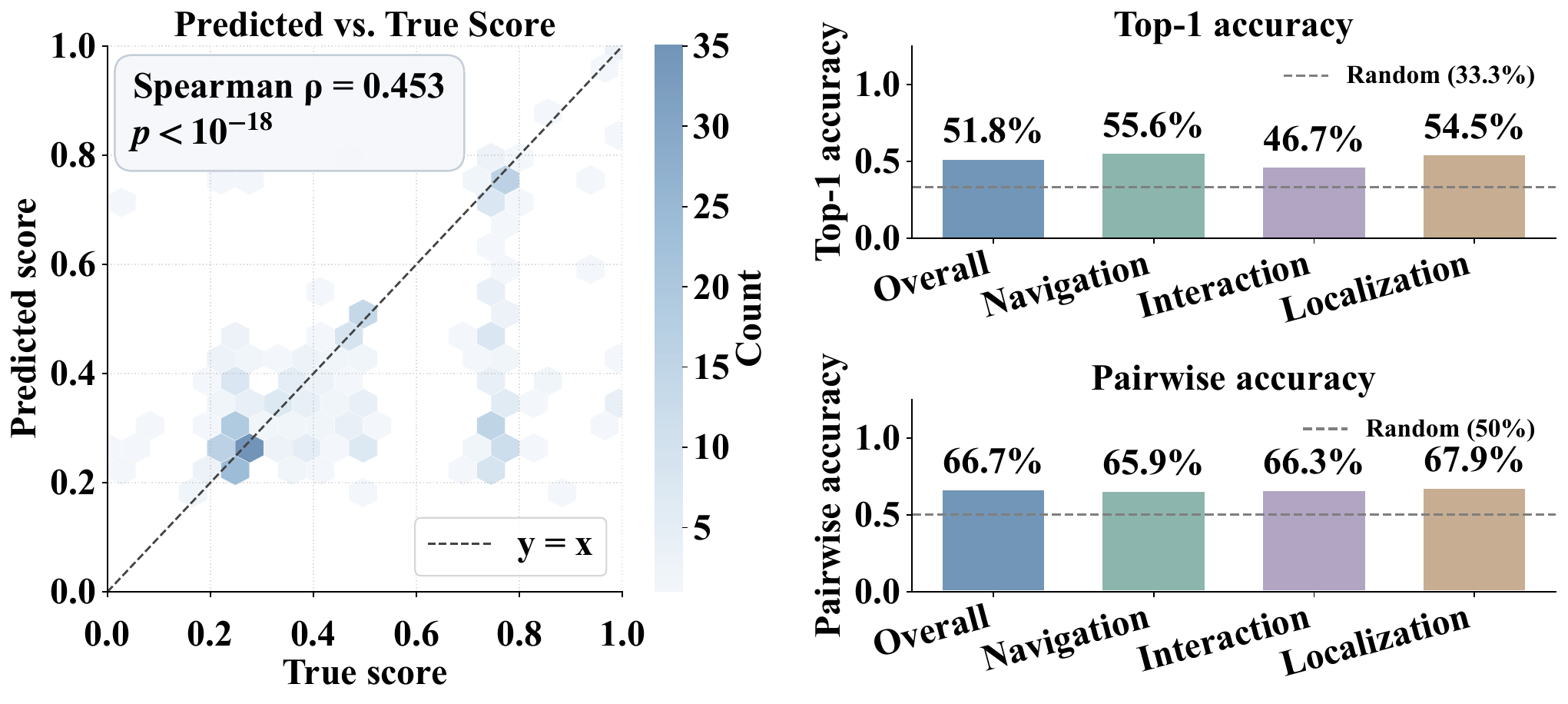}
	\caption{Validation of the GNN predictor. Left: predicted versus true configuration scores. Right: Top-1 selection and pairwise ranking accuracy of the score-based acquisition function.}
	\label{fig:predictor}
\end{figure*}

\subsection{Complete Modality Ablation}
\label{app:complete_modal}

Table~\ref{tab:complete_modal_ablation} compares all nonempty combinations of the three evolving modalities. Strategy-only evolution achieves the highest single-modality SR (43.86\%), whereas visual-only evolution yields 41.05\% SR and the highest IMR (15.58\%), indicating that improved observations still require effective planning and control. Relative to the full configuration, omitting vision produces the largest SR decrease, from 60.18\% to 39.82\%, identifying visual grounding as the most critical bottleneck in PokeGym. Omitting strategies or macros reduces SR to 47.02\% and 50.35\%, respectively. The full configuration combines the highest SR with the lowest IMR (5.55\%). Together, these results show that the gains depend on coordinating perception, strategy, and action rather than improving any one component alone.

\subsection{Sensitivity of the Disagreement Weight $\lambda$}
\label{app:lambda_sensitivity}

The disagreement weight $\lambda$ controls the balance between predicted performance $\mu(c_i)$ and disagreement $\sigma(c_i)$ in $U(c_i)=\mu(c_i)+\lambda\sigma(c_i)$. Table~\ref{tab:lambda_sensitivity} compares $\lambda\in{0,1,2}$. Pure exploitation ($\lambda=0$) reaches 51.93\% SR and 6.36\% IMR, while stronger exploration ($\lambda=2$) yields 52.46\% SR and 7.71\% IMR. The default $\lambda=1$ achieves both the highest SR (60.18\%) and the lowest IMR (5.55\%). The comparison highlights the value of balancing candidate quality with predictive disagreement: ignoring disagreement limits exploration, whereas over-weighting it can divert evaluations toward uncertain but less effective candidates.

\subsection{Sensitivity of MC-Dropout Samples $T$}
\label{app:mcdropout_sensitivity}

Table~\ref{tab:mcdropout_sensitivity} examines the number $T$ of stochastic GNN forward passes used to estimate predictive disagreement. Increasing $T$ from 5 to 10 raises average SR from 55.61\% to 60.18\% and lowers IMR from 8.02\% to 5.55\%, with improvements across all task categories. This is consistent with a better-resolved disagreement estimate providing more effective guidance for candidate selection. Even $T=5$ exceeds the strongest extended baseline's average SR (54.04\%), while $T=10$ delivers the stronger performance used in our default setting.

\subsection{Sensitivity of the Training Score}
\label{app:score_sensitivity}

The training score combines task success $v$, recovery rate $r$, and ineffective move rate $I$ as $0.5v+0.5[0.5r+0.5(1-I)]$. Table~\ref{tab:score_sensitivity} compares this score with a variant excluding recovery, $0.5v+0.5(1-I)$. Excluding $r$ reduces average SR from 60.18\% to 53.51\% and increases IMR from 5.55\% to 8.60\%. The default score improves Interaction and Localization, although the variant performs slightly better on Navigation. The recovery-aware objective therefore provides more useful supervision overall, helping the surrogate favor configurations that sustain progress after mistakes rather than judging them solely by completion and action efficiency.

\begin{figure*}[htbp!]
    \centering
    \includegraphics[width=1.0\textwidth]{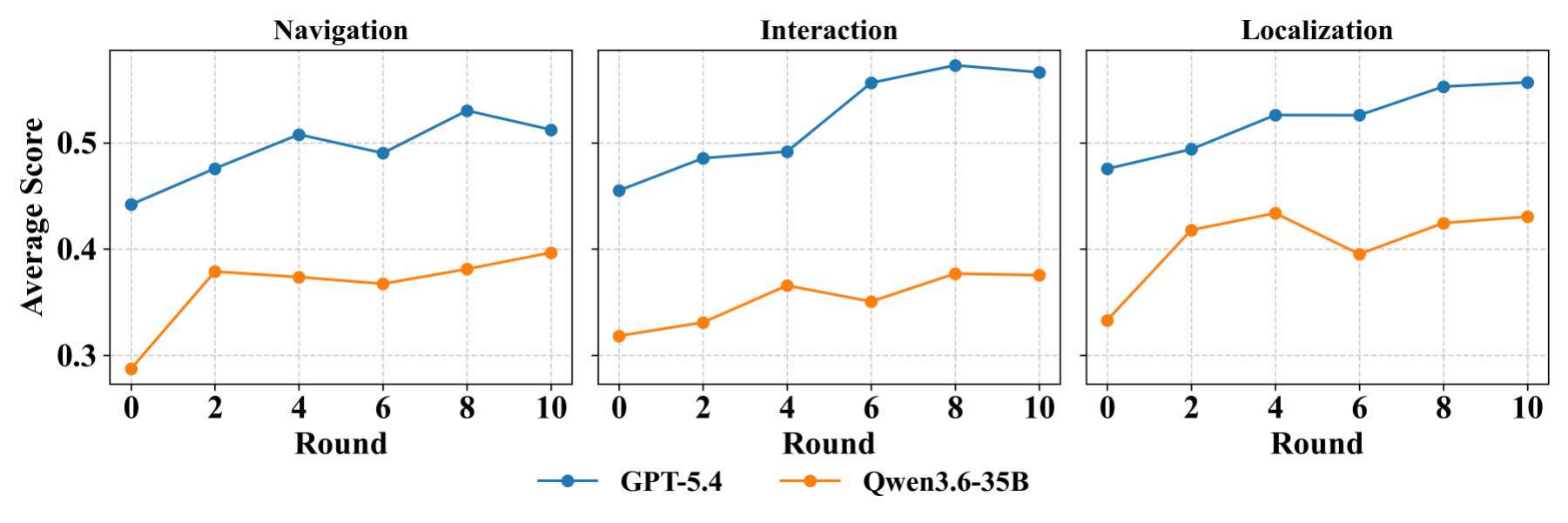}
    \caption{Average score across evolution rounds for GraphEvo with two backbone VLMs on the three task types.}
    \label{fig:score_trends}
\end{figure*}

\subsection{Offline Predictor Validation}
\label{app:predictor}

We evaluate every candidate in an evolution round in the simulator and compare the resulting measured scores with the surrogate's predictions (Fig.~\ref{fig:predictor}). Predicted and measured scores exhibit a positive rank correlation (Spearman $\rho=0.453$, $p<10^{-18}$). Acquisition-based selection identifies the highest-scoring candidate in 51.75\% of cases, compared with 33.3\% for random selection; the predictor also correctly orders 66.67\% of candidate pairs, compared with a 50\% random baseline. These diagnostics show that surrogate predictions provide actionable ranking information, enabling the selector to prioritize promising configurations before committing rollout budget.

\subsection{Evolution Score Trends}
\label{app:score_trends}

Figure~\ref{fig:score_trends} tracks average task scores over 10 evolution rounds for GPT-5.4 and Qwen3.6-35B. Both backbones exhibit an overall upward trend across task categories, with larger early improvements followed by more gradual gains. GPT-5.4 maintains higher scores, particularly on Interaction, while Navigation and Localization improve with some round-to-round fluctuations. These trajectories show that the configuration-update loop translates accumulated experience into progressively stronger behavior across backbones, rather than relying on a single successful configuration proposal. The gains taper off in later rounds, suggesting that GraphEvo converges to a strong and stable configuration within the evaluated budget.

\section{Qualitative Analysis}
\label{app:qualitative_analysis}


\subsection{Qualitative Trajectory Analysis}
\label{sec:qualitative}

\begin{figure*}[t]
    \centering
    \includegraphics[width=\textwidth]{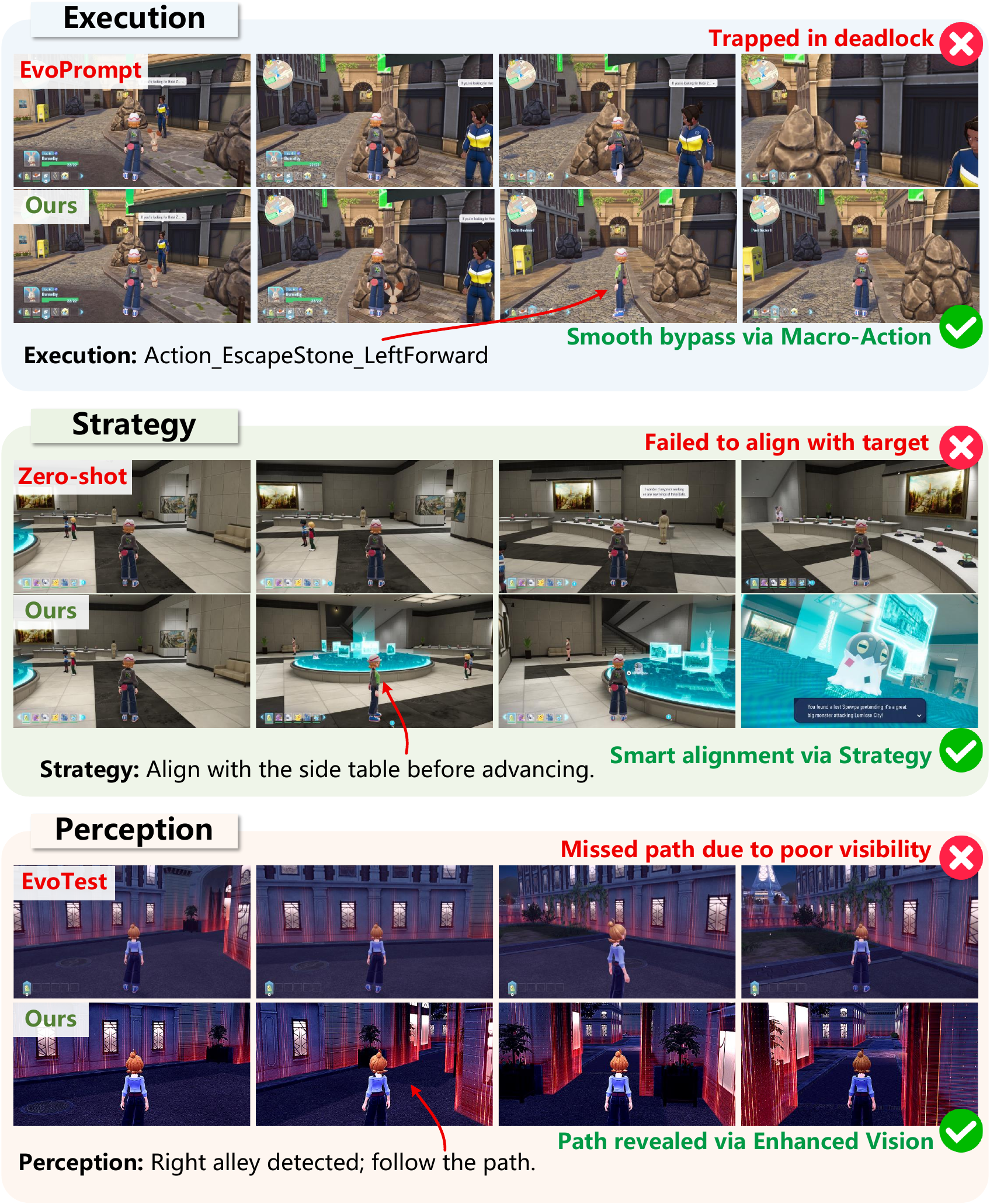}
    \caption{Trajectory comparison between GraphEvo and the other baselines.}
    \label{fig:traj_cmp}
\end{figure*}

Figure~\ref{fig:traj_cmp} illustrates how evolved components address distinct execution bottlenecks. An evolved macro action breaks a collision deadlock, an evolved strategy guides alignment with the interaction target, and visual enhancement makes a poorly visible route easier to identify. These examples connect configuration changes to concrete behavioral improvements: macros expand available maneuvers, strategies guide their use, and visual enhancement improves the observations on which decisions depend.

\subsection{Qualitative Failure Analysis}
\label{app:qualitative_failure}

\begin{figure*}[htbp!] 
	\centering
\includegraphics[width=\textwidth]{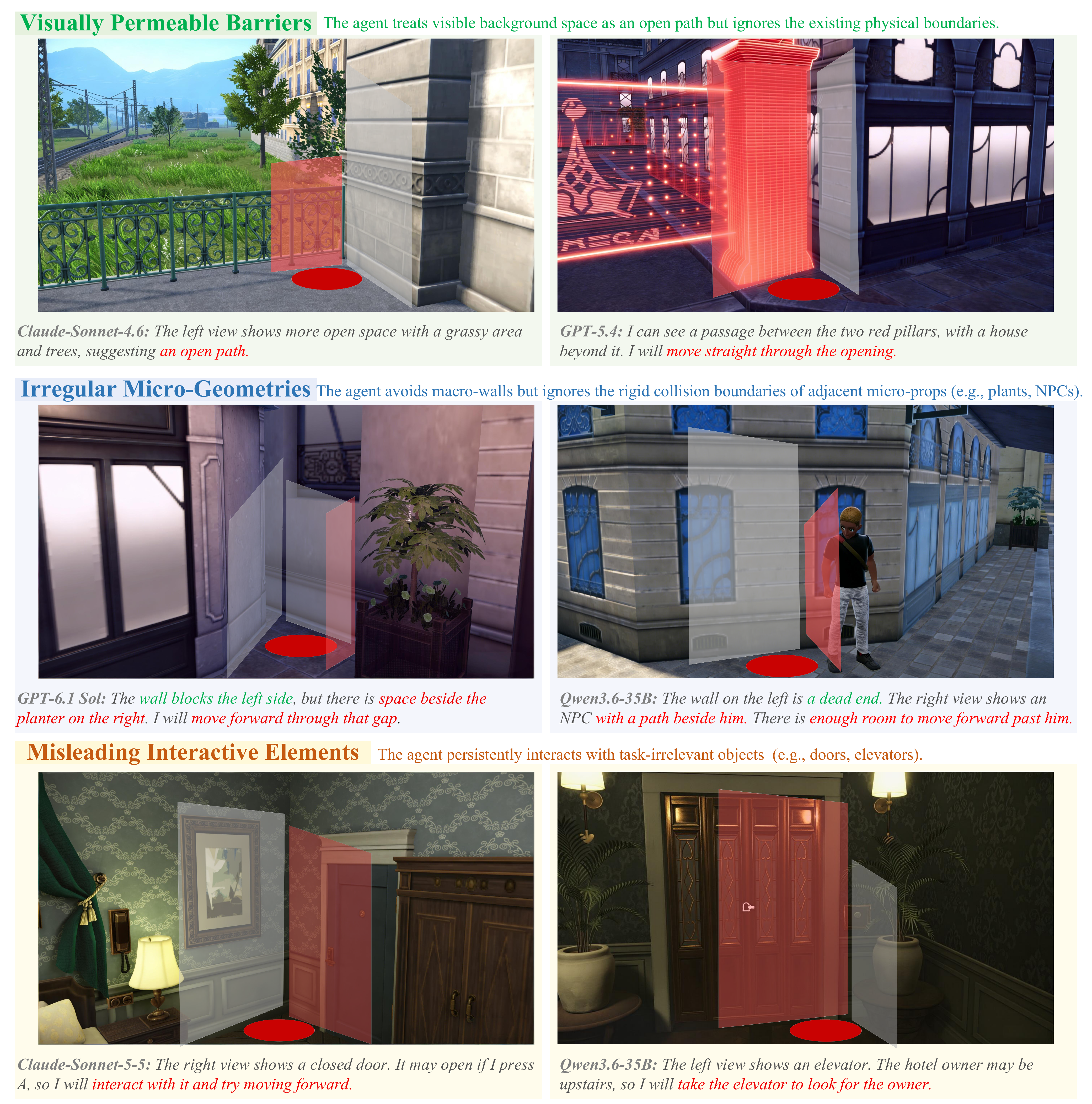}
	
\caption{Representative failure cases in PokeGym: (left) visual-spatial understanding failures, (middle) visual-perception and action-control failures, and (right) planning and reasoning failures.}
	\label{fig:locations}	
\end{figure*}

Figure~\ref{fig:locations} groups recurring failures into three categories. \textbf{Spatial grounding:} visible space beyond a barrier is mistaken for a traversable route. \textbf{Perception and control:} small obstacles are not translated into sufficiently precise collision-avoidance maneuvers. \textbf{Planning:} the agent repeatedly engages with irrelevant doors or signs instead of advancing the task. These cases illustrate the coupled nature of the remaining challenges: recognizing an object is insufficient without reasoning about traversability, executing suitable maneuvers, and maintaining task-directed progress.

\begin{table*}[t]
	\centering
	\small
	\setlength{\tabcolsep}{3pt}
	\caption{Token consumption and API cost per run. Input, output, and total token counts are reported in thousands (per episode). Costs are shown for proprietary closed-source models and omitted for the open-source model.}
	\label{tab:token_and_cost}
	\begin{tabular}{l rrr rrr rrr rrrr}
		\toprule
		& \multicolumn{3}{c}{\textbf{Navigation}} & \multicolumn{3}{c}{\textbf{Interaction}} & \multicolumn{3}{c}{\textbf{Localization}} & \multicolumn{4}{c}{\textbf{Overall}} \\
		\cmidrule{2-4} \cmidrule{5-7} \cmidrule{8-10} \cmidrule{11-14}
		\textbf{Model} & \textbf{In} & \textbf{Out} & \textbf{Total} & \textbf{In} & \textbf{Out} & \textbf{Total} & \textbf{In} & \textbf{Out} & \textbf{Total} & \textbf{In} & \textbf{Out} & \textbf{Total} & \textbf{Cost} \\
		\midrule
		\multicolumn{14}{l}{\textbf{\textit{Zero-shot VLMs}}} \\
		Qwen3.6-35B      &  51k & 24k &  75k &  42k & 18k &  60k &  53k & 24k &  77k &  48k & 22k &  70k & --   \\
		Claude-Sonnet-4.6 & 148k & 20k & 167k & 125k & 16k & 141k & 156k & 21k & 178k & 141k & 19k & 160k & \$0.708 \\
		GPT-5.4          &  64k & 12k &  76k &  49k &  9k &  58k &  66k & 12k &  78k &  59k & 11k &  69k & \$0.307 \\
		\midrule
		\multicolumn{14}{l}{\textbf{\textit{Game Agent Frameworks}}} \\
		Cradle           & 166k & 17k & 183k & 110k & 11k & 121k & 114k & 15k & 128k & 129k & 14k & 143k & \$0.540 \\
		Voyager          & 224k & 10k & 234k & 149k &  7k & 156k & 215k &  9k & 224k & 192k &  8k & 200k & \$0.630 \\
		\midrule
		\multicolumn{14}{l}{\textbf{\textit{Self-evolving/TTL Methods}}} \\
		Reflexion        & 146k & 10k & 156k & 110k &  8k & 118k & 154k & 11k & 165k & 134k &  9k & 144k & \$0.491 \\
		EvoPrompt        & 186k & 12k & 198k & 115k &  9k & 124k & 145k & 12k & 157k & 146k & 11k & 157k & \$0.542 \\
		PromptBreeder    & 119k & 13k & 132k &  55k & 10k &  65k &  71k & 12k &  84k &  80k & 12k &  92k & \$0.379 \\
		AgentSquare      & 145k & 13k & 158k &  99k &  9k & 108k & 145k & 14k & 159k & 127k & 12k & 139k & \$0.501 \\
		EvoTest          & 204k & 11k & 215k & 152k &  8k & 160k & 131k & 12k & 142k & 162k & 10k & 172k & \$0.572 \\
		\midrule
		\textbf{GraphEvo (Ours)} & 198k & 10k & 209k & 133k &  7k & 140k & 163k & 10k & 173k & 163k &  9k & 171k & \$0.560 \\
		\bottomrule
	\end{tabular}
\end{table*}

\section{Token Consumption and API Cost}
\Cref{tab:token_and_cost} reports the input and output token consumption for each evaluated model across the three task categories and in total per run.
Claude-Sonnet-4.6 consumes the most input tokens per run (141k), while GPT-5.4 is substantially more efficient (59k input / 11k output). Among the GPT-5.4-based methods, Voyager uses the most input tokens (192k) because of its skill-library queries, whereas PromptBreeder is the most input-efficient (80k). Qwen3.6-35B produces the largest output volume (22k), indicating more verbose responses. GraphEvo (Ours) consumes 163k input and 9k output tokens per run, achieving strong performance with a moderate token budget of \$0.560 per run.

\section{Quantitative Complexity Analysis}
\label{app:complexity_analysis}

To mathematically illustrate the challenge PokeGym poses to vision-language agents, we quantify the environment's complexity across three fundamental dimensions: state space, action space, and decision horizon.

\subsection{State Space Complexity}
We simplify the analysis by omitting the environmental states (\eg, dynamic NPCs) and focus on the spatial state. 
This state can be represented as $s = (x, z, \theta)$, where $(x, z)$ denotes the horizontal position and $\theta$ represents the camera yaw angle. 
We explicitly omit the vertical coordinate $y$ and the camera pitch angle, as they remain nearly constant in our evaluated tasks. 

To estimate the size of the state space $|S|$, we discretize the map with a spatial step size of $\Delta d = 1$ unit and the viewing direction with an angular step size of $\Delta \theta = 1^\circ$. Let $R$ denote the map area.
The resulting state space size can be approximated as:
\begin{equation}
	|S| \approx \left( \frac{R}{\Delta d^2} \right) \times \left( \frac{360^\circ}{\Delta \theta} \right).
\end{equation}

Since map sizes vary across tasks in PokeGym, we further estimate the state space range using the smallest map with area $R_{\min} = 186.65$ and the largest map with area $R_{\max} = 2418.12$:
\begin{equation}
	\scalebox{0.9}{$\displaystyle |S_{\min}| \approx \lceil 186.65 \rceil \times 360 = 187 \times 360 = 67,320,$}
\end{equation}
\begin{equation}
	\scalebox{0.9}{$\displaystyle |S_{\max}| \approx \lceil 2418.12 \rceil \times 360 = 2419 \times 360 = 870,840.$}
\end{equation}

This demonstrates that even under a highly simplified assumption with coarse discretization, the agent still faces a massive state space relying purely on visual observations.

\subsection{Action Space Complexity}
\label{action_space_complexity}


We analyze the action space complexity under the discrete high-level action paradigm used in PokeGym.
The base action set contains at least 7 macro actions (\eg, MoveForward, RotateLeft). 
Since three actions are executed per query, the size of the base action space per decision step is:
\begin{equation}
	|A_{base}| = 7^3 = 343.
\end{equation}
Through test-time evolution, GraphEvo can augment this set, yielding at most 12 macro actions. The resulting evolved action space per decision step is therefore:
\begin{equation}
	|A_{evolved}| = 12^3 = 1{,}728.
\end{equation}
This large discrete action space, which can grow substantially during evolution, requires agents to possess strong action sequencing and precise multi-step execution capability.

\subsection{Decision Horizon Complexity}

%
%

We evaluate the game tree complexity $\mathcal{O}(b^d)$, where $b$ represents the effective branching factor per environment step and $d$ is the maximum decision depth. 
According to our task budgets, the maximum effective horizon reaches up to $d = 360$ environment steps. 

For the discrete high-level action paradigm, the effective branching factor is at least $b = 7$ before evolution and can grow to at most $b = 12$ after GraphEvo augments the action set. The sizes of the corresponding decision trees are therefore:
\begin{equation}
	\text{Game Tree Size}_{base} \approx \mathcal{O}(7^{360}) \approx 10^{304},
\end{equation}
\begin{equation}
	\text{Game Tree Size}_{evolved} \approx \mathcal{O}(12^{360}) \approx 10^{389}.
\end{equation}

This explosion highlights that brute-force exploration or short-sighted planning is intractable in PokeGym. To succeed, the agent must maintain a coherent, long-term semantic plan and robust error-recovery strategies.
\section{Limitations and Future Work}
\label{app:limitations}

While PokeGym and GraphEvo advance test-time learning in visually-driven, long-horizon 3D games, several limitations remain and motivate future work.

\subsection{Evaluation and Computational Cost}

The GNN-based performance predictor substantially reduces the number of expensive emulator rollouts, yet each GraphEvo run still consumes around 163k input tokens and costs approximately \$0.560 per run with GPT-5.4.
This cost limits large-scale hyper-parameter sweeps and broad accessibility for researchers without generous API budgets.


\subsection{Single-Episode and Online Adaptation}

GraphEvo currently evolves configurations across multiple episodes or rounds.
A natural next step is \textit{intra-episode} or fully online adaptation, where the agent revises its visual pipeline, language strategy, and macro actions within a single playthrough based on real-time feedback.
Such online learning would better mimic human-like improvisation and could recover from unforeseen situations that are not captured by the initial rollout history.
Developing lightweight update rules and memory mechanisms that support safe, stable online evolution is an important research direction.

\begin{figure*}[htbp!]
    \centering
    \includegraphics[width=\textwidth]{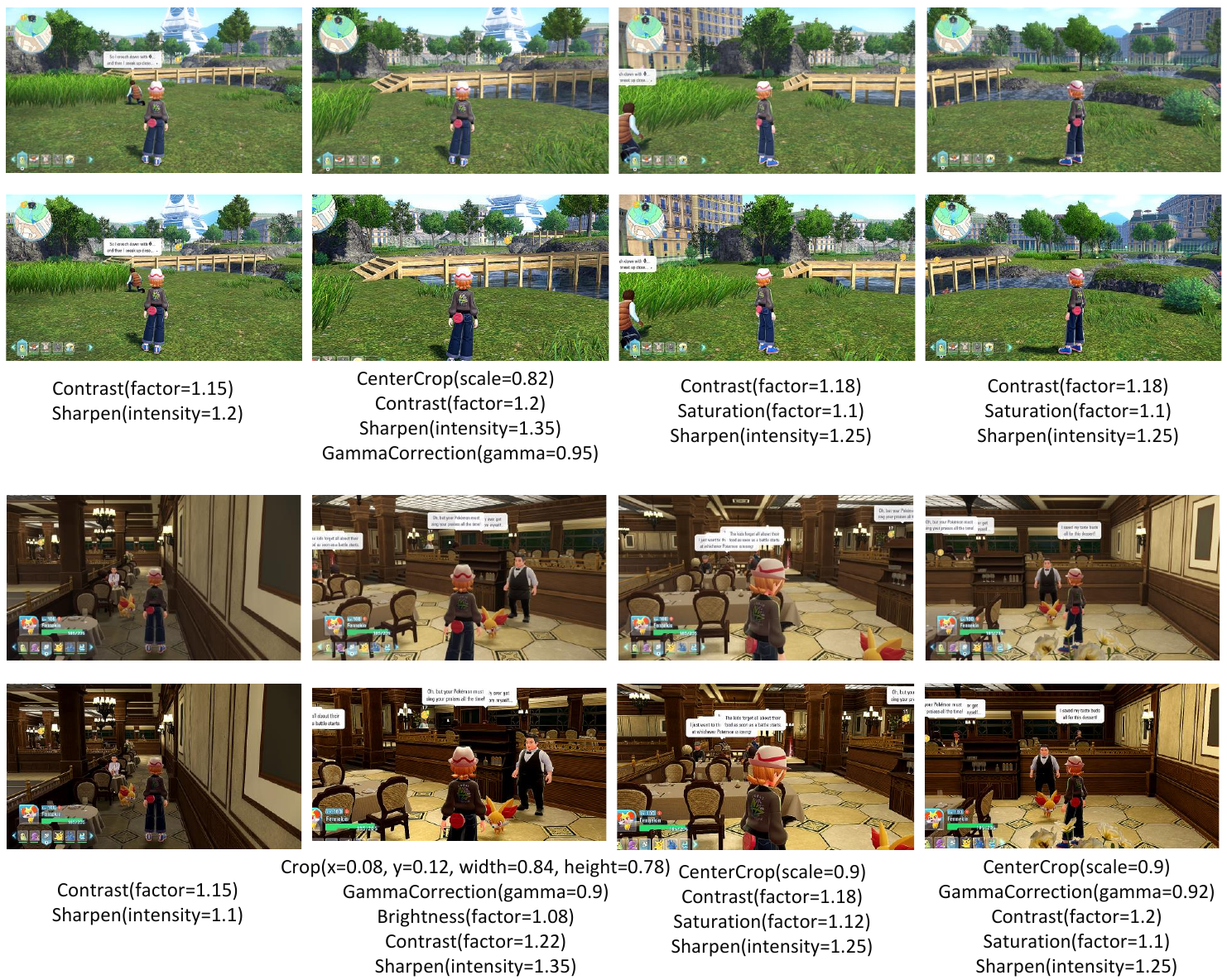}
    \caption{Examples of visual enhancements evolved by GraphEvo (part~1). Each figure shows original frames in the top row and their enhanced counterparts in the bottom row after per-frame operations.}
    \label{fig:enhancements1}
\end{figure*}

\begin{figure*}[htbp]
    \centering
    \includegraphics[width=\textwidth]{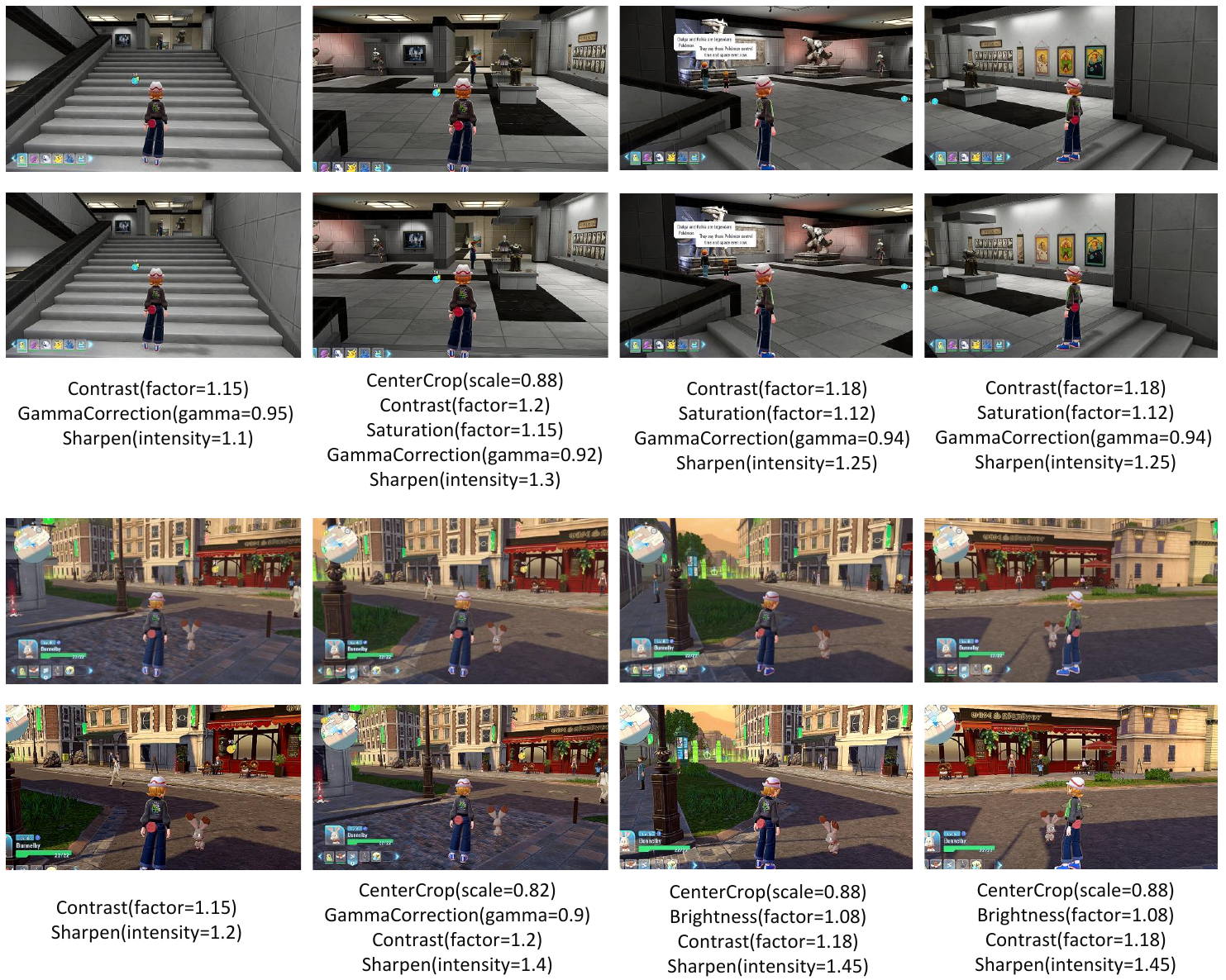}
    \caption{Examples of visual enhancements evolved by GraphEvo (part~2).}
    \label{fig:enhancements2}
\end{figure*}

\begin{figure*}[htbp]
    \centering
    \includegraphics[width=\textwidth]{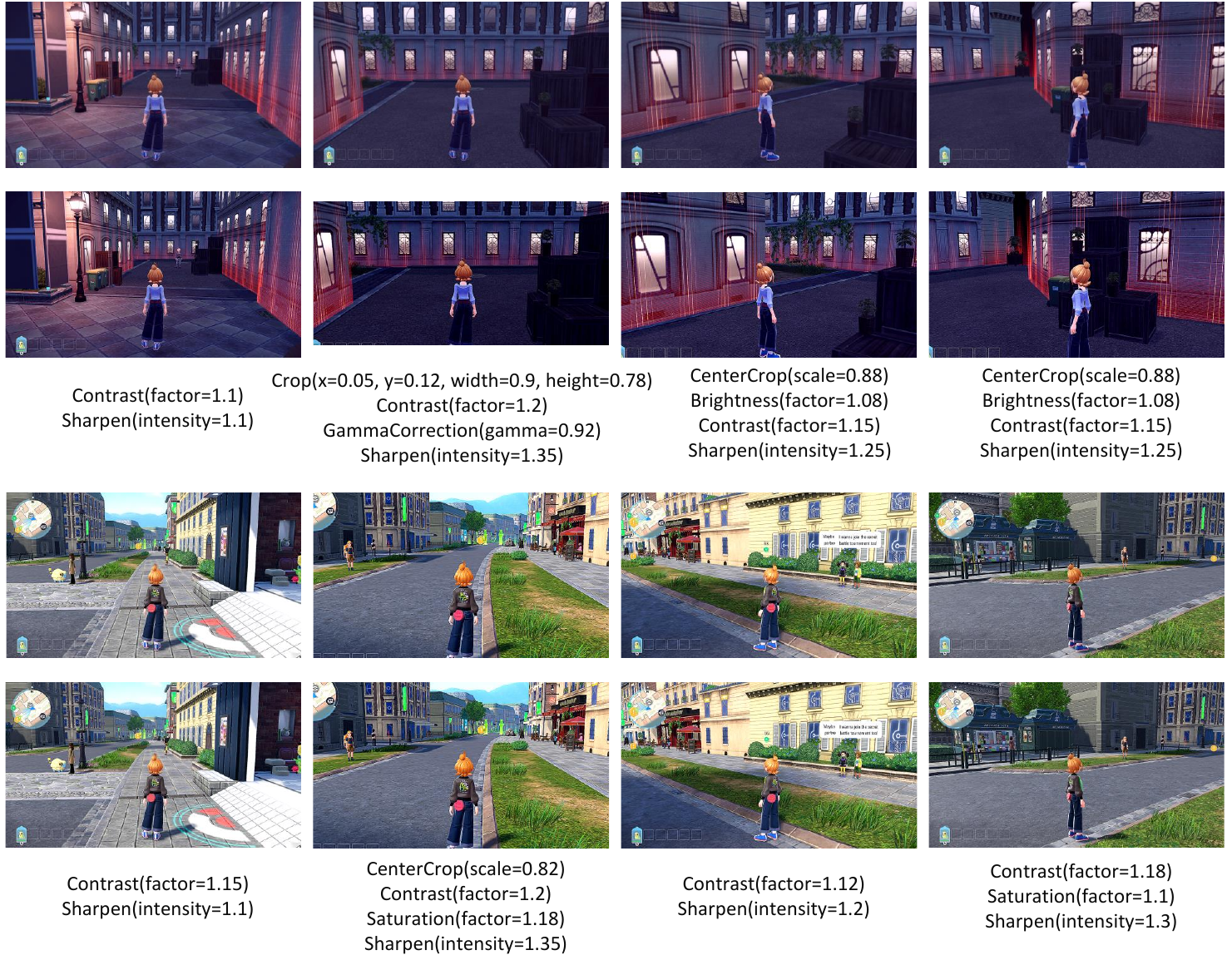}
    \caption{Examples of visual enhancements evolved by GraphEvo (part~3).}
    \label{fig:enhancements3}
\end{figure*}

\begin{figure*}[htbp]
    \centering
    \includegraphics[width=\textwidth]{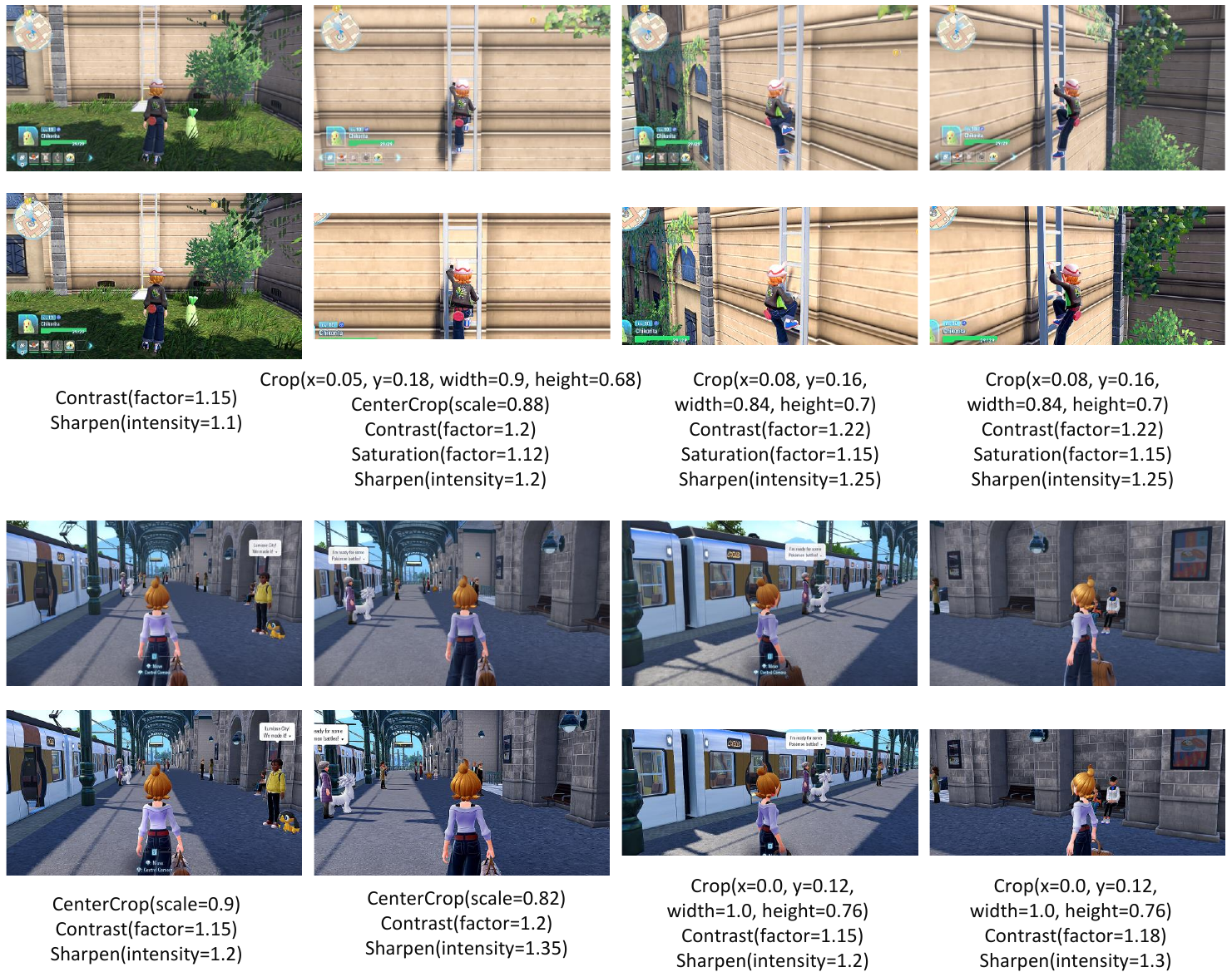}
    \caption{Examples of visual enhancements evolved by GraphEvo (part~4).}
    \label{fig:enhancements4}
\end{figure*}

\subsection{Broader Future Directions}

Looking ahead, we identify several high-impact extensions.
First, integrating explicit 3D scene understanding---such as depth estimation, semantic mapping, or neural radiance fields---could mitigate the spatial-reasoning failures observed in our analysis.
Second, combining GraphEvo with reinforcement learning or model-based planning may produce stronger policies by unifying discrete evolutionary search with gradient-based or sample-based optimization.
Finally, incorporating human demonstrations or external game knowledge as an initialization prior could accelerate evolution and reduce reliance on costly trial-and-error.
%
%
%

\section{Examples and Prompts}
\label{app:examples}

\Cref{fig:enhancements1,fig:enhancements2,fig:enhancements3,fig:enhancements4} show examples of visual enhancements evolved by GraphEvo.
Language strategy examples are shown in \Cref{fig:strategy_example_1,fig:strategy_example_2,fig:strategy_example_3}, and macro-action examples are shown in \Cref{fig:macro_action_example_1,fig:macro_action_example_2}.
The prompts used in PokeGym and GraphEvo are shown in \Cref{prompt:planning}, \Cref{prompt:summary}, and \Cref{prompt:evolution}.

\clearpage

\newtcolorbox{strategybox}[1][]{
colback=black!2,
colframe=black!35,
colbacktitle=black!7,
coltitle=black,
fonttitle=\normalfont\small\bfseries,
fontupper=\normalfont\small,
title=#1,
boxrule=0.4pt,
arc=1mm,
before upper={\raggedright\setlength{\parindent}{0pt}\setlength{\parskip}{2pt}},
left=2mm, right=2mm, top=2mm, bottom=2mm
}

\begin{figure*}[p]
\centering
\begin{strategybox}[{Strategy Example 1}]
1. Treat side-view marker sightings as coarse navigation only; only press A when the female NPC with the blue marker is front-centered from the open street side and the wall is not dominating the front view.
\par
2. If the front view becomes wall-heavy or building-close, immediately create depth with multiple backward steps before rotating; never keep turning in place while pinned to the wall.
\par
3. In dense NPC clusters, do not interact from a parallel wall-hugging angle. First step out toward open space, then re-approach with the marked female slightly centered and isolated from the neighboring NPC.
\par
4. If generic ambient dialogue appears, assume wrong target or wrong angle. Finish/close it, back away, and retry from a meaningfully different street-side position rather than making tiny local corrections.
\par
5. Once the marked female NPC is clearly engaged and dialogue has started, commit to advancing with A until the conversation ends; do not resume movement unless there is strong evidence the wrong NPC was selected.
\par
\end{strategybox}
\caption{Example of an evolved language strategy for NPC interaction tasks.}
\label{fig:strategy_example_1}
\end{figure*}

\begin{figure*}[p]
\centering
\begin{strategybox}[{Strategy Example 2}]
1. Use side views only to choose turning direction. If the ladder, bridge, or stairs are visible only in Left View or Right View, rotate toward that side first; do not move forward until the target is mostly in the front view.
\par
2. At the ladder, prioritize clean front alignment over proximity. When the ladder is centered and close, press A immediately; once mounted, stop turning and chain MoveForward until the top platform is clearly reached.
\par
3. Treat apparent open space beyond a near railing, hedge, pillar, or wall edge as suspect. If one or two forward moves produce little scene-scale change, assume a blocker, back up, rotate, and re-approach from a cleaner angle.
\par
4. Avoid micro-oscillation near bridge and stair entrances. Once the bridge or stairs are broadly centered in front and side views are not dominated by a near wall, commit to repeated MoveForward instead of trying to perfect alignment.
\par
5. Clear dialogue instantly with PressA, then resume route execution with a decisive action. Repeated dialogue usually means the agent is stalling in place, so prefer a larger reorientation or forward commit over small corrective turns afterward.
\par
\end{strategybox}
\caption{Example of an evolved language strategy for navigation tasks with vertical traversal.}
\label{fig:strategy_example_2}
\end{figure*}

\begin{figure*}[p]
\centering
\begin{strategybox}[{Strategy Example 3}]
1. Use the orange wall as a hazard boundary, not as a path to hug; if the wall dominates the front or right view, first create lateral/backup clearance before trying to advance.
\par
2. Pass the two-barrier alley by staying centered and committing to an early wide bypass around the first obstacle cluster, rather than alternating tiny left-right corrections while already in contact.
\par
3. When the far end looks like a close dead-end wall, assume it is a collision trap until a clear post/opening or explicit leave prompt is stably aligned; do not PressA on ambiguous close-up orange geometry.
\par
4. If 2-3 actions produce almost no scene change, treat it as a real deadlock: back up multiple steps, rotate decisively, and reacquire the corridor from a less cramped viewpoint before moving forward again.
\par
5. After clearing the barrier section, anchor on the leave interaction immediately; avoid drifting into the open street/plaza unless the leave transition is already confirmed.
\par
\end{strategybox}
\caption{Example of an evolved language strategy for constrained corridor navigation.}
\label{fig:strategy_example_3}
\end{figure*}

\newtcolorbox{actionbox}[1][]{
colback=black!2,
colframe=black!35,
colbacktitle=black!7,
coltitle=black,
fonttitle=\normalfont\small\bfseries,
fontupper=\normalfont\small,
title=#1,
boxrule=0.4pt,
arc=1mm,
before upper={\raggedright\setlength{\parindent}{0pt}\setlength{\parskip}{2pt}},
left=2mm, right=2mm, top=2mm, bottom=2mm
}

\begin{figure*}[p]
\centering
\begin{actionbox}[{Macro Action Example 1}]
[1] Action\_\allowbreak{}EscapeDoorframe\_\allowbreak{}LeftBias
\par
Description: Reliable escape when pressed against a wall, hedge, or left-side doorframe. Clears collision with two backward moves, then rotates left to re-approach from a new angle.
\par
Code:
\par
private static async Task Action\_\allowbreak{}EscapeDoorframe\_\allowbreak{}LeftBias()
\par
\{
\par
\hspace*{1.4em}Console.WriteLine("[Action API] Macro: Escape Doorframe Left Bias");
\par
\hspace*{1.4em}await Action\_\allowbreak{}MoveBackward();
\par
\hspace*{1.4em}await Action\_\allowbreak{}MoveBackward();
\par
\hspace*{1.4em}await Action\_\allowbreak{}RotateLeft();
\par
\hspace*{1.4em}await Action\_\allowbreak{}MoveForward();
\par
\}
\par
\smallskip
[2] Action\_\allowbreak{}EscapeDoorframe\_\allowbreak{}RightBias
\par
Description: Reliable escape when pressed against a wall, hedge, or right-side doorframe. Clears collision with two backward moves, then rotates right to re-approach from a new angle.
\par
Code:
\par
private static async Task Action\_\allowbreak{}EscapeDoorframe\_\allowbreak{}RightBias()
\par
\{
\par
\hspace*{1.4em}Console.WriteLine("[Action API] Macro: Escape Doorframe Right Bias");
\par
\hspace*{1.4em}await Action\_\allowbreak{}MoveBackward();
\par
\hspace*{1.4em}await Action\_\allowbreak{}MoveBackward();
\par
\hspace*{1.4em}await Action\_\allowbreak{}RotateRight();
\par
\hspace*{1.4em}await Action\_\allowbreak{}MoveForward();
\par
\}
\par
\smallskip
[3] Action\_\allowbreak{}CommitEnter\_\allowbreak{}LeftFacing
\par
Description: Use when the entrance/interior is visible to the left and the front is nearly clear. Rotates into the doorway and chains forward movement to cross the threshold instead of hesitating at the frame.
\par
Code:
\par
private static async Task Action\_\allowbreak{}CommitEnter\_\allowbreak{}LeftFacing()
\par
\{
\par
\hspace*{1.4em}Console.WriteLine("[Action API] Macro: Commit Enter Left Facing");
\par
\hspace*{1.4em}await Action\_\allowbreak{}RotateLeft();
\par
\hspace*{1.4em}await Action\_\allowbreak{}MoveForward();
\par
\hspace*{1.4em}await Action\_\allowbreak{}MoveForward();
\par
\}
\par
\smallskip
\end{actionbox}
\caption{Example of evolved macro actions for NPC interaction and collision recovery.}
\label{fig:macro_action_example_1}
\end{figure*}

\begin{figure*}[p]
\centering
\ContinuedFloat
\begin{actionbox}[{Macro Action Example 1 (continued)}]
[4] Action\_\allowbreak{}CommitEnter\_\allowbreak{}RightFacing
\par
Description: Use when the entrance/interior is visible to the right and the front is nearly clear. Rotates into the doorway and chains forward movement to cross the threshold cleanly.
\par
Code:
\par
private static async Task Action\_\allowbreak{}CommitEnter\_\allowbreak{}RightFacing()
\par
\{
\par
\hspace*{1.4em}Console.WriteLine("[Action API] Macro: Commit Enter Right Facing");
\par
\hspace*{1.4em}await Action\_\allowbreak{}RotateRight();
\par
\hspace*{1.4em}await Action\_\allowbreak{}MoveForward();
\par
\hspace*{1.4em}await Action\_\allowbreak{}MoveForward();
\par
\}
\par
\smallskip
[5] Action\_\allowbreak{}RecenterAndTalk
\par
Description: Final interaction macro for when the clerk is already nearly centered in front. Makes a small orientation correction, steps into interaction range, then presses A.
\par
Code:
\par
private static async Task Action\_\allowbreak{}RecenterAndTalk()
\par
\{
\par
\hspace*{1.4em}Console.WriteLine("[Action API] Macro: Recenter And Talk");
\par
\hspace*{1.4em}await Action\_\allowbreak{}RotateLeft();
\par
\hspace*{1.4em}await Action\_\allowbreak{}MoveForward();
\par
\hspace*{1.4em}await Action\_\allowbreak{}PressA();
\par
\}
\par
\end{actionbox}
\caption{Example of evolved macro actions for NPC interaction and collision recovery. (continued)}
\end{figure*}

\begin{figure*}[p]
\centering
\begin{actionbox}[{Macro Action Example 2}]
[1] Action\_\allowbreak{}EscapeCorner\_\allowbreak{}BackBackRotateRightAdvance
\par
Description: Use when the front view is blocked by a bush, wall, railing, or pillar and the route seems more open to the right. This creates space before re-aiming, preventing repeated scraping in the same corner.
\par
Code:
\par
private static async Task Action\_\allowbreak{}EscapeCorner\_\allowbreak{}BackBackRotateRightAdvance()
\par
\{
\par
\hspace*{1.4em}Console.WriteLine("[Action API] Macro: Escape Corner - Back Back Rotate Right Advance");
\par
\hspace*{1.4em}await Action\_\allowbreak{}MoveBackward();
\par
\hspace*{1.4em}await Action\_\allowbreak{}MoveBackward();
\par
\hspace*{1.4em}await Action\_\allowbreak{}RotateRight();
\par
\hspace*{1.4em}await Action\_\allowbreak{}MoveForward();
\par
\}
\par
\smallskip
[2] Action\_\allowbreak{}EscapeCorner\_\allowbreak{}BackBackRotateLeftAdvance
\par
Description: Mirror version for cases where the route is clearer to the left. Useful for rooftop railing traps and ladder-base wall entanglement.
\par
Code:
\par
private static async Task Action\_\allowbreak{}EscapeCorner\_\allowbreak{}BackBackRotateLeftAdvance()
\par
\{
\par
\hspace*{1.4em}Console.WriteLine("[Action API] Macro: Escape Corner - Back Back Rotate Left Advance");
\par
\hspace*{1.4em}await Action\_\allowbreak{}MoveBackward();
\par
\hspace*{1.4em}await Action\_\allowbreak{}MoveBackward();
\par
\hspace*{1.4em}await Action\_\allowbreak{}RotateLeft();
\par
\hspace*{1.4em}await Action\_\allowbreak{}MoveForward();
\par
\}
\par
\smallskip
[3] Action\_\allowbreak{}LadderCenterAndInteract\_\allowbreak{}RightBias
\par
Description: Use when the ladder is near the front but slightly right-offset or when the right side suggests the ladder lane. This stabilizes approach and immediately attempts the climb.
\par
Code:
\par
private static async Task Action\_\allowbreak{}LadderCenterAndInteract\_\allowbreak{}RightBias()
\par
\{
\par
\hspace*{1.4em}Console.WriteLine("[Action API] Macro: Ladder Center And Interact - Right Bias");
\par
\hspace*{1.4em}await Action\_\allowbreak{}RotateRight();
\par
\hspace*{1.4em}await Action\_\allowbreak{}MoveForward();
\par
\hspace*{1.4em}await Action\_\allowbreak{}PressA();
\par
\}
\par
\smallskip
\end{actionbox}
\caption{Example of evolved macro actions for vertical traversal and corridor escape.}
\label{fig:macro_action_example_2}
\end{figure*}

\begin{figure*}[p]
\centering
\ContinuedFloat
\begin{actionbox}[{Macro Action Example 2 (continued)}]
[4] Action\_\allowbreak{}LadderCenterAndInteract\_\allowbreak{}LeftBias
\par
Description: Mirror version for when the ladder is near the front but slightly left-offset. Designed to convert a side glimpse into a centered ladder mount attempt.
\par
Code:
\par
private static async Task Action\_\allowbreak{}LadderCenterAndInteract\_\allowbreak{}LeftBias()
\par
\{
\par
\hspace*{1.4em}Console.WriteLine("[Action API] Macro: Ladder Center And Interact - Left Bias");
\par
\hspace*{1.4em}await Action\_\allowbreak{}RotateLeft();
\par
\hspace*{1.4em}await Action\_\allowbreak{}MoveForward();
\par
\hspace*{1.4em}await Action\_\allowbreak{}PressA();
\par
\}
\par
\smallskip
[5] Action\_\allowbreak{}BridgeOrStairCommitForward
\par
Description: Use when the bridge or stairs are already mostly centered in front and no close wall dominates either side. This prevents over-rotation and forces decisive progression through the route segment.
\par
Code:
\par
private static async Task Action\_\allowbreak{}BridgeOrStairCommitForward()
\par
\{
\par
\hspace*{1.4em}Console.WriteLine("[Action API] Macro: Bridge/Stair Commit Forward");
\par
\hspace*{1.4em}await Action\_\allowbreak{}MoveForward();
\par
\hspace*{1.4em}await Action\_\allowbreak{}MoveForward();
\par
\hspace*{1.4em}await Action\_\allowbreak{}MoveForward();
\par
\}
\par
\end{actionbox}
\caption{Example of evolved macro actions for vertical traversal and corridor escape. (continued)}
\end{figure*}

\begin{figure*}[t]
\centering
\begin{promptbox}[Prompt 1: Planning]
\small
You are an experienced player of Pokemon Legends ZA. Your goal is to complete the main quests one by one.\\
Task: \{task\}\\
\\
=== ACTION SPACE ===\\
Movement: MoveForward, MoveBackward, MoveLeft, MoveRight (Move the character a short distance forward or backward, left or right in the current direction, to reach or approach the target location.)\\
Camera: RotateLeft, RotateRight (Turn the camera view to the left or right to find something or to provide a better perspective.)\\
Interaction: PressA (Confirm the dialog box/Talk/Pick up)\\
\{macro\_actions\}\\
\\
=== VISUAL INPUT DEFINITION ===\\
You will receive 4 images representing the agent's status:\\
1. \textbf{Previous Screen}: The state before your last action. \\
2. \textbf{Current Screen (Front)}: The current main view.\\
3. \textbf{Left View}: The current visual information to your left.\\
4. \textbf{Right View}: The current visual information to your right.\\
\\
\{strategies\}\\
\\
Provide the step-by-step reasoning:\\
1. describe the differences between the previous and current screens if available and verify the effectiveness of the previous action execution\\
2. analyze the key information from the left view frame and the right frame view\\
3. plan the next 3 actions based on the reasoning of the first and second steps\\
\\
After the step-by-step reasoning, you will finish by returning in this JSON format as follows:\\
```json\\
\{\{\\
    "actions": ["action1", "action2", ""]\\
\}\}\\
```\\
If no action is needed (waiting/idling), use an empty string `""`.\\
\end{promptbox}
\caption{Prompt of Planning}
\label{prompt:planning}
\end{figure*}

\begin{figure*}[t]
\centering
\begin{promptbox}[Prompt 2: Trajectory Summarization]
\small
You are an expert AI behavior analyst specializing in an agent playing Pokémon Legends: Z-A. \\
Your task is to analyze the raw trajectory log of an agent playing a single episode and provide a structured, clear summary.\\
\\
The agent only has pure visual input (what it sees on screen) and can output discrete actions. \\
The trajectory may be very long, so you need to group continuous steps into logical "Behavioral Phases".\\
\\
<task>\\
\{task\}\\
</task>\\
\\
<trajectory>\\
\{trajectory\}\\
</trajectory>\\
\\
<result>\\
\{result\}\\
</result>\\
\\
Please summarize the trajectory following this structure:\\
\\
1. \textbf{Overall Performance \& Outcome}: \\
   - State whether the task was successful or failed based on the <result>.\\
   - Provide a 1-2 sentence high-level summary of what the agent actually did vs. what it was supposed to do.\\
\\
2. \textbf{Phase-by-Phase Breakdown}:\\
   Group the multiple steps into 3-5 logical phases. For each phase, clearly extract:\\
   - \textbf{Visual Perception}: What did the agent see?\\
   - \textbf{Logical Reasoning}: What was the agent's intent?\\
   - \textbf{Action Sequence}: A brief summary of the actions taken.\\
\\
3. \textbf{Critical Errors, Deadlocks, and Bottlenecks}:\\
   Identify specific moments where the agent wasted steps or got stuck. Specifically look for:\\
   - \textbf{Deadlocks \& Spatial Errors}: Did the agent get physically stuck? Did it experience an "Unaware Deadlock" (hallucinating progress while stuck on an obstacle) or an "Aware Deadlock" (knowing it's stuck but failing to escape)? Describe the exact sequence of primitive actions that kept it trapped.\\
   - \textbf{Interaction Errors}: Did it get trapped in a dialog box because it forgot to confirm the interaction? Did it interact with the wrong object?\\
   - Explain *why* the agent made this mistake based on its visual reasoning.\\
   \\
4. \textbf{Visual Quality \& Observation Bottlenecks}:\\
   Analyze whether the agent's perception was hindered by the raw image quality or structural composition.\\
   - \textbf{Illumination \& Color}: Was the environment too dark, too bright, or washed out?\\
   - \textbf{Scale \& Focus}: Were key objects too small, distant, or surrounded by irrelevant background?\\
   - \textbf{Clarity}: Were important details noisy or blurry?\\
   - State exactly which views (Previous Screen, Current Front, Left View, Right View) suffered from these visual bottlenecks.\\
\end{promptbox}
\caption{Prompt of Trajectory Summarization}
\label{prompt:summary}
\end{figure*}

\begin{figure*}[p]
\centering
\begin{promptbox}[Prompt 3: Cross-Trajectory Evolution]
\small
You are a master strategy optimizer for an agent playing Pokémon Legends: Z-A, relying purely on visual screenshots to complete long-horizon tasks.\\
\\
You are provided with the summaries of 5 recent rollout trajectories for the same task. These 5 trajectories might be all failures, all successes, or a mix of both. \\
\\
<task>\\
\{task\}\\
</task>\\
\\
<current\_actions>\\
\{current\_actions\}\\
</current\_actions>\\
\\
<trajectories>\\
\{trajectories\}\\
</trajectories>\\
\\
Your goal is to cross-analyze these 5 trajectories to distill high-level, generalizable strategies and experiences that can guide the agent's actions in future runs.\\
\\
Conduct your analysis following these steps:\\
1. \textbf{Cross-Trajectory Analysis}: \\
   - If there are both successes and failures: What critical decision or action separated the successful runs from the failed ones?\\
   - If all are failures: What is the common bottleneck?\\
   - If all are successes: What are the most robust patterns that led to success? How can it be done more efficiently?\\
2. \textbf{Abstracting to General Rules}: Translate specific observations into general spatial/visual rules. Focus heavily on 3D spatial reasoning, obstacle avoidance, and correct NPC interaction protocols.\\
3. \textbf{Deadlock Analysis \& Macro Action Design}: Identify specific physical deadlocks (e.g., stuck on corners, trapped by visually permeable barriers, oscillating between two states). If primitive actions fail to escape these, design new Macro Actions by combining existing functions in `<current\_actions>`. For example, a reliable deadlock escape might require moving backward multiple times before rotating.\\
\end{promptbox}
\caption{Prompt of Cross-Trajectory Evolution}
\label{prompt:evolution}
\end{figure*}

\begin{figure*}[p]
\ContinuedFloat
\centering
\begin{promptbox}[Prompt 3: Cross-Trajectory Evolution (continued)]
\small
4. \textbf{Visual Enhancement Design}: Analyze the visual observation bottlenecks from the summaries. Propose specific image processing operations to apply to the 4 camera views (Previous Screen, Current Screen (Front), Left View, Right View) to enhance future perception.\\
   Available operations and required parameter formats:\\
   - "Crop": \{\{"x": float, "y": float, "width": float, "height": float\}\} (normalized 0.0 to 1.0)\\
   - "CenterCrop": \{\{"scale": float\}\} (e.g., 0.8 for zooming into the center 80\%)\\
   - "Brightness": \{\{"factor": float\}\} (>1.0 to brighten, <1.0 to darken)\\
   - "Contrast": \{\{"factor": float\}\} (>1.0 to increase contrast)\\
   - "Saturation": \{\{"factor": float\}\} (>1.0 to boost colors)\\
   - "GammaCorrection": \{\{"gamma": float\}\}\\
   - "MedianBlur": \{\{"kernel\_size": int\}\} (e.g., 3 or 5, for noise reduction)\\
   - "Sharpen": \{\{"intensity": float\}\} (to enhance edges/details)\\
\\
Based on your analysis, extract at most 5 highly effective strategies, propose at most 5 necessary Macro Actions, and provide the optimal visual enhancement pipeline. \\
\\
You must output your response EXACTLY as a JSON object with the following three keys:\\
- "reasoning": A string containing your detailed step-by-step analysis based on the two steps above.\\
- "strategies": A list of strings, where each string is a clear, generalizable strategy based on the trajectories.\\
- "new\_macro\_actions": A list of objects detailing the new composite action functions.\\
- "visual\_enhancements": An object mapping each of the 4 views to a list of sequential image processing operations.\\
\\
\end{promptbox}
\caption{Prompt of Cross-Trajectory Evolution (continued)}
\end{figure*}

\begin{figure*}[p]
\ContinuedFloat
\centering
\begin{promptbox}[Prompt 3: Cross-Trajectory Evolution (continued)]
\small
Ensure your output is valid JSON format as follows:\\
```json\\
\{\{\\
    "reasoning": "Your step-by-step cross-trajectory analysis, abstraction process, and deadlock analysis.",\\
    "strategies": [\\
        "First extracted general strategy here.",\\
        "Second extracted general strategy here."\\
    ],\\
    "new\_macro\_actions": [\\
        \{\{\\
            "name": "Action\_BackAndTurnRight",\\
            "description": "A macro action designed to escape when stuck against a wall or obstacle. It forces the agent to back up significantly and turn to clear the collision box.",\\
            "code": "private static async Task Action\_EscapeDeadlock\_BackAndTurn()\textbackslash\{\}\textbackslash\{\}n\{\{\textbackslash\{\}\textbackslash\{\}n    Console.WriteLine(\textbackslash\{\}\textbackslash\{\}"[Action API] Macro: Escape Deadlock - Back and Turn\textbackslash\{\}\textbackslash\{\}");\textbackslash\{\}\textbackslash\{\}n    await Action\_MoveBackward();\textbackslash\{\}\textbackslash\{\}n    await Action\_MoveBackward();\textbackslash\{\}\textbackslash\{\}n    await Action\_RotateRight();\textbackslash\{\}\textbackslash\{\}n\}\}"\\
        \}\}\\
    ],\\
    "visual\_enhancements": \{\{\\
        "Previous Screen": [],\\
        "Current Screen (Front)": [\\
            \{\{ "operation": "CenterCrop", "parameters": \{\{"scale": 0.8\}\} \}\},\\
            \{\{ "operation": "Brightness", "parameters": \{\{"factor": 1.2\}\} \}\}\\
        ],\\
        "Left View": [\\
            \{\{ "operation": "Sharpen", "parameters": \{\{"intensity": 1.5\}\} \}\}\\
        ],\\
        "Right View": []\\
    \}\}\\
\}\}\\
```\\
\end{promptbox}
\caption{Prompt of Cross-Trajectory Evolution (continued)}
\end{figure*}


\end{document}